%% file: TAI_template.tex
\documentclass[journal]{IEEEtai}

\usepackage[colorlinks,urlcolor=blue,linkcolor=blue,citecolor=black]{hyperref}

\usepackage{color,array}

\usepackage{graphicx}

\usepackage[figuresright]{rotating}
\usepackage[numbers]{natbib}
\usepackage{amsmath,amssymb,amsfonts}
\usepackage{algorithmic}
\usepackage{graphicx}
\usepackage{hhline}
\usepackage{tabularx}
\usepackage{multirow}
\usepackage{booktabs}
\usepackage{textcomp}
\usepackage{xcolor}
\usepackage{tikz}
\usetikzlibrary{trees}
\usepackage{subcaption}
\usepackage{scalerel}
\usepackage{tikz-qtree}
\usepackage{lipsum,adjustbox}
\usepackage{hyperref}
\usepackage{titlesec}
\usepackage{diagbox}
\usepackage{makecell}
\usepackage{hhline}
\usepackage{rotating}
\usepackage{fixltx2e}
\usepackage{setspace}
\usepackage{booktabs}
\usepackage{array}
\usepackage{tabularx}
\usepackage{calc}
\usepackage{makecell}
\usepackage{mathtools}
\usepackage{url}
\usepackage{multirow}
\usepackage{graphicx}
\usepackage{ifthen}
\usepackage{xcolor}
\usepackage{pgfplots}
\usetikzlibrary{matrix,calc}
\usepackage{tikzsymbols}
\usepackage{tikz}
\usetikzlibrary{shapes}
\usetikzlibrary{calc, arrows, quotes}
\usepackage{pgfplots}
\pgfplotsset{compat=1.13, axis line style=thick}
\tikzstyle{arrow} = [thick,->,>=stealth]
\pgfplotsset{
  compat=newest,
  xlabel near ticks,
  ylabel near ticks
}

\usetikzlibrary{arrows, automata}
\newcommand{\hasper}{\textsc{HaSPeR}}
\newcommand{\haspers}{\textsc{HaSPeR} }
\def\hasperlogo{\raisebox{0.0\height}{\scalerel*{\includegraphics{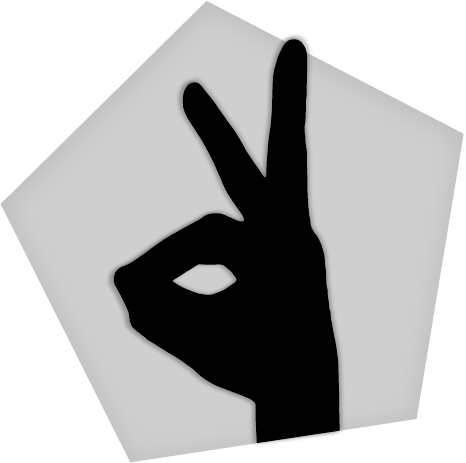}}{b}}}
\usepackage{academicons}
\usepackage{orcidlink}
\definecolor{orcidlogocol}{HTML}{A6CE39}
\usepackage{moresize}
\usepackage{array}
\newcommand{\PreserveBackslash}[1]{\let\temp=\\#1\let\\=\temp}
\newcolumntype{C}[1]{>{\PreserveBackslash\centering}p{#1}}
\newcolumntype{R}[1]{>{\PreserveBackslash\raggedleft}p{#1}}
\newcolumntype{L}[1]{>{\PreserveBackslash\raggedright}p{#1}}
\usepackage{siunitx}
\usepackage{float}
\usepackage{esvect}
\usepackage{epigraph}
\usepackage[activate=false,kerning]{microtype}
\SetExtraKerning[unit=character]{encoding=*}{\textemdash={100,100}}
\DeclareMathOperator*{\argmax}{arg\,max}
\DeclareMathOperator*{\argmin}{arg\,min}
\definecolor{ForestGreen}{rgb}{0.13, 0.55, 0.13}
\begin{document}

\title{{\HUGE\hasperlogo}\,\textsc{HaSPeR}: An Image Repository for Hand Shadow Puppet Recognition} 


\author{Syed Rifat Raiyan\textsuperscript{\textdagger} \orcidlink{0009-0004-3558-3418}, Zibran Zarif Amio \orcidlink{0009-0000-5575-0819}, and Sabbir Ahmed \orcidlink{0000-0001-5928-4886}
\thanks{\textsuperscript{\textdagger}Corresponding Author}
\thanks{Syed Rifat Raiyan is affiliated with the Systems and Software Lab (SSL) of the Department of Computer Science and Engineering at the Islamic University of Technology, Boardbazar, Gazipur-1704, Dhaka, Bangladesh (e-mail: \href{mailto:rifatraiyan@iut-dhaka.edu}{rifatraiyan@iut-dhaka.edu}).}
\thanks{Zibran Zarif Amio, was affiliated with the Networking Research Group of the Department of Computer Science and Engineering at the Islamic University of Technology, Boardbazar, Gazipur-1704, Dhaka, Bangladesh. He is now with BrillMark LLC., 192 Benmore Dr, Hayward 94542, California, USA. (e-mail: \href{mailto:zibranzarif@iut-dhaka.edu}{zibranzarif@iut-dhaka.edu}).}
\thanks{Sabbir Ahmed is affiliated with the Computer Vision Lab (CVLab) of the Department of Computer Science and Engineering at the Islamic University of Technology, Boardbazar, Gazipur-1704, Dhaka, Bangladesh (e-mail: \href{mailto:sabbirahmed@iut-dhaka.edu}{sabbirahmed@iut-dhaka.edu}).}
}
\markboth{IEEE/CVF International Conference on Computer Vision (ICCV) Workshop on Cultural Continuity of Artists (WCCA 2025)}
{Raiyan \MakeLowercase{\textit{et al.}}: \textsc{HaSPeR}: An Image Repository for Hand Shadow Puppet Recognition}

\maketitle

\begin{abstract}
Hand shadow puppetry, also known as shadowgraphy or ombromanie, is a form of theatrical art and storytelling where hand shadows are projected onto flat surfaces to create illusions of living creatures. The skilled performers create these silhouettes by hand positioning, finger movements, and dexterous gestures to resemble shadows of animals and objects. Due to the lack of practitioners and a seismic shift in people's entertainment standards, this art form is on the verge of extinction. To facilitate its preservation and proliferate it to a wider audience, we introduce \textsc{HaSPeR}, a novel dataset consisting of 15,000 images of hand shadow puppets across 15 classes extracted from both professional and amateur hand shadow puppeteer clips. We provide a detailed statistical analysis of the dataset and employ a range of pretrained image classification models to establish baselines. Our findings show a substantial performance superiority of skip-connected convolutional models over attention-based transformer architectures. We also find that lightweight models, such as \textsc{MobileNetV2}, suited for mobile applications and embedded devices, perform comparatively well. We surmise that such low-latency architectures can be useful in developing ombromanie teaching tools, and we create a prototype application to explore this surmission. Keeping the best-performing model \textsc{ResNet34} under the limelight, we conduct comprehensive feature-spatial, explainability, and error analyses to gain insights into its decision-making process and explore architectural improvements. To the best of our knowledge, this is the first documented dataset and research endeavor to preserve this dying art for future generations, with computer vision approaches. Our code and data are publicly available at \url{https://github.com/Starscream-11813/HaSPeR}.
\end{abstract}

\begin{IEEEImpStatement}
This research is an impetus towards utilizing AI tools to revitalize the hitherto underexplored cinematic art form of hand shadow puppetry. Such tools may help understand the creativity frontier in generative models, facilitate the development of applications to teach shadowgraphy, and unveil several prospects for entertainment. The existing works, though distally relevant to shadowgraphy, 
explore the digitization of such precursory art forms via approaches that have since been rendered primitive and obsolete. The novel dataset that we introduce in this paper, namely \hasper, consists of 15,000 diverse samples garnered from performance clips of variably skilled puppeteers. Our extensive benchmarking reveals that the task of classifying the puppet silhouettes is reasonably solvable using lightweight and convolutional feature extractor models, with accuracies of $94.97\%$ by \textsc{ResNet34} and $92.38\%$ by \textsc{MobileNetV2}. \hasper, as a data resource for all intents and purposes, can be a potential stride towards systematically preserving this artistic practice.
\end{IEEEImpStatement}

\setlength{\epigraphwidth}{0.9\linewidth}
\vspace{-4mm}
\begin{flushleft}
\epigraph{``Will he not fancy that the shadows which he formerly saw are truer than the objects which are now shown to him?"}{Plato, \textit{The Republic} (Book VII, Allegory of the Cave)}
\end{flushleft}
\input{sections/1_Introduction}

\input{sections/2_Literature_Review}
\input{sections/3_Dataset_Construction}

\input{sections/4_Statistical_Analysis}

\input{sections/5_Developing_Benchmark}
\input{sections/7_Conclusion}
\bibliographystyle{IEEEtranN}
\bibliography{ref2}
\newpage
\appendices
\section{Experimental Setup: Additional Details}
\label{appendix:exp_setup}
\subsection{\textsc{ResNet34} Architectural Enhancements}
\subsubsection{Silhouette Polygonization}
We augment \textsc{ResNet34} with handcrafted polygonal features extracted from the silhouette contours of hand shadow puppets, using Douglas-Peucker approximation \cite{douglas1973algorithms} and geometric shape descriptors \cite{zhang2004review}. These features capture structural cues such as convexity, angularity, and polygonal regularity that complement the visual representations learned by the CNN feature extractor model. Hand shadow puppets inherently possess distinct geometric forms and outlines, which traditional CNNs might implicitly learn but not explicitly represent. To augment the classification process with this crucial geometric information, we test the integration of features derived from polygonal approximations and contour analysis.
\begin{figure}[h]
    \includegraphics[width=1.0\linewidth]{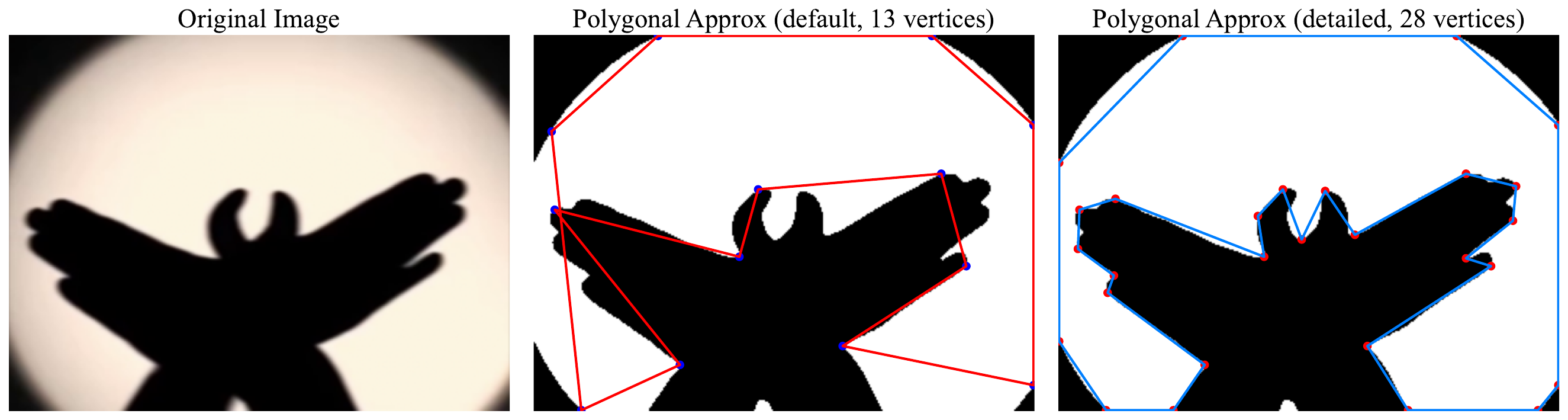}
    \caption{Comparison of polygonal approximations for a hand shadow puppet silhouette.}
    \label{fig:polygon_comparison}
\end{figure}
\begin{enumerate}
    \item \textbf{Silhouette Preprocessing:} The input image is first converted to grayscale and then binarized using Otsu's method \cite{otsu1979threshold}. Morphological closing and opening operations \cite{serra1982image} are subsequently applied to remove small holes, smooth contours, and eliminate noise, ensuring a clean and consistent silhouette.
    \item \textbf{Basic Contour Properties:} For the largest detected contour (representing the main puppet silhouette), fundamental geometric properties (area, perimeter, compactness, aspect ratio, solidity) are computed using OpenCV \cite{opencv_library}.
    \item \textbf{Convexity Defects Analysis:} Hand shadow puppets often feature distinct \textit{``finger-like"} protrusions and indentations. We analyze convexity defects (the regions between the contour and its convex hull) to quantify these characteristics. Features include the number of significant defects, as well as the mean, standard deviation, and maximum depth of these defects. A specific check is incorporated to identify \textit{``finger-like"} defects based on their depth relative to the perimeter and the sharpness of the angle at the defect point.
    \item \textbf{Hu Moments:} These are seven scale, rotation, and translation-invariant moments derived from the central moments of the contour \cite{hu1962visual}. Hu moments are powerful shape descriptors, capturing intrinsic characteristics regardless of the puppet's position, size, or orientation in the image. We use their logarithmically scaled values for better numerical stability.
    \item \textbf{Ellipse Fitting:} An ellipse is fitted to the main contour, and the ratio of the contour's area to the fitted ellipse's area is computed. This metric provides insight into how well the silhouette can be approximated by an elliptical shape.
    \item \textbf{Polygon Approximation Features:} The Douglas-Peucker algorithm (\texttt{cv2.approxPolyDP}) \cite{douglas1973algorithms} is used to simplify the contour into a polygon with a reduced number of vertices, controlled by an epsilon ($\epsilon$) factor (as evident in \ref{fig:polygon_comparison}).
    \item \textbf{Skeleton-Based Features:} A morphological skeleton of the binary silhouette is generated \cite{lam1992thinning}. This skeleton represents the medial axis of the shape and provides valuable structural information.
    \begin{itemize}
        \item \textit{Skeleton Ratio:} The ratio of skeleton pixels to total silhouette pixels, indicating the \textit{``thinness"} of the shape.
        \item \textit{Branch Points and Endpoints:} The counts of these critical points on the skeleton, which correspond to junctions and extremities (like fingertips) in the hand shadow puppet.
    \end{itemize}
\end{enumerate}
These features collectively form a robust and comprehensive polygonal descriptor vector, explicitly encoding geometric characteristics that are highly relevant for distinguishing between different hand shadow puppet forms.
\subsubsection{Topological Features}
This variant integrates \textsc{ResNet34} with topological descriptors derived from skeletonized silhouettes, including branch/end-point counts and skeleton-to-area ratios \cite{blum1967transformation, lam1992thinning}. Such features model the internal articulation and connectivity of hand shapes, enabling finer discrimination of visually similar gesture classes. While CNNs excel at extracting hierarchical and abstract visual patterns, they may not explicitly capture the fundamental \textit{``shape"} or connectivity of objects, which is crucial for silhouette-based recognition. To address this, we integrate topological features derived from persistent homology \cite{edelsbrunner2008computational} into the classification pipeline. The topological feature extraction module is designed to quantify intrinsic shape properties invariant to continuous deformations, such as stretching or bending. This is particularly relevant for hand shadow puppets, where variations in hand posture can alter geometric appearance while preserving the underlying topological form.
\begin{enumerate}
    \item \textbf{Betti Curves (Simplified Persistent Homology):} We approximate Betti numbers by analyzing the image at various filtration levels (thresholds from 0 to 1). $\beta_0$ is the number of connected components, reflecting the fragmentation or unity of the silhouette. $\beta_1$ is an estimation of the number of \textit{``holes"} or loops within the silhouette. We derive this by considering the difference in pixel counts between the binary image and its morphologically filled counterpart, normalized to reduce sensitivity to small noise. These Betti curves provide a multi-scale topological signature of the image.
    \item \textbf{Critical Points Analysis:} We identify local maxima and minima within the smoothed grayscale image. The counts and densities (normalized by total pixels) of these critical points offer insights into the image's \textit{``peaks"} and \textit{``valleys,"} which correspond to salient features of the silhouette's shape.
    \item \textbf{Morphological Features:} Standard morphological operations, opening and closing, are applied to the binary silhouette at various kernel sizes (3, 5, 7). The ratio of pixels in the opened/closed image to the original binary image's pixel count provides measures of the object's robustness to small protrusions/indentations and its overall compactness.
    \item \textbf{Euler Characteristic at Multiple Scales:} The Euler characteristic ($\chi=\text{connected components}-\text{holes}$) is a fundamental topological invariant. We compute this characteristic at different binarization thresholds (0.3, 0.5, 0.7) to capture how the global topology of the silhouette evolves across different levels of detail.
    \item \textbf{Gradient-Based Features:} To capture edge information, Sobel filters are used to compute horizontal and vertical gradients. Statistical properties (mean, standard deviation, 90th percentile) of the gradient magnitude provide a summary of the image's edge strength and complexity.
    \item \textbf{Contour-Based Features:} Utilizing OpenCV's contour detection, we extract features directly from the silhouette's boundaries. This includes the number of distinct contours, and the mean and standard deviation of their areas and perimeters. These features directly characterize the complexity and size of the hand shadow's outline.
\end{enumerate}
These diverse topological features are concatenated into a single, fixed-dimension vector, designed to provide a comprehensive, invariant representation of the silhouette's inherent shape.
\input{sections/8_Acknowledgments}
\newpage
\begin{IEEEbiography}[{\includegraphics[width=1in,height=1.25in,clip,keepaspectratio]{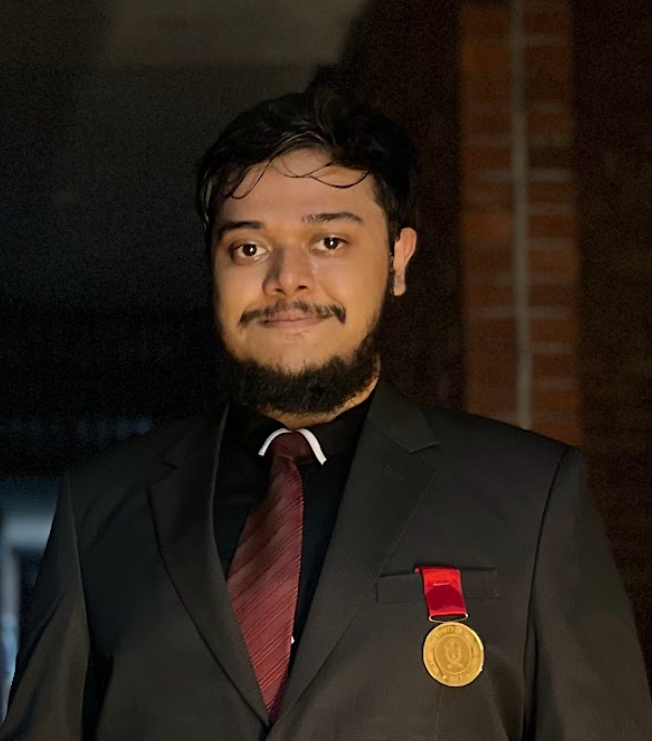}}]{Syed Rifat Raiyan} was born in Agrabad, Chattogram, Bangladesh in 2001. His alma maters include PlaySchool, Milestone College, RAJUK Uttara Model College, and Notre Dame College. He attained the Bachelor of Science (B.Sc.) degree with Honors from the Department of Computer Science and Engineering (CSE) at the Islamic University of Technology (IUT), Board Bazar, Gazipur-1704, Dhaka, in 2023.

He worked as an Industrial Trainee at Battery Low Interactive Ltd. in 2021. Since the 16th of August 2023, he has been working as a Lecturer in the Department of Computer Science and Engineering (CSE) at the Islamic University of Technology (IUT). He is currently affiliated with the Systems and Software Lab (SSL) research group of IUT. His research interests lie broadly in natural language processing, computer vision, and deep learning. His works have been published in the Findings of the Association for Computational Linguistics: ACL 2023 and in the Proceedings of the 61st Annual Meeting of the Association for Computational Linguistics (Volume 4: Student Research Workshop). His current endeavors revolve around projects concerning mathematical reasoning in language models, in-context learning for text classification, Bangla sentiment analysis, and image classification.

Mr. Raiyan has been a member of the Association for Computational Linguistics (ACL) since 2023. He was awarded the prestigious IUT Gold Medal in recognition of his stellar academic performance while pursuing his B.Sc. (Engg.) degree, in 2023.
    
\end{IEEEbiography}

\begin{IEEEbiography}[{\includegraphics[width=1in,height=1.25in,clip,keepaspectratio]{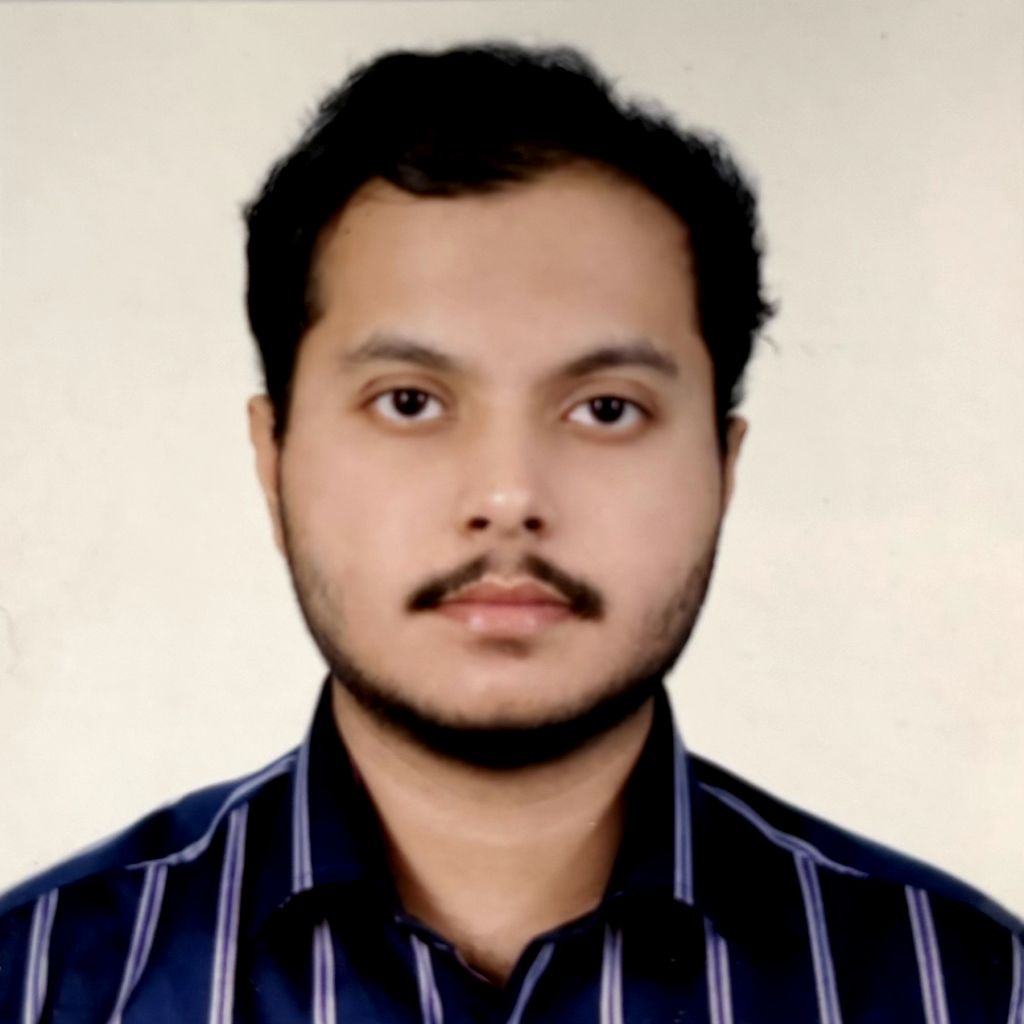}}]{Zibran Zarif Amio} was born in Dulahazara, Cox's Bazar, Bangladesh in 2000. His alma mater includes BIAM Laboratory School, Cox's Bazar Govt. High School, and Dhaka Residential Model College. He attained the Bachelor of Science (B.Sc.) degree with First Class from the Department of Computer Science and Engineering (CSE) at the Islamic University of Technology (IUT), Board Bazar, Gazipur-1704, Dhaka, in 2023.

He worked as an Android Developer at BYDO Academy in 2020. He was an Industrial Trainee at Battery Low Interactive Ltd. in 2021. Since the 1st of June 2023, he has been working as the Chief Technology Officer at ReplyMind AI Ltd., Grameen Telecom Bhaban, Mirpur-1216, Dhaka. His work involves leading and developing Generative AI-enabled software solutions. His research interests lie broadly in natural language processing, computer vision, and deep learning. His current endeavors include Meta Business Automation with Large Language Models.
\end{IEEEbiography}




\begin{IEEEbiography}[{\includegraphics[width=1in,height=1.25in,clip,keepaspectratio]{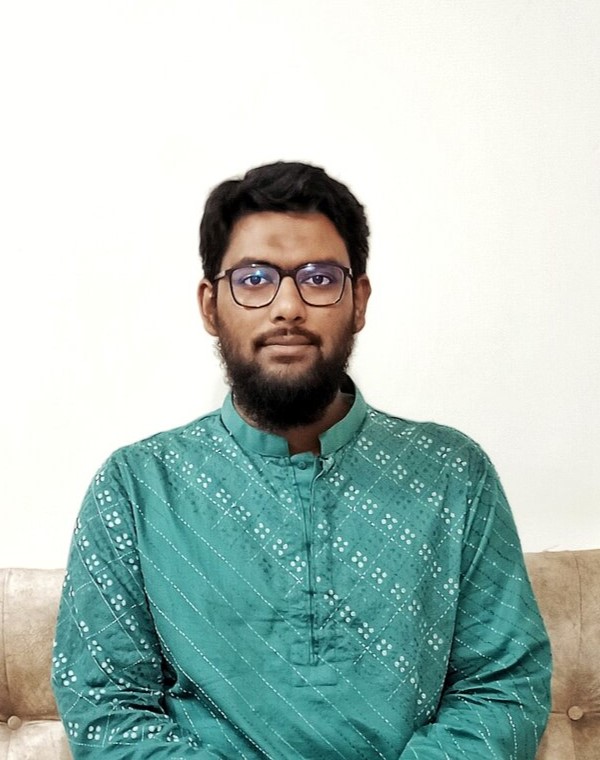}}]{Sabbir Ahmed} 
  was born in Dhaka,
Bangladesh, in 1996. He received the B.Sc. Engg.
degree (Hons.) in computer science (CS) from the
Islamic University of Technology (IUT), Gazipur,
Bangladesh, in 2017, where he is pursuing the
M.Sc. degree in CS.

From 2018 to 2022, he was a Lecturer
of the Department of Computer Science and
Engineering, IUT, and since 2022 he has been working as an Assistant Professor. His research interests include
pattern recognition, deep learning in computer
vision, and intelligent agriculture. He received the IUT Gold Medal for his B.Sc. Engg. degree.
    
\end{IEEEbiography}
\end{document}

%% file: sections/1_Introduction.tex
 \begin{figure}[t]
    \centering
    \subfloat[A generic hand shadow puppetry setup.]{\includegraphics[width=0.49\textwidth]{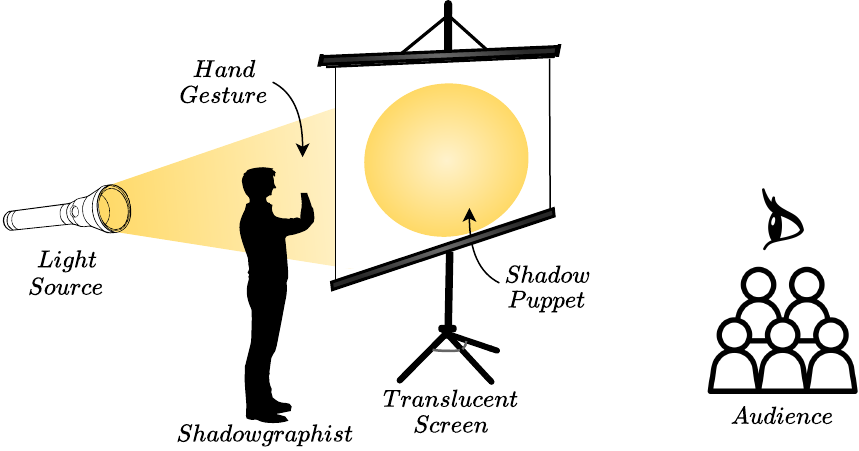}}\\
    \vspace{2mm}
    \subfloat[Rabbit]{\includegraphics[scale=0.4125]{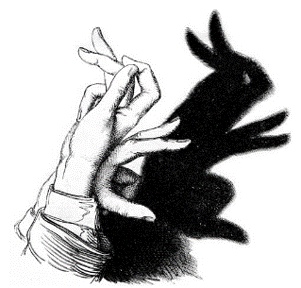}}\hspace{0.1cm}
    \hfil
    \subfloat[Bird]
    {\includegraphics[scale=0.4125]{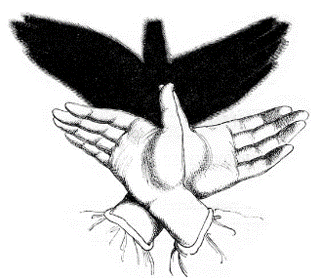}}
    \hfil
    \subfloat[Dog]
    {\includegraphics[scale=0.4125]{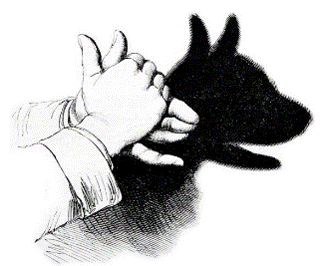}}
    \caption{Ombromanie in a nutshell.\protect\footnotemark}
    \label{fig:fig1}
\end{figure}
\footnotetext{The shadowgraphy cliparts are adapted from ClipArt ETC, Florida Center for Instructional Technology, College of Education, University of South Florida. Link: \url{https://etc.usf.edu/clipart/galleries/266-hand-shadow-puppetry}}
\section{Introduction}
\label{sec:introduction}
Ombromanie, the ancient art of hand shadow puppetry, is a form of art that involves the mesmerizing interplay of light and shadow through the construction and manipulation of shadow figures or silhouettes on a surface, typically a screen or a wall, using one's hands, body, or props \cite{almoznino2002art,Sraj_2012}. The alias \textit{``cinema in silhouette"}\footnote{
\url{https://en.wikipedia.org/wiki/Shadowgraphy_(performing_art)}} is sometimes used to refer to this proto-cinematic medium of entertainment. Its working principle is very straightforward---the puppeteer adeptly positions their hands between a radiant light source and a translucent screen, consequently conjuring shadows and silhouettes that emulate different creatures,
 as shown in Figure \ref{fig:fig1}.

Despite its rich history and captivating allure across many cultures,\footnote{
\url{https://www.geniimagazine.com/wiki/index.php/Shadowgraphy}} there exists a notable dearth of resources specifically tailored to this artistic domain. With properly annotated and sourced data, researchers could study the intricacies of hand silhouette movements, shapes, and storytelling techniques, thereby enabling the development of sophisticated Artificial Intelligence (AI) systems for automatic recognition, classification, or even generation of ombromanie performances \cite{maerten2023paintbrush}. The generation aspect is particularly relevant given the demonstrable impotency of AI image generator models in accurately creating hands and fingers \cite{samuel2024generating}. Apart from that, the development of applications that can facilitate the learning of ombromanie has the potential to breathe new life into this waning art form \cite{rsaritha2017nurturing}. In 2011, UNESCO recognized shadow puppetry as an endangered artistic tradition by adding it to the Intangible Cultural Heritage list \cite{lu2011shadowstory}, which is why it necessitates more preservatory apparatus and research efforts.

In tandem with this motivation, this work introduces a seminal addition to the realm of data resources, \textsc{HaSPeR} (\textbf{Ha}nd \textbf{S}hadow \textbf{P}upp\textbf{e}t Image \textbf{R}epository), a methodically curated novel image dataset of hand shadow puppets. The dataset comprises an assemblage of 15,000 samples, that we painstakingly source and verify from 68 professional shadowgraphist clips and 90 amateur shadowgraphist clips. We label and categorize the images with utmost precision to elicit robustness in the image classification models that will undergo training with these images. The samples in \haspers are diverse in nature since the source clips are recorded in a plethora of different poses, orientations, and background lighting conditions of the translucent screen. We also inculcate silhouette motion diversity via optical flow estimation \cite{Horn_1981} in the frame extraction process. We conduct a detailed analysis of \textsc{HaSPeR}'s statistical characteristics. We also employ a variety of state-of-the-art (SOTA) pretrained image classification models to establish a performance benchmark for validating the integrity of the dataset. Additionally, we conduct a thorough evaluation of several facets of the ace \textsc{ResNet34} model, including its feature representations, feature fusions, interpretability, explainability, and classification errors that it encounters. In an effort to assess the potential of digitized ombromanie teaching tools, we create a simple and lightweight prototype Android application using Flutter for classifying hand shadow puppet images from the phone's camera feed. We posit that our dataset possesses the potential to offer a wealth of opportunities for exploration and analysis into the artistic domain of hand shadow puppetry.
\vspace{-1mm}

%% file: sections/2_Literature_Review.tex
\begin{figure*}

        \centering

        \includegraphics[width=0.985\textwidth]{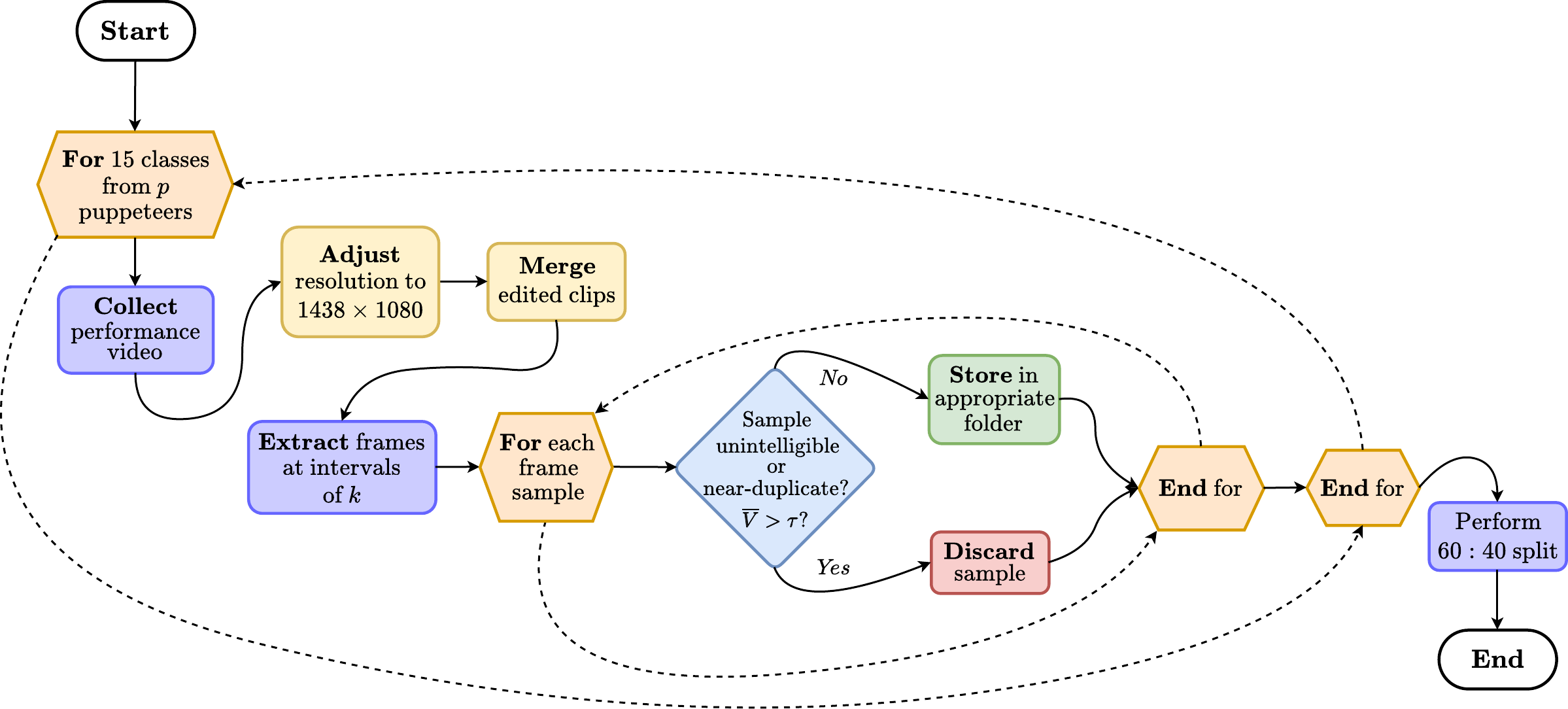}

        \caption{A flowchart depicting the dataset construction process.}

        \label{fig:fig2}

\end{figure*}

\section{Literature Review}

\label{sec:lit_review}

The recognition and classification of hand shadow puppet images are intriguing problemspaces in the context of deep learning, albeit relatively underexplored.
After rigorously analyzing the existing pool of research on the topic, we could identify several quasi-related works.

\vspace{-2mm}
\subsection{Image Classification and Recognition}
Among the pioneering endeavors in hand shadow image classification was that of \citet{huang1999shadow}, who created \textsc{Shadow Vision}---a system to emulate an immersive virtual shadow puppet theater experience, employing a user's hand gestures over an overhead projector to control the creation and manipulation of objects within a 3D Open Inventor\footnote{Open Inventor™ toolkit --- \url{https://www.openinventor.com/}} environment. The chain of stages underlying the implementation of \textsc{Shadow Vision} were acquisition, segmentation, feature extraction, and recognition of the infrared shadow puppet images. 
They also adopted a 3-layer neural network and the centralized contour moments modeling technique, using 13 features (7 moments of the object, length, angle, and the 4 endpoints of the axis of inertia). The data used for this study isn't publicly available, and 
the methodology can be deemed somewhat obsolete in the modern purview, due to being supplanted by the emergence of deep learning models. 
Some recent works explore different convolutional models to assess their efficacy in Indonesian shadow puppet recognition. 
\citet{sudiatmika2018indonesian,sudiatmika2021indonesian} used the deep CNN models, \textsc{AlexNet} \cite{krizhevsky2014one} and \textsc{VGG-16} \cite{simonyan2014very}, and constructed a dataset of 2,530 images spanning 6 classes of puppets 
from museums in Bali. 
They also experimented with other convolutional models, such as \textsc{Mask R-CNN} \cite{he2017mask} and \textsc{MobileNet} \cite{howard2017mobilenets}, in two separate studies \cite{sudiatmika2021mask,prabowo2021indonesian}.

In a similar spirit, our work is an endeavor towards establishing a performance benchmark of the recent SOTA feature extractor models for hand shadow puppet contour images, in a more large-scale and comprehensive manner.

\vspace{-2mm}

\subsection{3D Modeling and Human Motion Capture}

One of the earliest works involving silhouettes is a study by \citet{brand1999shadow} that explored the mapping of monocular monochromatic 2D shadow image sequences of humans to animated 3D body poses, using a configural and dynamical manifold created from data with a topologically special hidden Markov model (HMM), acquired via the process of entropy minimization without resorting to any articulatory body model. Several advances in vision-based human motion capture and analysis since then have leveraged human silhouette templates \cite{collins2002silhouette,moeslund2006survey}, more specifically, hand and finger silhouettes \cite{tsuji2018real,tsuji2019real,tsuji2021telecommunication}.

\vspace{-2mm}

\subsection{Robotics}

\citet{huang2017shadow} introduced computer vision-aided shadow puppetry with robotics by matching shape correspondences of input images. They claimed that due to the physical limitations of human arms, it is often not feasible to construct complex shadow forms. Instead, they developed a framework that enabled them to produce shadow images with the mechanical arms of a robot. The authors built a library of shadow images and used them to orient the robotic arms into a formation resembling the intended shadow puppet. The data used for this study isn't publicly available.

\vspace{-2mm}

\subsection{Human-Computer Interaction}

The authors of \cite{zhang2012chinese} proposed a framework for controlling two Chinese shadow puppets---a human model and an animal model, with the use of body gestures via a Microsoft Kinect sensor. \citet{carr2014shadow} conducted a similar work by building a real-time Indonesian shadow puppet storytelling application that is capable of mimicking the full-body actions of the user, using the Microsoft Kinect sensor. In order to leverage contactless gesture recognition (CGR) to teach traditional Chinese shadow puppetry to beginners, \citet{tsai2016study} developed a system using Leap Motion sensors. These studies on digitizing the art of shadow puppetry, or puppetry in general, were influenced to some extent by other similar works in the gesture recognition domain \cite{gudukbay2000beyond,gao2011digital,lu2011shadowstory,liang2015exploitation,yan2016interactive,liang2017hand}. 
\citet{tang2023intelligent} developed an intelligent shadow play system, called \textsc{ShadowTouch}, which includes a multidimensional somatosensory interaction module coupled with an automatic choreography module, to facilitate natural interaction between the shadow play figures and the human users.

The motif of our work tessellates well with the core objectives of the aforementioned research works. The utilization of digitized traditional arts serves as a means to preserve their inherent legacies, and \haspers can be a potent contribution to the contemporary pool of resources to facilitate such innovative digitization for ombromanie.

%% file: sections/3_Dataset_Construction.tex
\section{Dataset Construction}
\label{sec:dataset_construction}
The series of steps involved in our data acquisition process is broadly divided into three tasks---(a) procuring the performance clips, (b) extraction of the frames, and (c) categorization of each sample frame with a proper label. Figure \ref{fig:fig2} portrays this workflow behind our dataset preparation. We incorporate manual oversight at each step of the dataset creation in order to reconcile any exigencies pertaining to the quality of \hasper.
\begin{table}[t]
    \caption{Statistical summary of \hasper.}
    \centering
    \small
    \def\arraystretch{1.1}
    \begin{tabular}{|c|cc|ccc|}
    \Xhline{2\arrayrulewidth}
    \multirow{2}{*}{\textbf{\begin{tabular}[c]{@{}c@{}}Silhouette\\ Class\end{tabular}}} & \multicolumn{2}{c|}{\textbf{Clips}} & \multicolumn{3}{c|}{\textbf{Sample Distribution}} \\ \cline{2-6} 
     & \multicolumn{1}{c|}{\textbf{Pro.}} & \textbf{Nov.} & \multicolumn{1}{c|}{\textbf{\begin{tabular}[c]{@{}c@{}}Training\end{tabular}}} & \multicolumn{1}{c|}{\textbf{\begin{tabular}[c]{@{}c@{}}Validation\end{tabular}}} & \multicolumn{1}{c|}{\textbf{\begin{tabular}[c]{@{}c@{}}Total\end{tabular}}} \\ \Xhline{2\arrayrulewidth}
    Bird & \multicolumn{1}{c|}{6} & 6 & \multicolumn{1}{c|}{600} & \multicolumn{1}{c|}{400} & 1000 \\ \hline
    Chicken & \multicolumn{1}{c|}{2} & 6 & \multicolumn{1}{c|}{600} & \multicolumn{1}{c|}{400} & 1000 \\ \hline
    Cow & \multicolumn{1}{c|}{2} & 6 & \multicolumn{1}{c|}{600} & \multicolumn{1}{c|}{400} & 1000 \\ \hline
    Crab & \multicolumn{1}{c|}{4} & 6 & \multicolumn{1}{c|}{600} & \multicolumn{1}{c|}{400} & 1000 \\ \hline
    Deer & \multicolumn{1}{c|}{6} & 6 & \multicolumn{1}{c|}{600} & \multicolumn{1}{c|}{400} & 1000 \\ \hline
    Dog & \multicolumn{1}{c|}{7} & 6 & \multicolumn{1}{c|}{600} & \multicolumn{1}{c|}{400} & 1000 \\ \hline
    Elephant & \multicolumn{1}{c|}{5} & 6 & \multicolumn{1}{c|}{600} & \multicolumn{1}{c|}{400} & 1000 \\ \hline
    Horse & \multicolumn{1}{c|}{8} & 6 & \multicolumn{1}{c|}{600} & \multicolumn{1}{c|}{400} & 1000 \\ \hline
    Llama & \multicolumn{1}{c|}{2} & 6 & \multicolumn{1}{c|}{600} & \multicolumn{1}{c|}{400} & 1000 \\ \hline
    Moose & \multicolumn{1}{c|}{3} & 6 & \multicolumn{1}{c|}{600} & \multicolumn{1}{c|}{400} & 1000 \\ \hline
    Panther & \multicolumn{1}{c|}{2} & 6 & \multicolumn{1}{c|}{600} & \multicolumn{1}{c|}{400} & 1000 \\ \hline
    Rabbit & \multicolumn{1}{c|}{4} & 6 & \multicolumn{1}{c|}{600} & \multicolumn{1}{c|}{400} & 1000 \\ \hline
    Snail & \multicolumn{1}{c|}{4} & 6 & \multicolumn{1}{c|}{600} & \multicolumn{1}{c|}{400} & 1000 \\ \hline
    Snake & \multicolumn{1}{c|}{3} & 6 & \multicolumn{1}{c|}{600} & \multicolumn{1}{c|}{400} & 1000 \\ \hline
    Swan & \multicolumn{1}{c|}{10} & 6 & \multicolumn{1}{c|}{600} & \multicolumn{1}{c|}{400} & 1000 \\ \Xhline{2\arrayrulewidth}
    \multirow{2}{*}{\textbf{Total}} & \multicolumn{1}{c|}{68} & 90 & \multicolumn{1}{c|}{\multirow{2}{*}{9000}} & \multicolumn{1}{c|}{\multirow{2}{*}{6000}} & \multirow{2}{*}{15000} \\ \cline{2-3}
     & \multicolumn{2}{c|}{158} & \multicolumn{1}{c|}{} & \multicolumn{1}{c|}{} &  \\ \Xhline{2\arrayrulewidth}
    \end{tabular}
    \vspace{-5mm}
    \label{tab:tab1}
    \end{table}
\subsection{Collating Shadowgraphy Clips}
At the outset of the process, we procure 68 different clips of 14 different professional shadowgraphists from YouTube.\footnote{\url{https://www.youtube.com}} The video sources are licensed under fair use and a list consisting of the links to all of them is available in our GitHub\footnote{GitHub repository --- \url{https://github.com/Starscream-11813/HaSPeR}} repository. 
We record the relevant portions of the performance videos using the open-source recording software OBS Studio.\footnote{Open Broadcaster Software\textsuperscript{\textregistered} --- \url{https://obsproject.com/}} Six novice volunteer shadowgraphists collectively produce 90 additional clips, with each contributing one clip for every class. As a consequence, the total number of source clips aggregates to $68 + (15 \times 6) = 158$.
\begin{figure*}[t]
    \centering
    \subfloat[Bird]{\includegraphics[width=0.19\linewidth]{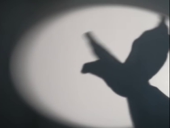}\label{fig:fig3a}}
    \hfil
    \subfloat[Chicken]{\includegraphics[width=0.19\linewidth]{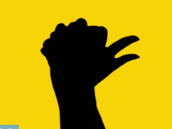}\label{fig:fig3b}}
    \hfil
    \subfloat[Cow]
    {\includegraphics[width=0.19\linewidth]{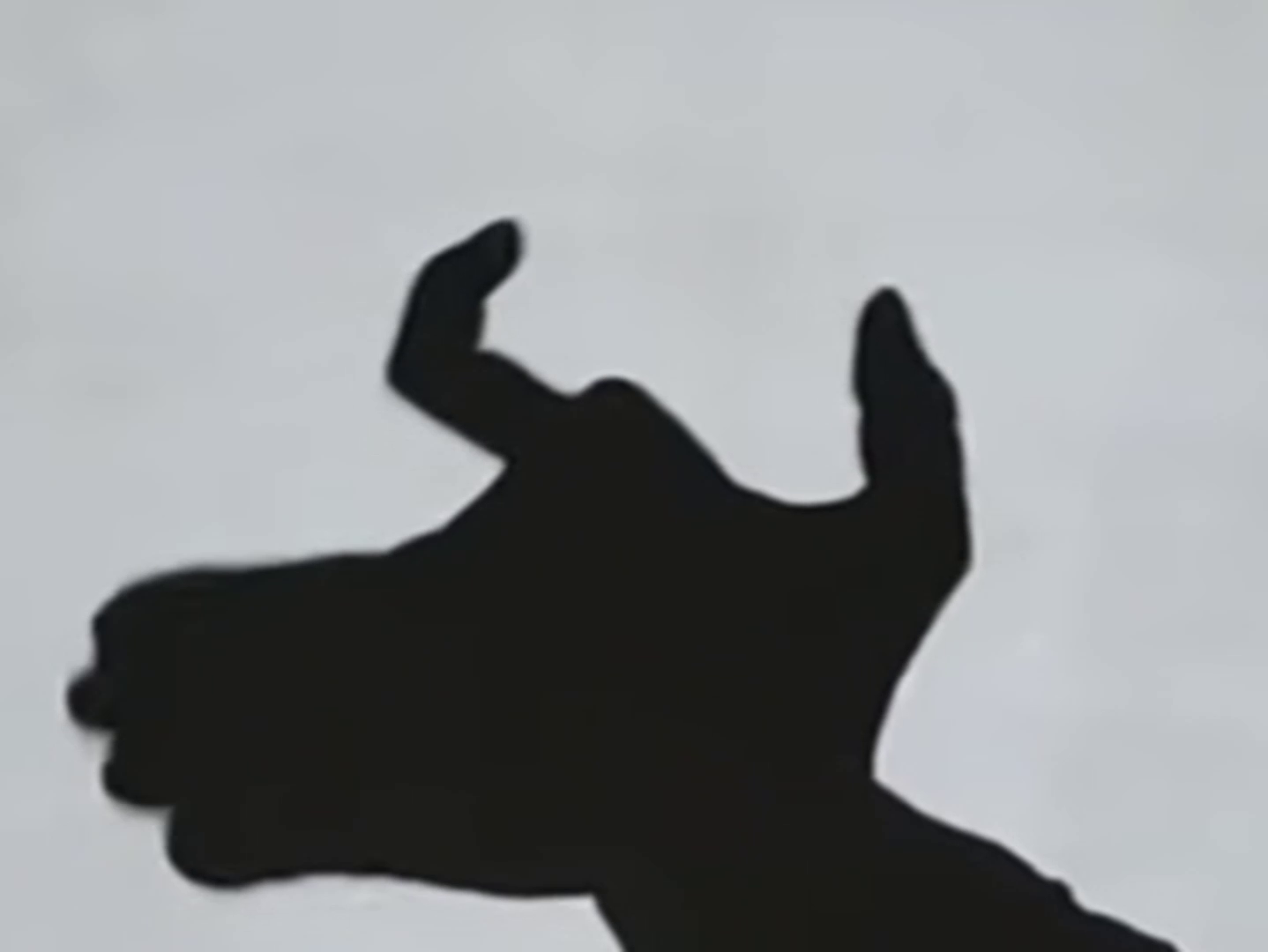}\label{fig:fig3c}}
    \hfil
    \subfloat[Crab]
    {\includegraphics[width=0.19\linewidth]{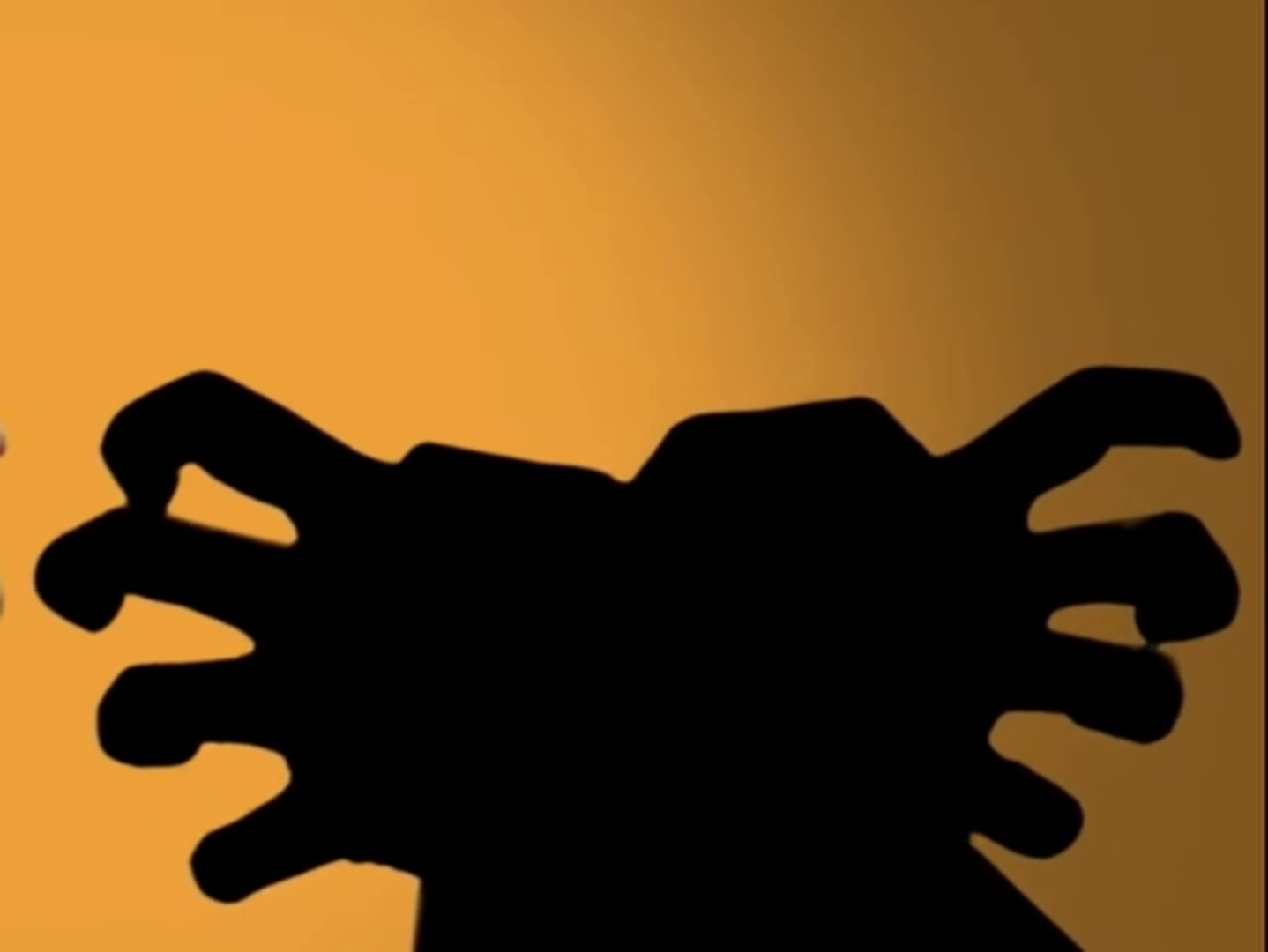}\label{fig:fig3d}}
    \hfil
    \subfloat[Deer]
    {\includegraphics[width=0.19\linewidth]{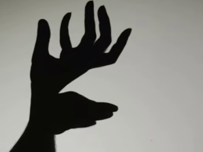}\label{fig:fig3e}}\\
    \vspace{1.1mm}
    \subfloat[Dog]
    {\includegraphics[width=0.19\linewidth]{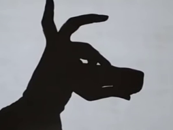}\label{fig:fig3f}}
    \hfil
    \subfloat[Elephant]
    {\includegraphics[width=0.19\linewidth]{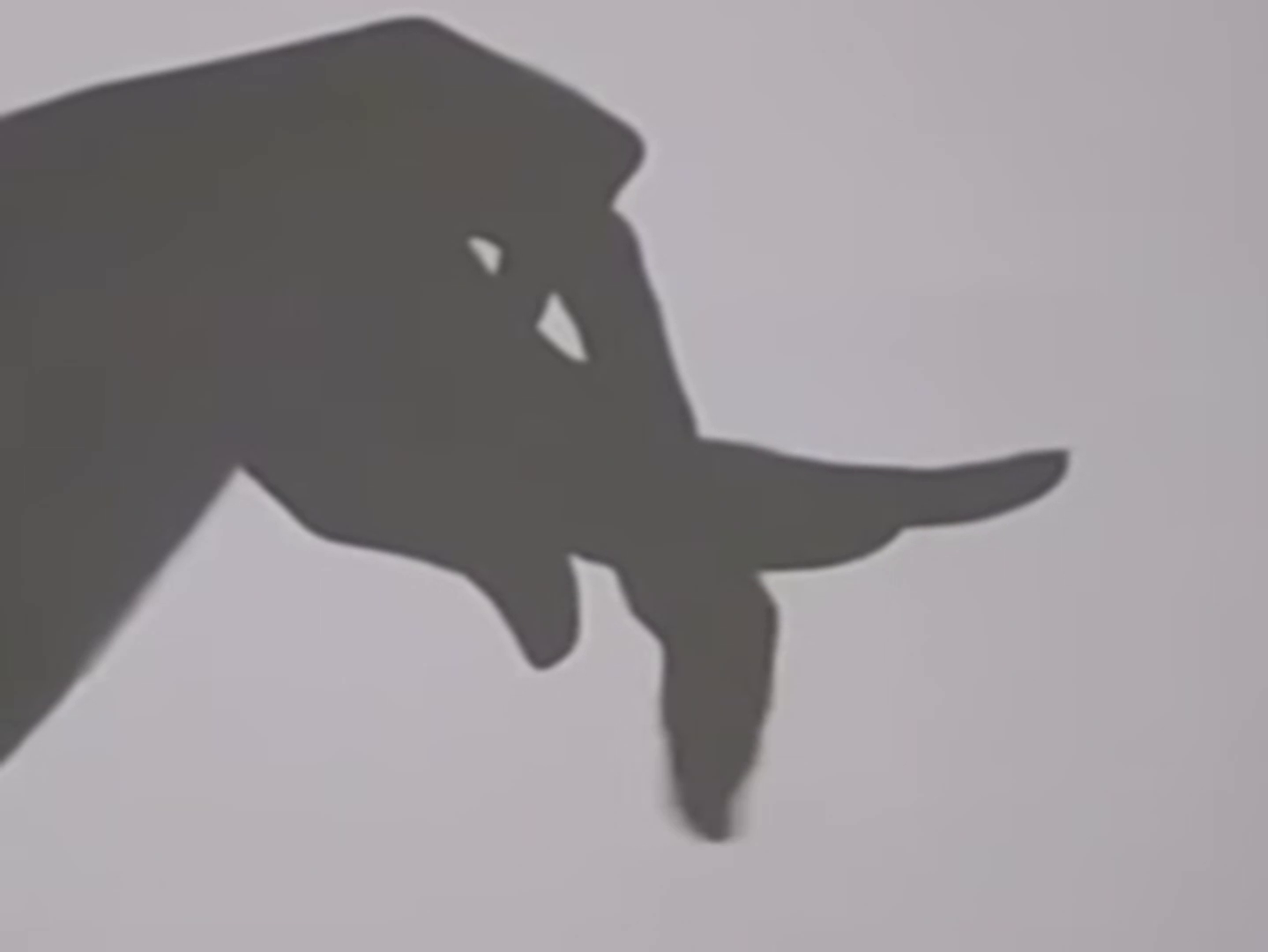}\label{fig:fig3g}}
    \hfil
    \subfloat[Horse]
    {\includegraphics[width=0.19\linewidth]{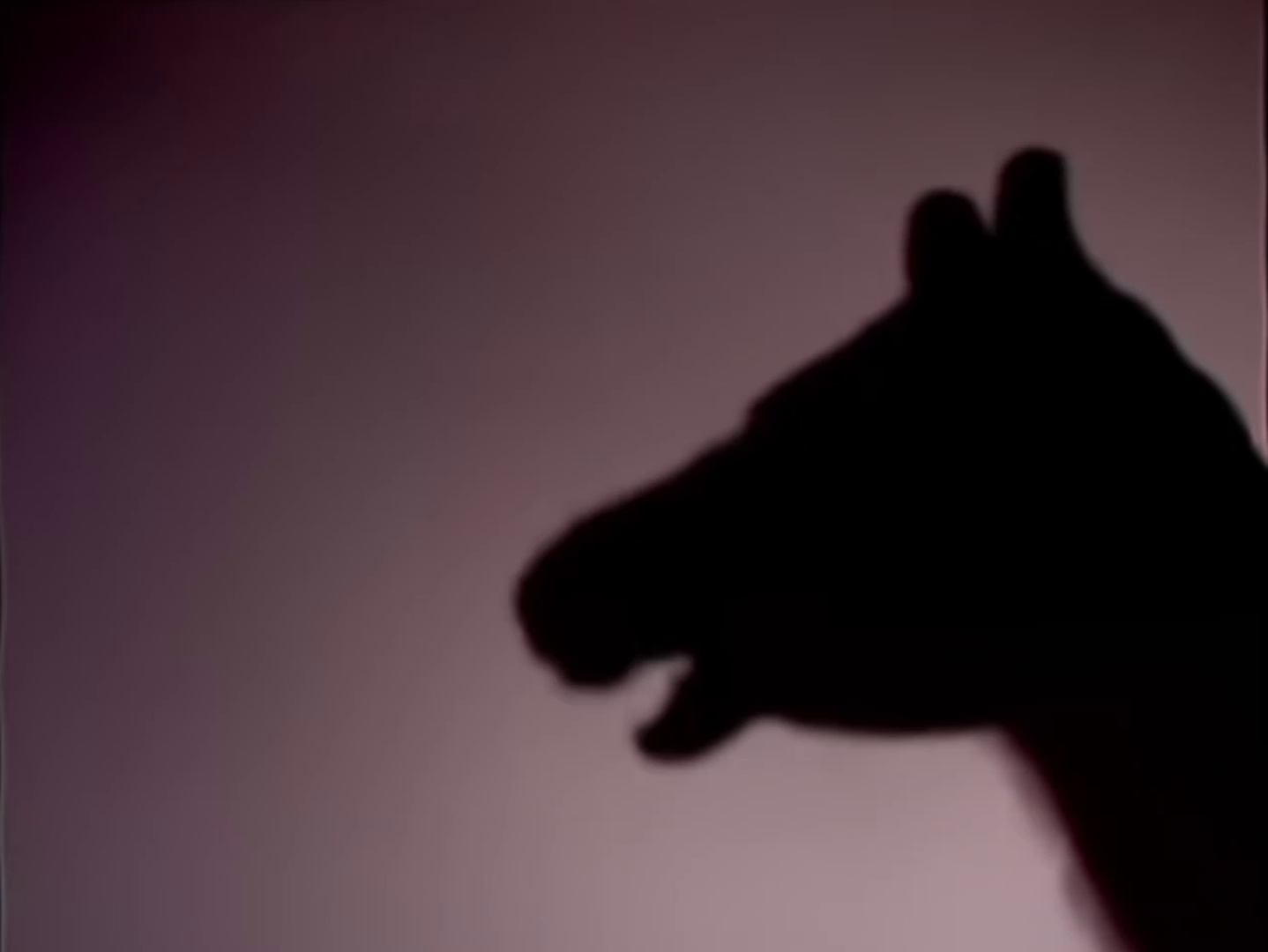}\label{fig:fig3h}}
    \hfil
    \subfloat[Llama]
    {\includegraphics[width=0.19\linewidth]{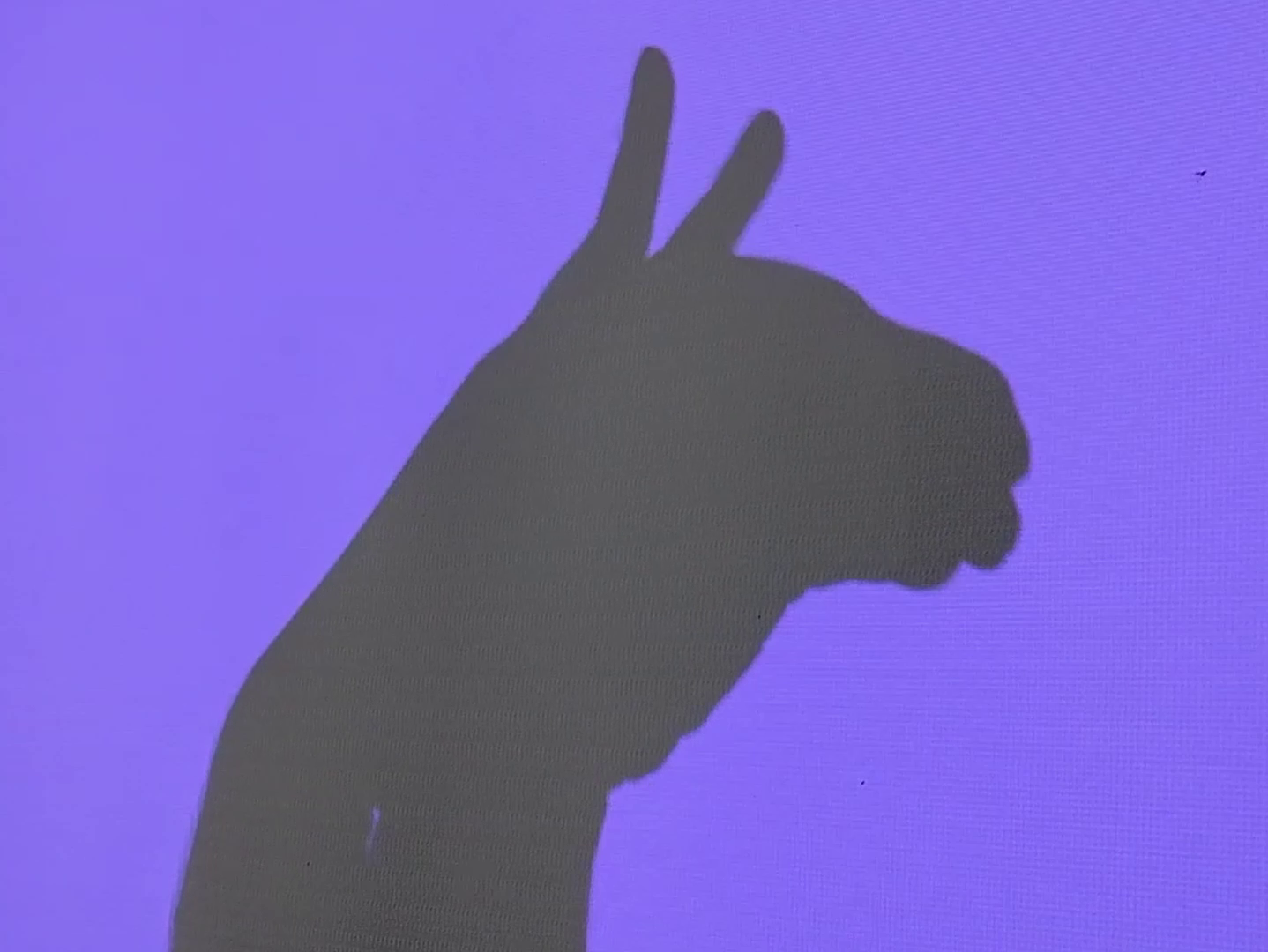}\label{fig:fig3i}}
    \hfil
    \subfloat[Moose]
    {\includegraphics[width=0.19\linewidth]{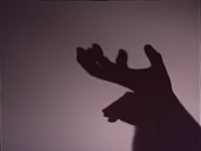}\label{fig:fig3j}}\\
    \vspace{1.1mm}\subfloat[Panther]
    {\includegraphics[width=0.19\linewidth]{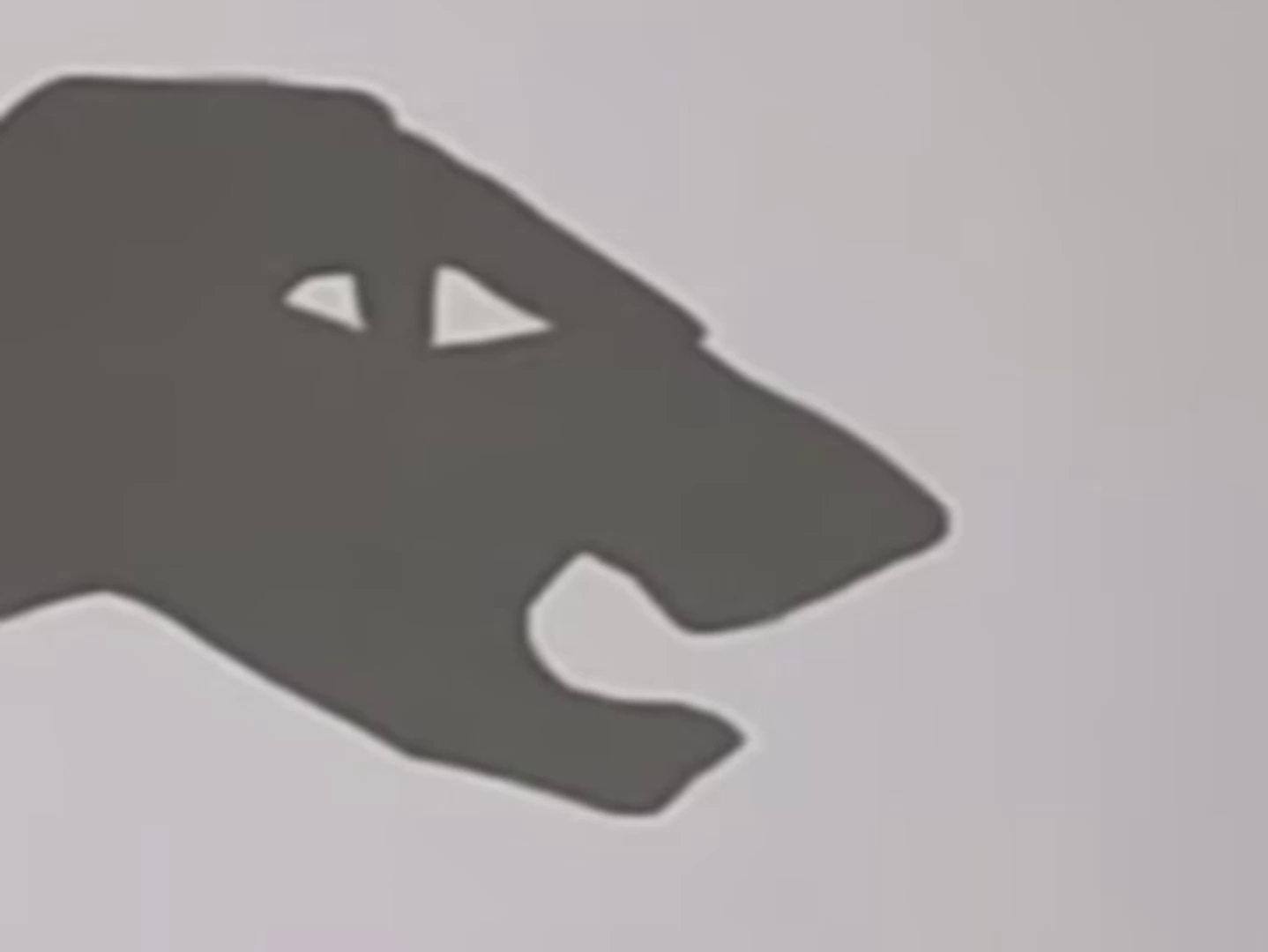}\label{fig:fig3k}}
    \hfil
    \subfloat[Rabbit]
    {\includegraphics[width=0.19\linewidth]{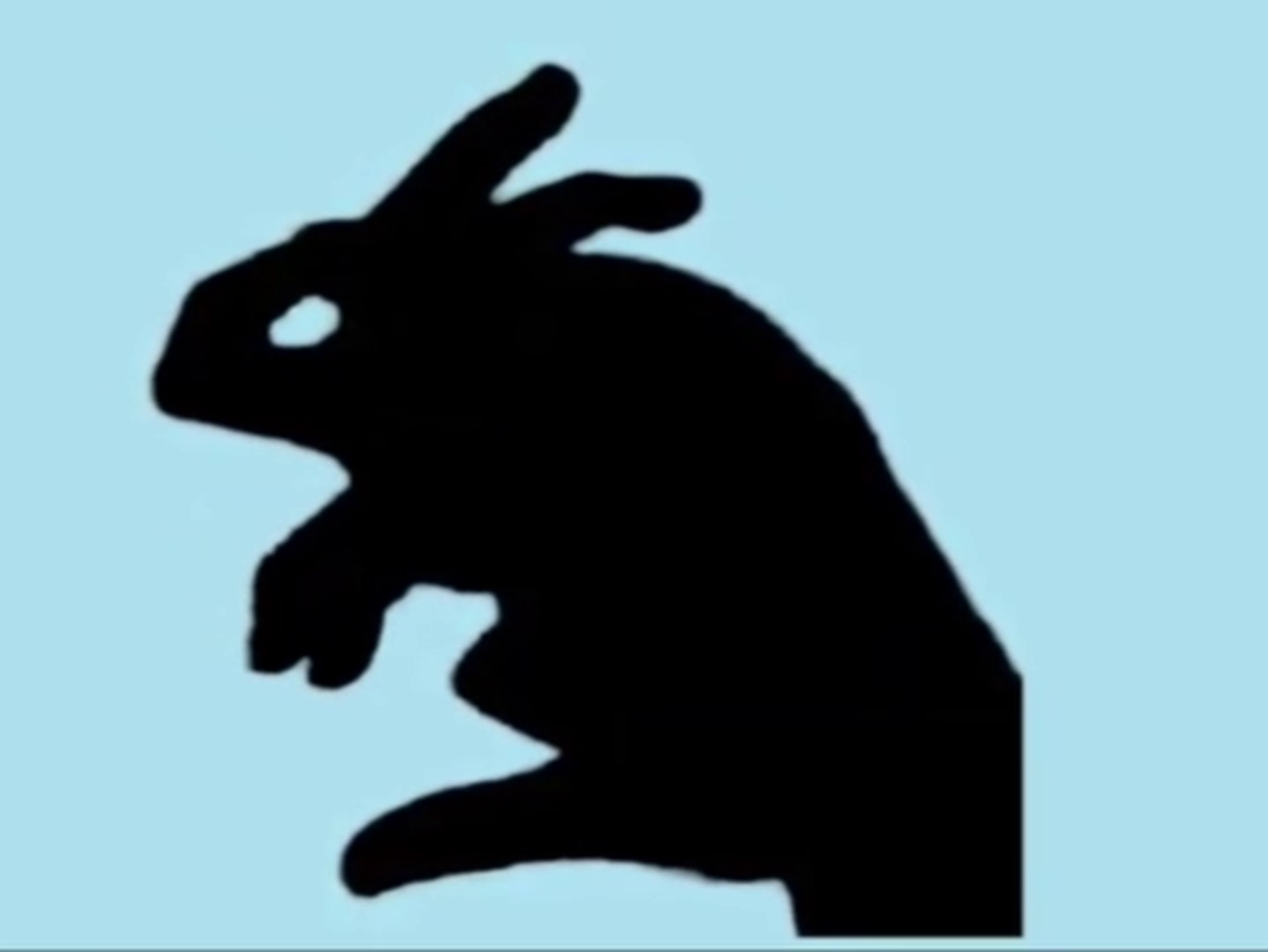}\label{fig:fig3l}}
    \hfil
    \subfloat[Snail]
    {\includegraphics[width=0.19\linewidth]{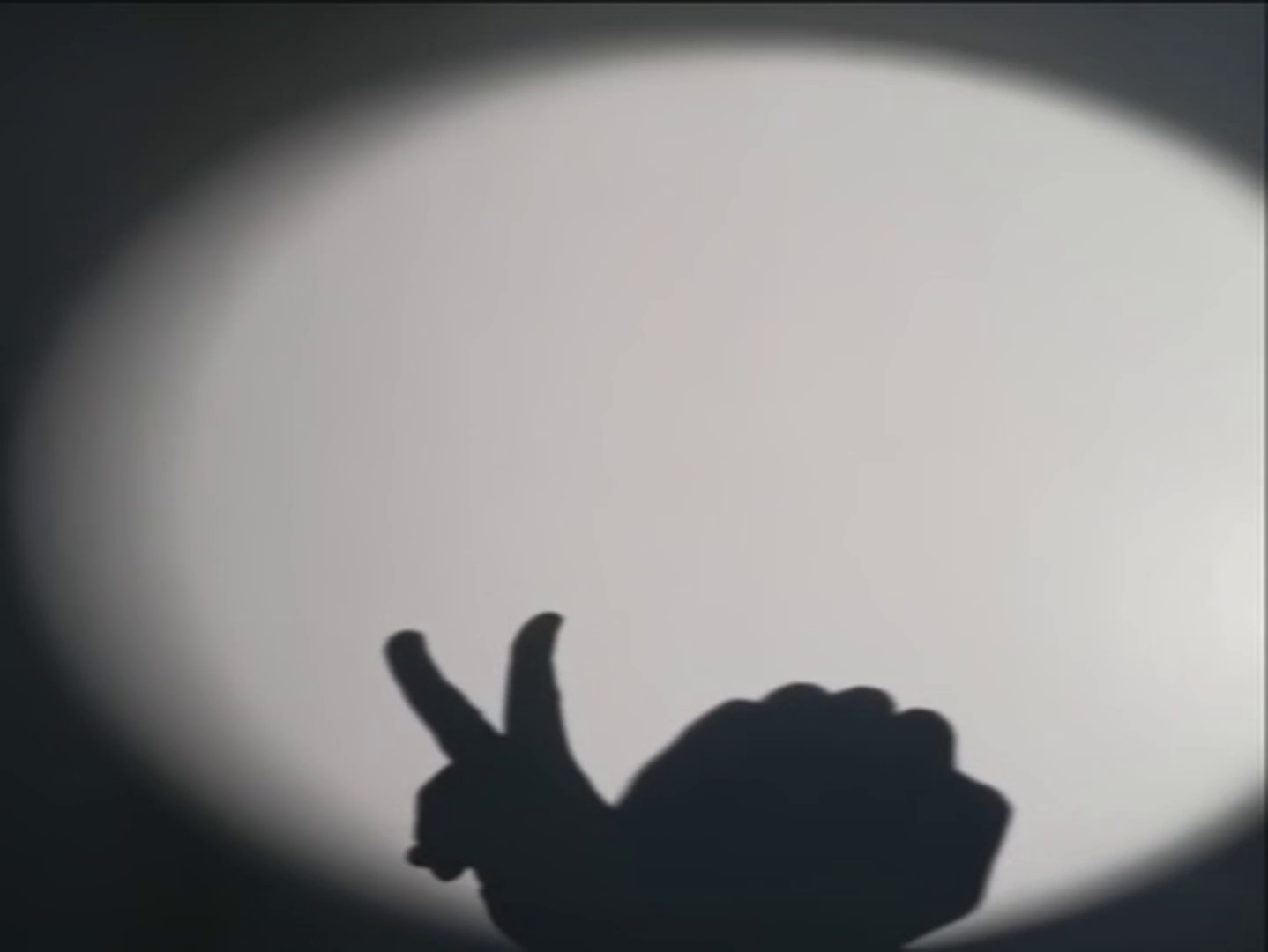}\label{fig:fig3m}}
    \hfil
    \subfloat[Snake]
    {\includegraphics[width=0.19\linewidth]{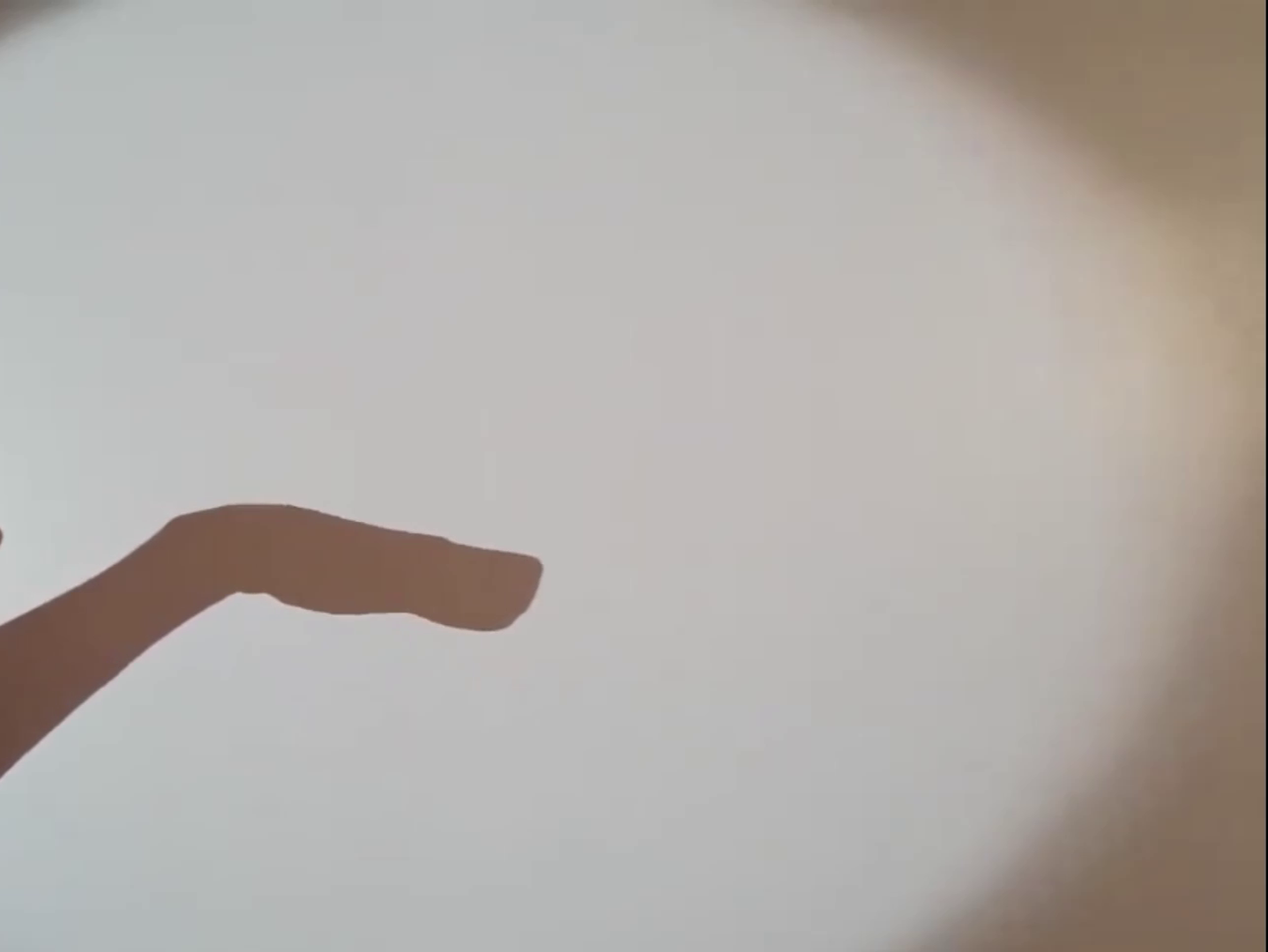}\label{fig:fig3n}}
    \hfil
    \subfloat[Swan]
    {\includegraphics[width=0.19\linewidth]{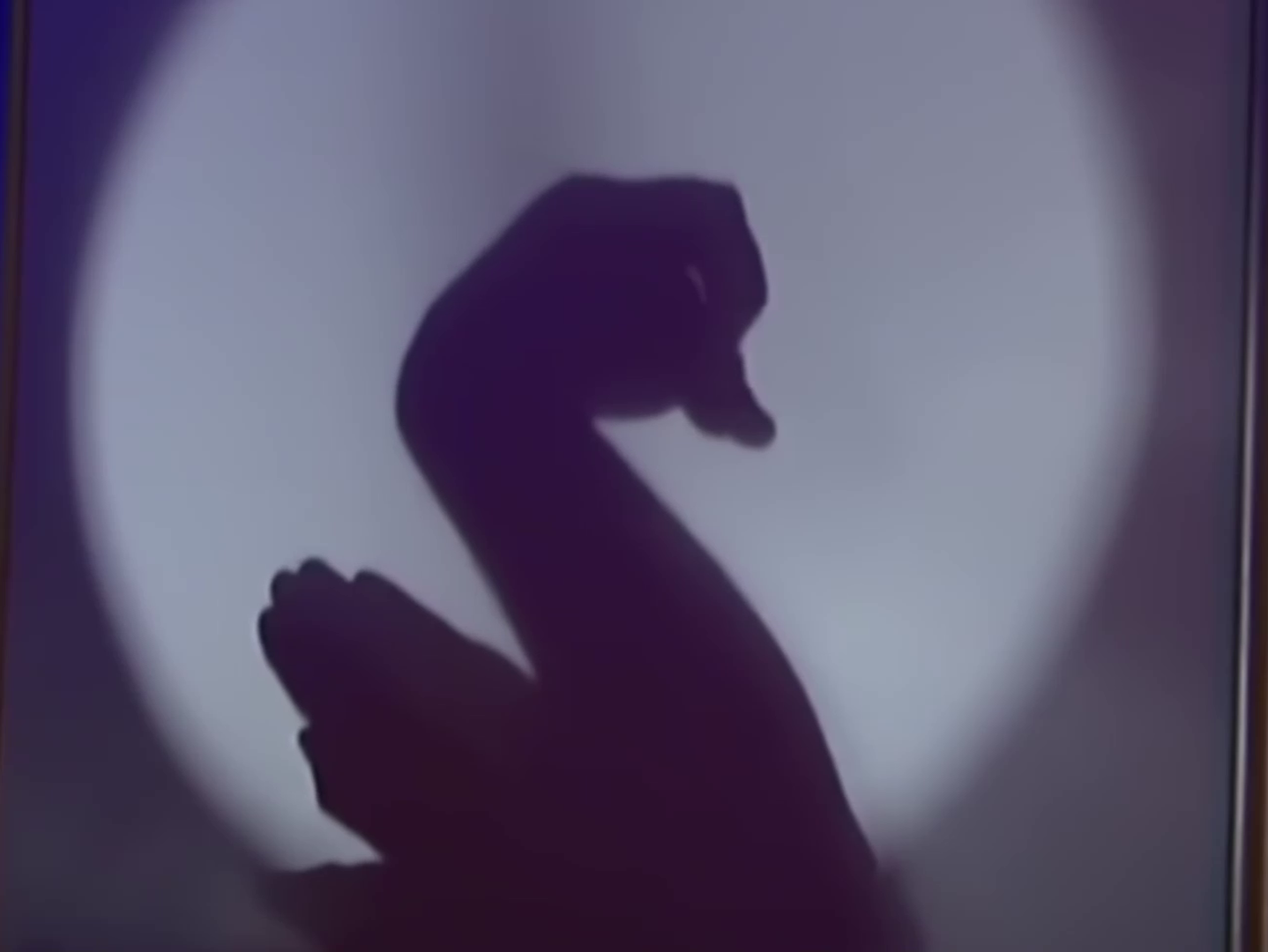}\label{fig:fig3o}}
    \caption{Samples from each class of the dataset.}
    \label{fig:fig3}
    \end{figure*}
\subsection{Extracting Samples}
\begin{figure}[t]
    \centering
      \subfloat[\centering $t$th frame]{\includegraphics[width=0.31\linewidth]{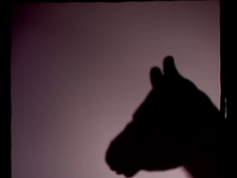}\label{fig:fig_ofa}}
      \hspace{0.0cm}
      \subfloat[\centering $(t+k)$th frame]{\includegraphics[width=0.31\linewidth]{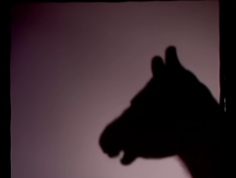}\label{fig:fig_ofb}}
      \\
      \vspace{1.1mm}
      \subfloat[\centering LK method]{\includegraphics[width=0.31\linewidth]{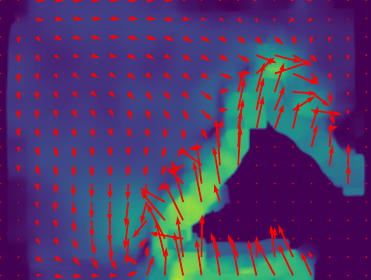}\label{fig:fig_ofc}}
      \hspace{0.0cm}
      \subfloat[\centering TV-$L^1$ method]{\includegraphics[width=0.31\linewidth]{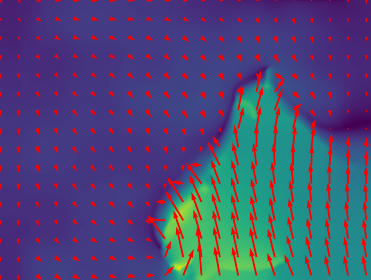}\label{fig:fig_ofd}}
      \hspace{0.0cm}
      \subfloat[\centering {\footnotesize $\max(\text{LK}, \text{TV-}L^1)$}]{\includegraphics[width=0.31\linewidth]{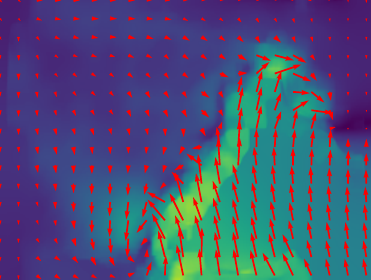}\label{fig:fig_ofe}}
    \caption{Optical flow estimation of contiguous candidate frames from the `Horse' class.}
    \label{fig:fig_of}
  \end{figure}

To mitigate the presence of excessively similar and redundant image samples, we extract frames from these clips at reasonable intervals of $k$ after downsampling the clips to a resolution of $1438 \times 1080$. The values of $k$ are judiciously chosen for the clips of each class, and every $k$th frame is selected as a candidate image sample (\textit{e.g.}, with $k \approx 180$, $200$, $220$ for a 60 FPS clip). From this sequence of extracted candidate frames, we prioritize those exhibiting significant motion. To this end, we estimate the motion vector field by calculating the optical flow between the consecutive $t$th frame and the $(t+k)$th frame. The magnitude of this motion is quantified by the mean L2 norm of the resulting flow field. We retain the frame pairs with an average flow magnitude surpassing a certain requisite threshold $\tau$, thereby ensuring the inclusion of dynamically distinct frames. We synergistically amalgamate two optical flow estimation methods: the Lucas--Kanade (LK) method \cite{lucas1981iterative} and the Total Variation L1 Regularization (TV-$L^1$) method \cite{zachtvl1}. The undergirding assumption beneath the LK method is brightness constancy and spatial coherence of the flow in a local neighbourhood of the pixel (say, the patch $W$) under consideration. It employs a multi-scale gradient descent optimization approach for the constraint equation shown in Equation \ref{eq:eq1}.
{\small
\begin{align}
  \label{eq:eq1}
  I_x \cdot u + I_y \cdot v + I_t = 0
\end{align}
}
where, $I_x = \frac{\partial I}{\partial x}$ and $I_y = \frac{\partial I}{\partial y}$ are the spatial gradients of the image intensity $I$, and $I_t = \frac{\partial I}{\partial t}$ is the temporal gradient. $u$ and $v$ are the horizontal and vertical components of the optical flow vector, respectively. The motion is then estimated by iteratively minimizing the cost function in Equation \ref{eq:eq2} at increasingly granular image resolutions, from coarse to fine.
{\small
\begin{align}
  \label{eq:eq2}
  \begin{bmatrix}
    u \\ v
  \end{bmatrix}_{\text{LK}} = \argmin_{u,v}\sum_{\{i,j\} \in W} \left[I_x(i,j) \cdot u_{ij} + I_y(i,j) \cdot v_{ij} + I_t(i,j)\right]^2
\end{align}
}
which can be then solved with the closed-form solution in Equation \ref{eq:eq3}.
{\small
\begin{align}
  \label{eq:eq3}
  \begin{bmatrix}
    u \\ v
  \end{bmatrix}_{\text{LK}} = \begin{bmatrix}
    \sum_{\{i,j\} \in W} I_x^2 & \sum_{\{i,j\} \in W} I_x I_y \\
    \sum_{\{i,j\} \in W} I_x I_y & \sum_{\{i,j\} \in W} I_y^2
  \end{bmatrix}^{-1} \begin{bmatrix}
    -\sum_{\{i,j\} \in W} I_x I_t \\
    -\sum_{\{i,j\} \in W} I_y I_t
  \end{bmatrix}
\end{align}
}
As evident from Figure \ref{fig:fig_ofc}, the LK method can track the edge movement of the homogeneous silhouette patches, given the slight texture offered by the penumbral region of the shadow. However, as it is limited by local window constraints, it fails to capture the global motion of the shadow puppet. To ameliorate this issue, we resort to the TV-$L^1$ method, which is a variational method that minimizes the total variation of the flow field, subject to the L1 norm of the data fidelity term, which together form the energy function $E$ in Equation \ref{eq:eq4}.
{\footnotesize
\begin{align}
  \label{eq:eq4}
  \begin{bmatrix}
    u \\ v
  \end{bmatrix}_{\text{TV-}L^1} &= \argmin_{u,v} E(u,v)\\
  &= \argmin_{u,v} \int_{\Omega} (\lambda \underbrace{\left\lVert \nabla I \cdot \vv{w} + I_t \right\rVert_1}_{\text{Data term}} + \underbrace{\left\lVert \nabla u \right\rVert_1 + \left\lVert \nabla v \right\rVert_1}_{\text{L1 Regularization term}}) \text{\,d}x \text{\,d}y
\end{align}
}
where $\vv{w} = \langle u,v\rangle$ is the optical flow vector, $\nabla u$ and $\nabla v$ are the spatial gradients of the flow, $\lambda$ is the parameter for balancing data fidelity and regularization, and $\Omega \subseteq \mathbb{R}^2$ represents the spatial domain of the entire image. The TV-$L^1$ method is more adept at capturing the global motion of the homogeneous and spatially consistent inner portion of the shadow puppet, as depicted in Figure \ref{fig:fig_ofd}. We then take the element-wise maximum of the LK and TV-$L^1$ flow fields' L2 norms to obtain a more hoilistic optical flow field, as portrayed in Figure \ref{fig:fig_ofe}.
\begin{align}
  \label{eq:eq5}
  \begin{bmatrix}
    u \\ v
  \end{bmatrix}^* &= \argmax_{u,v} \left( \left\lVert \begin{bmatrix}
    u \\ v
  \end{bmatrix}_{\text{LK}} \right\rVert_2, \left\lVert \begin{bmatrix}
    u \\ v
  \end{bmatrix}_{\text{TV-}L^1} \right\rVert_2 \right)
\end{align}
Then from this maximum combination flow field of resolution $M \times N$, we compute the average L2 norm $\overline{V}$ using Equation \ref{eq:eq6}.
\begin{align}
  \label{eq:eq6}
  \overline{V} = \frac{\displaystyle\sum_{i=0}^{M-1} \displaystyle\sum_{j=0}^{N-1} \sqrt{u_{ij}^{*\;2} + v_{ij}^{*\;2}}}{MN}
\end{align}
If $\overline{V} > \tau$, we retain the corresponding frame pair as candidate samples, otherwise we continue the process with the $(t+k)$th and the $(t+2k)$th frames. Table \ref{tab:tab1} encapsulates some essential statistical information related to \haspers and provides a superficial overview of the dataset.


\subsection{Labeling}
After the extraction of the frames, the samples undergo manual scrutiny by 3 annotators who are pursuing undergraduate studies in Computer Science and Engineering (CSE). If a series of contiguous samples \textit{prima facie} exhibit substantial similarity, we only keep a single image from that set of samples. The rest are discarded to avoid redundancy and to instill diversity. Another criterion that dictates the legitimacy of an image sample is its intelligibility. If the majority of the annotators agree on the unintelligibility of a sample, they discard it in unison. After performing this omission of unsuitable samples for each class, we end up with 15 different directories of images, each containing the curated samples of a particular class. The images in these folders are then further partitioned into training and validation sets, maintaining a 60:40 split.
We also pragmatically incorporate a proper distribution of the samples sourced from amateur clips over both the training and validation sets, to avoid making the latter unfairly difficult for the classification models.

\section{Dataset Description}
To provide a tangible exposition of the diverse samples in the dataset, Figure \ref{fig:fig3} presents a collection of representative images across all 15 classes. With minimally astute perspicacity, we can observe that the samples vary in terms of the nature of the backgrounds, the anatomical structure of the puppeteers' hands, the photometric opacity and sharpness of the projected silhouettes, and a panoply of other aspects. 

\vspace{-2mm}
\subsection{Background Variance}
\begin{figure}[t]
    \centering
    \includegraphics[width=\linewidth]{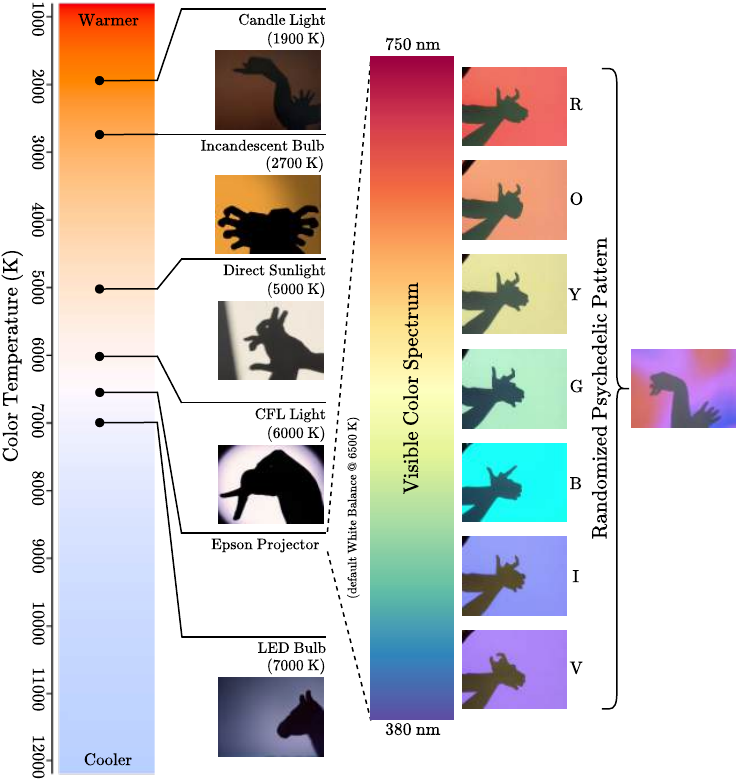}
    \setlength{\belowcaptionskip}{-5mm}
    \caption{Light sources for background diversity in \hasper.}
    \label{fig:fig_lighting}
\end{figure}

The hand shadow puppetry setup that a puppeteer's crew arranges before the performance greatly dictates the nature of the background on which the shadow puppets are displayed. If the location of the light source is very near to the wall or the translucent screen, then we can observe an elliptical shadow contour on the background as evident in Figures \ref{fig:fig3a} and \ref{fig:fig3m}. The angular directionality of the light also manifests a gradient effect on the background as can be seen in Figures \ref{fig:fig3d} and \ref{fig:fig3h}. The temperature and color of the light emanated by the light sources onto the screens also add to the diversity. To achieve this, we use six different light sources---candlelight, an incandescent bulb, sunlight, a CFL light, an Epson EB-972 XGA projector, and an LED bulb---each with different color temperatures. Historically, many other light sources were used by shadowgraphists such as marrow-fat lamps, flame torches, halogen lamps, lime lights, etc. Figure \ref{fig:fig_lighting} depicts an overarching illustration of the monocular polychromatic background lighting diversity that we maintain in \hasper. We avail the overhead projector to emit light from across the visible spectral range ($380$--$750$ nm). Additionally, we use a random-patterned combination of these colors (colloquially referred to as the \textit{psychedelic pattern}) as the backdrop for a subset of the organically created samples.
\vspace{-2mm}

\subsection{Nature of the Silhouettes}
\begin{figure}[t]
    \centering
    \subfloat[Sharp, high opacity]{\includegraphics[width=0.48\linewidth]{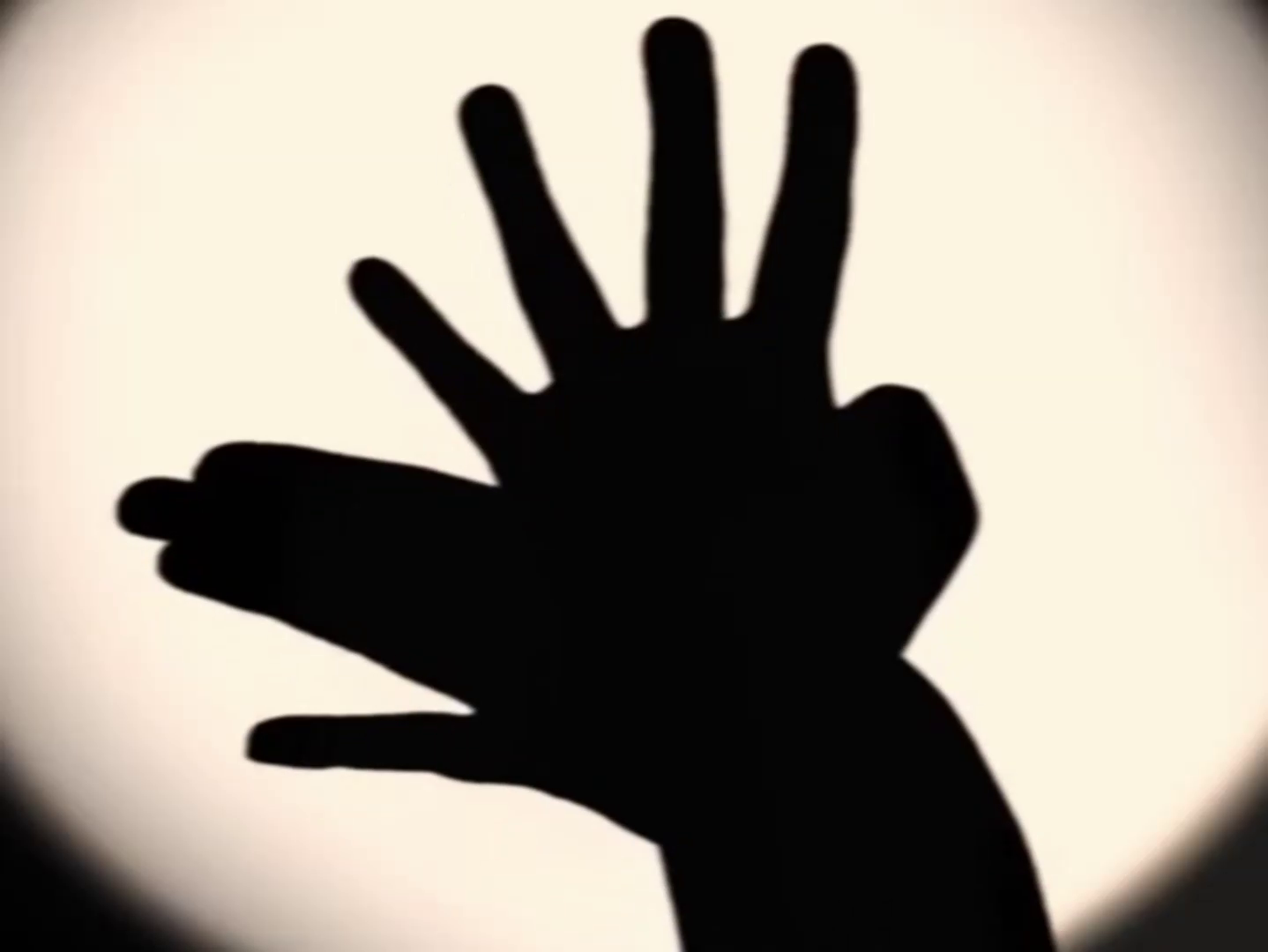}\label{fig:fig4a}}\hspace{0.2cm}\subfloat[Diffuse, low opacity]{\includegraphics[width=0.48\linewidth]{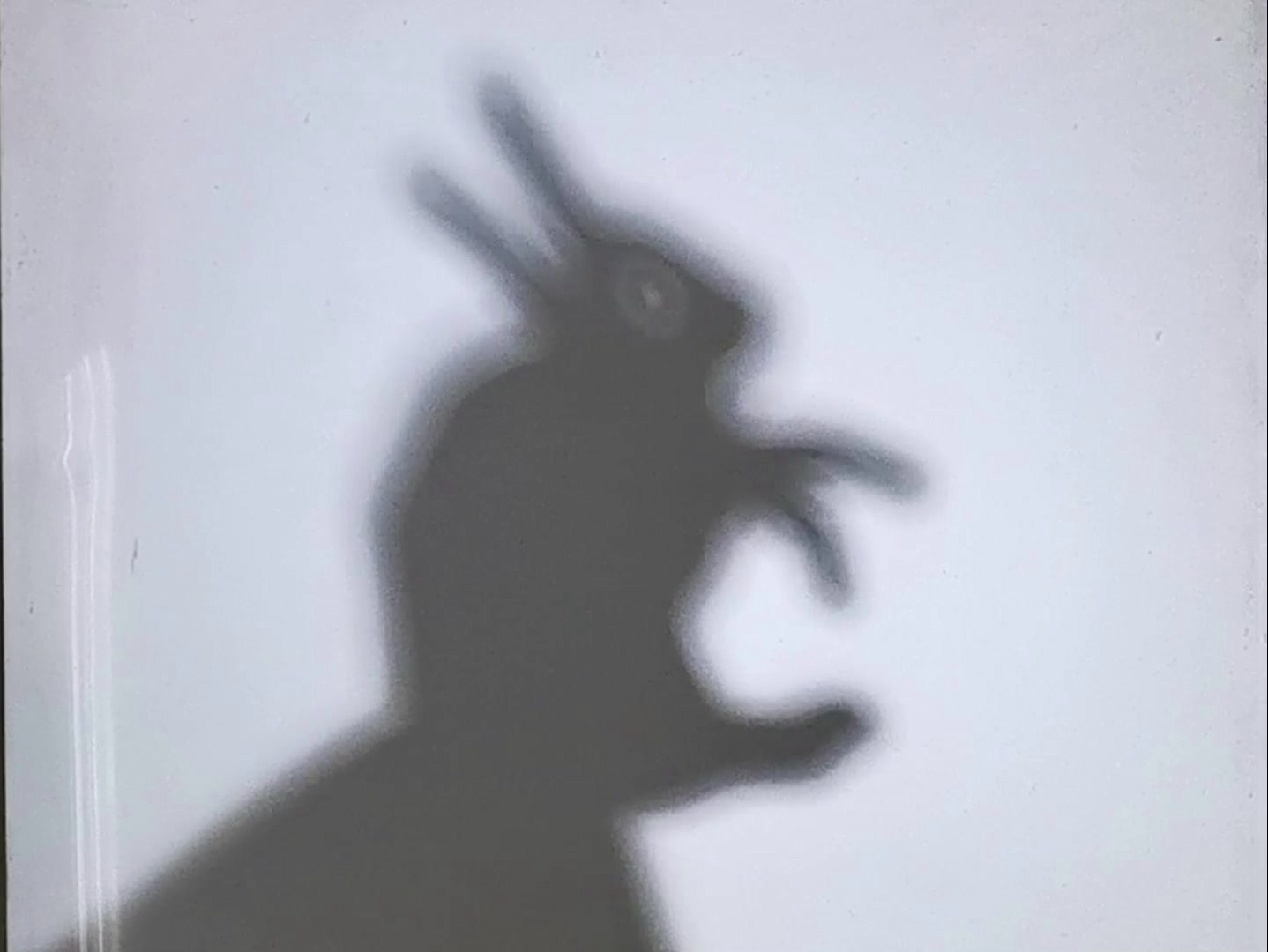}\label{fig:fig4b}}
    \caption{Samples with different silhouette properties.}
    \label{fig:fig4}
\end{figure}

The positioning of the light source with respect to the puppeteer's hands plays a role in shaping the shadows' quality. As per the natural laws of optics, proximity to the light source yields crisp, well-defined shadows (\textit{e.g.}, Figure \ref{fig:fig4a}), while increasing the distance fosters softer, more diffuse shadows (\textit{e.g.}, Figure \ref{fig:fig4b}) with a central umbra and peripheral penumbra. The higher the contrast between the silhouettes and their respective backdrops, the more visible and well-contoured the shadow puppets are. The direction of the light source influences the orientation and shape of the shadows. Shadows cast by overhead lighting sources may appear elongated, while shadows cast by low-angle lighting sources may exhibit softer edges and less pronounced contrast and sharpness. Similarly, due to varied values of the lateral incident angle at which the light sources are kept relative to the screen's normal, we see horizontally elongated and compressed shadows. The shadows also differ in terms of the magnitude of their opacity, \textit{i.e.}, the degree to which the hands prevent the transmission of light being projected onto the screen.

\vspace{-2mm}
\subsection{Hand Anatomy and Stylistic Flair of the Puppeteers}
\begin{figure}[t]
    \centering
    \subfloat{\includegraphics[scale=0.055]{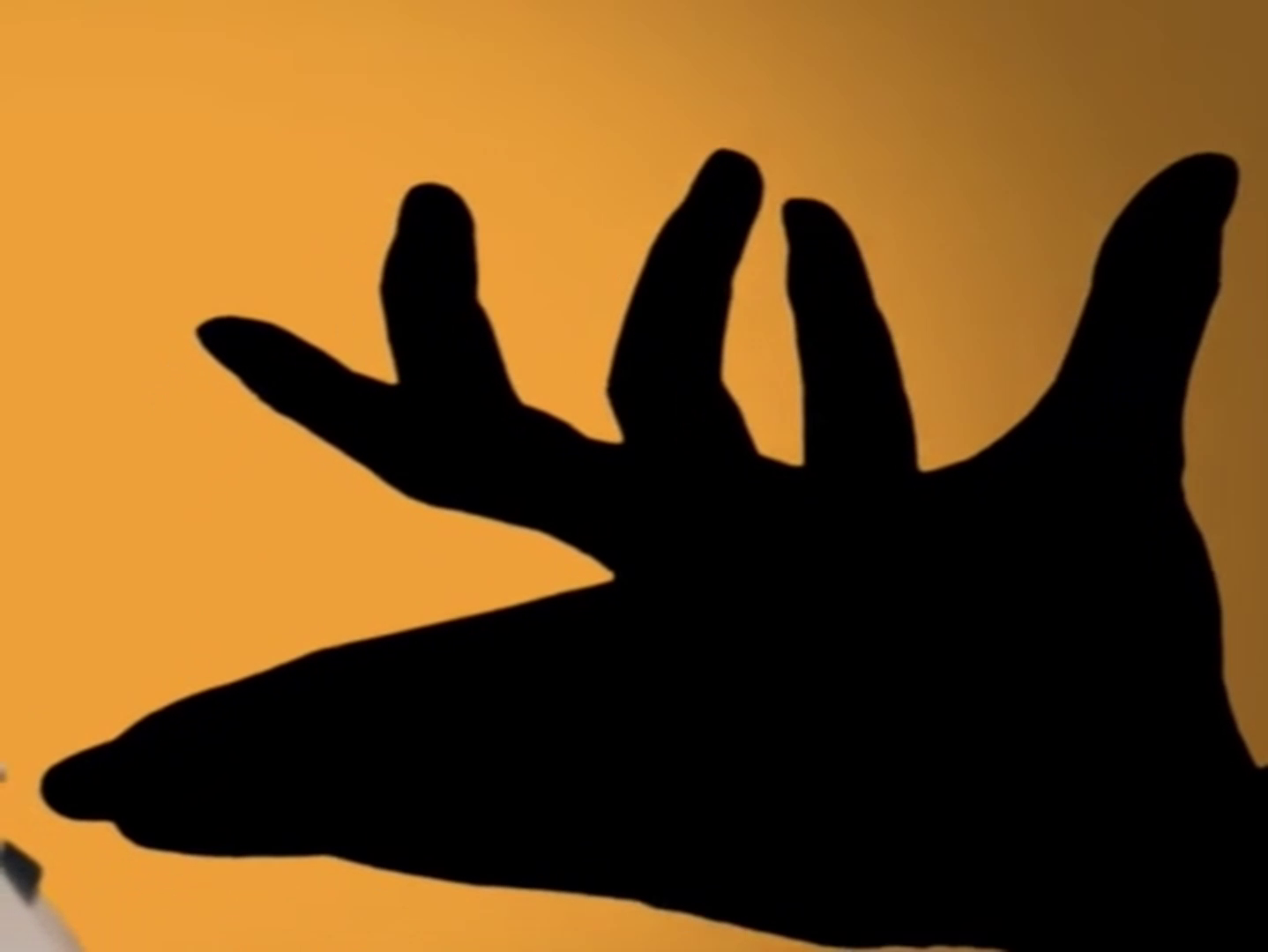}\label{fig:fig5a}}\hspace{0.15cm}\subfloat{\includegraphics[scale=0.055]{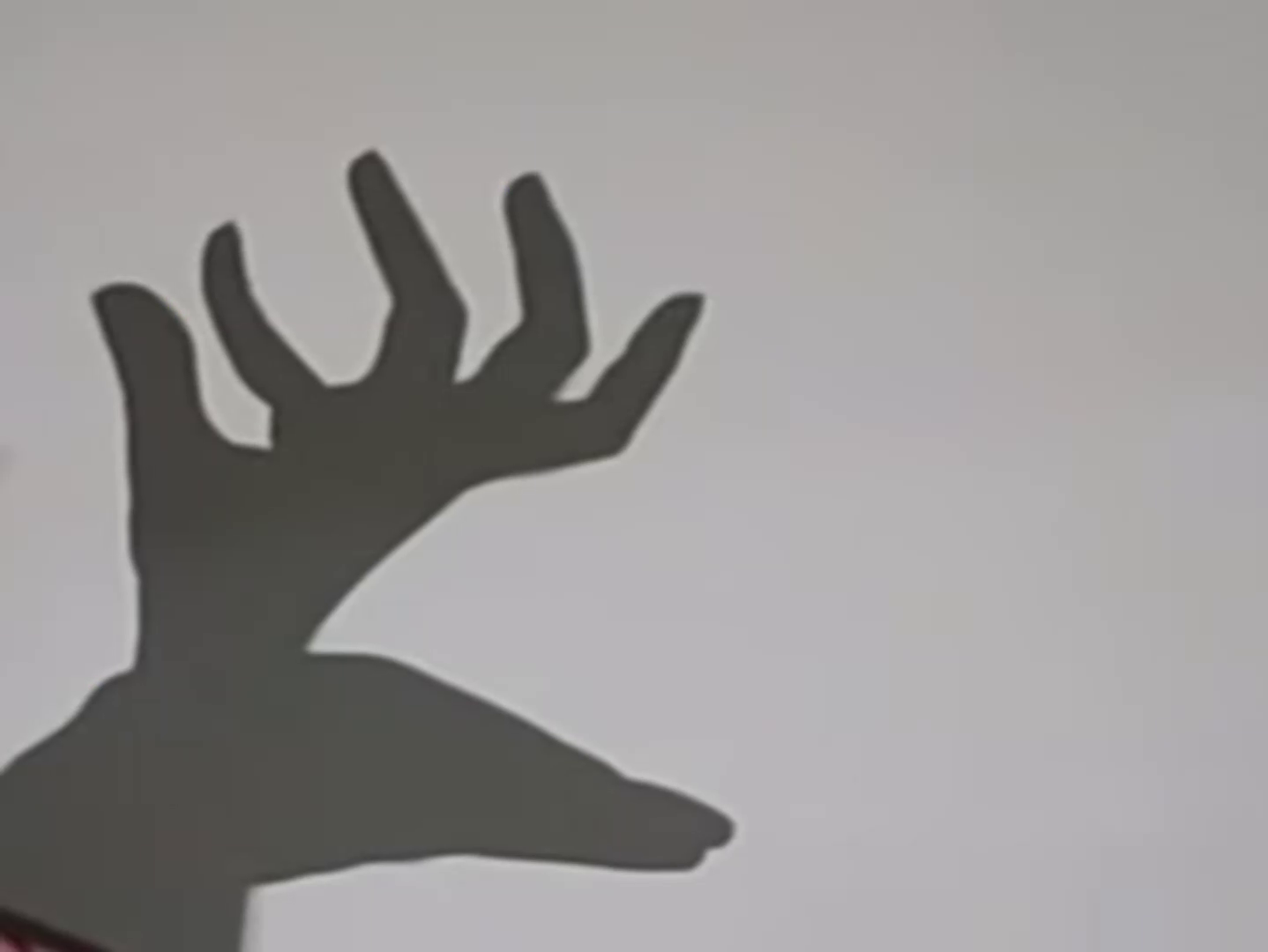}\label{fig:fig5b}}
    \hspace{0.15cm}\subfloat{\includegraphics[scale=0.055]{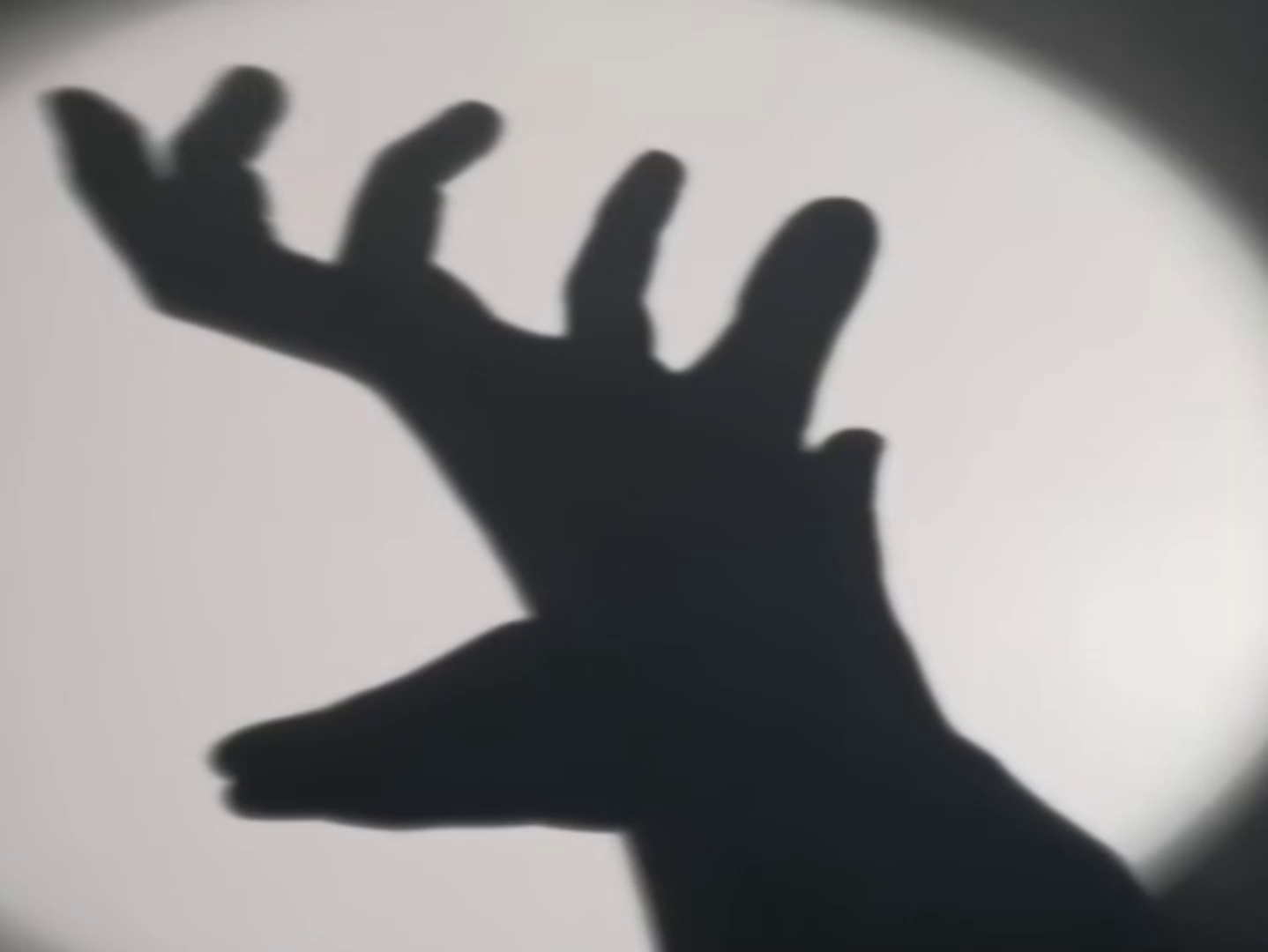}\label{fig:fig5c}}\\\vspace{1.1mm}
    \subfloat{\includegraphics[scale=0.055]{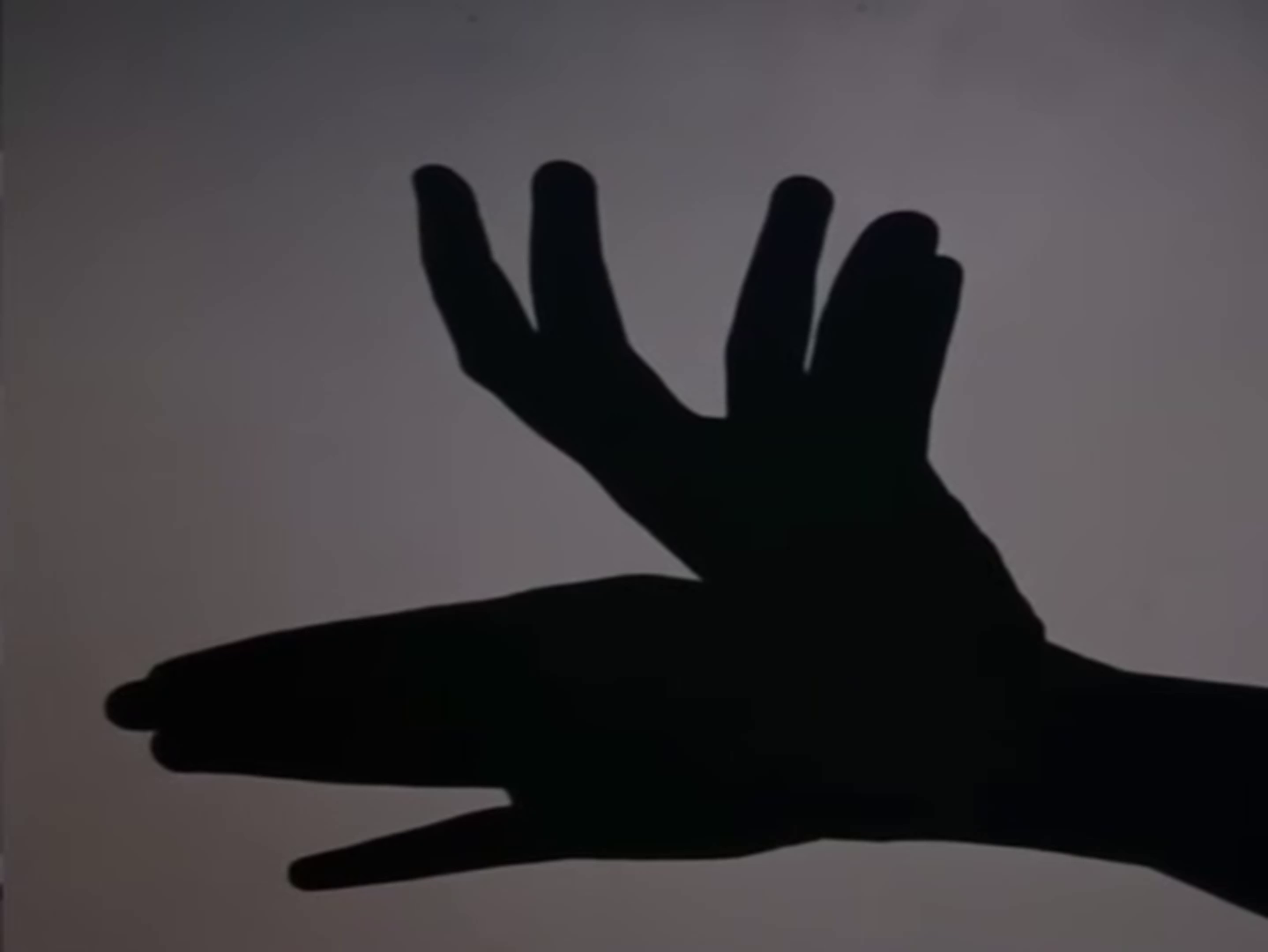}\label{fig:fig5d}}
    \hspace{0.15cm}\subfloat{\includegraphics[scale=0.07333]{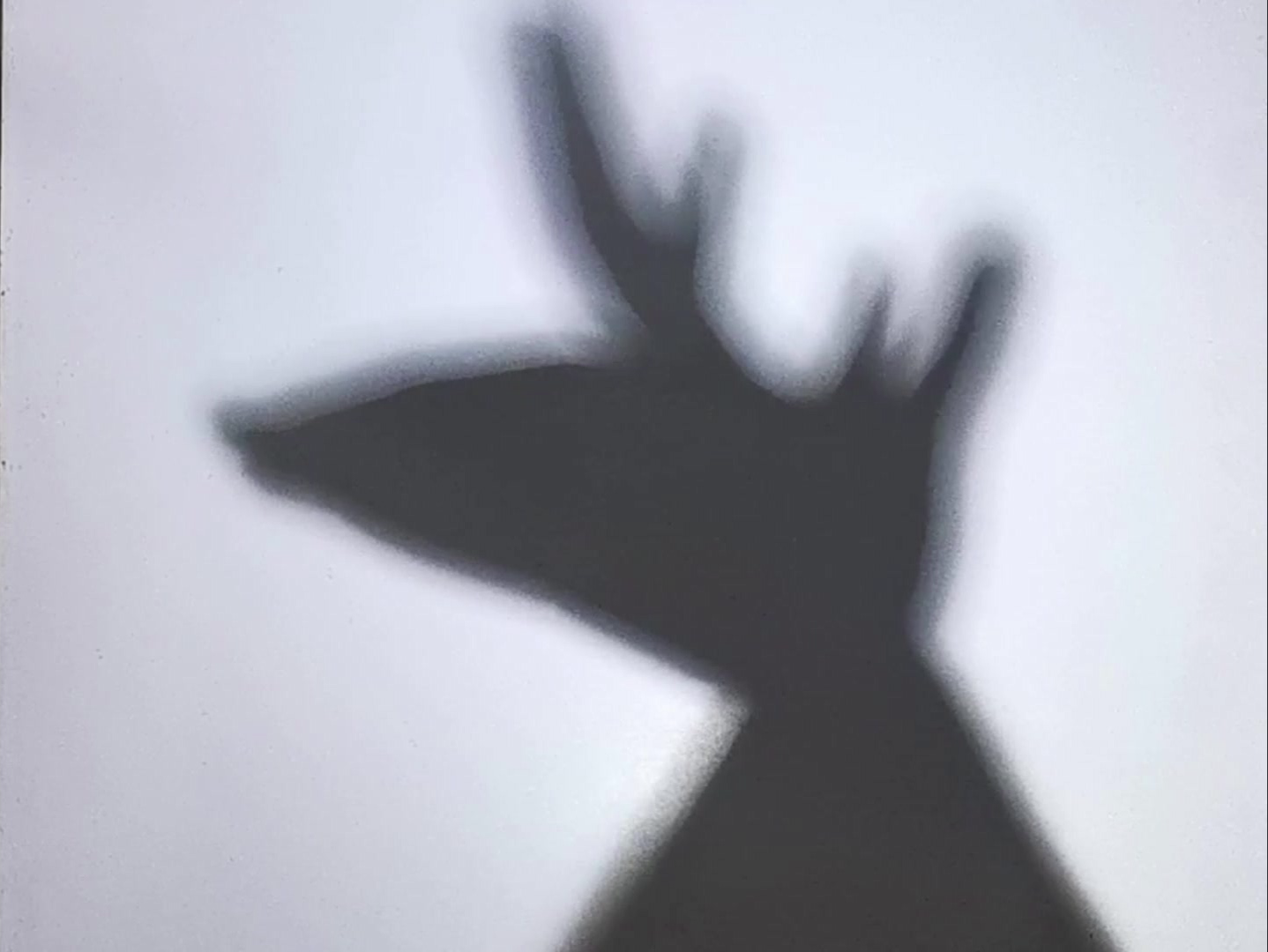}\label{fig:fig5e}}\hspace{0.15cm}\subfloat{\includegraphics[scale=0.07333]{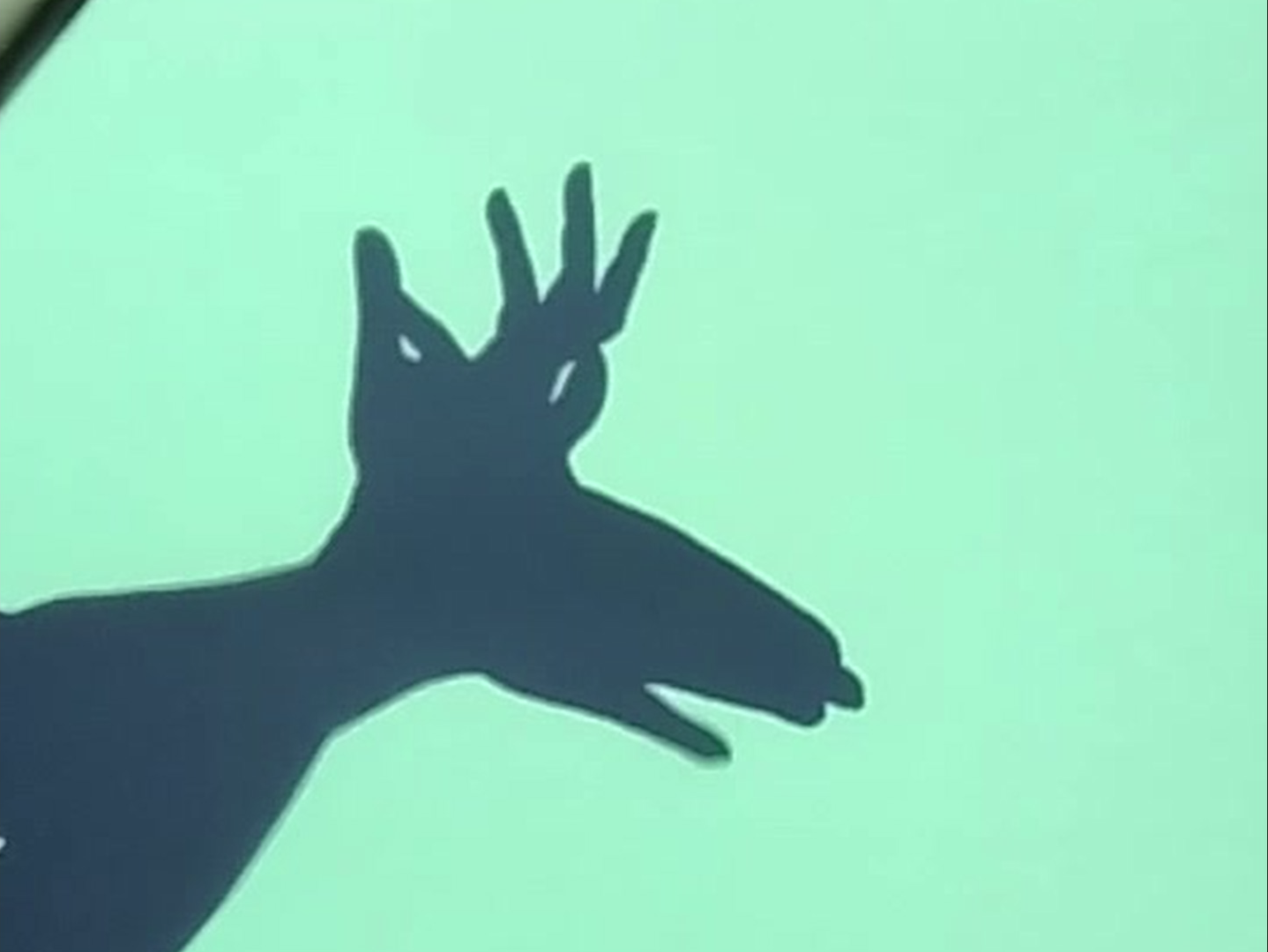}\label{fig:fig5f}}
    \caption{`Deer' samples with different artistic representations.}
    \label{fig:fig5}
\end{figure}

The physiological properties of the puppeteers' hands can vary significantly due to a combination of genetic factors, environmental influences, and lifestyle choices. These nuanced anatomical variations of the wrists, palms, and digits of the puppeteers, along with the different stylistic choices they employ in their choreography, contribute as yet another avenue of diversity of the image samples in \hasper. Human beings, by nature, exhibit morphometric variations in finger length, palm width, and forearm thickness based on age and gender. As such, the cohort of novice shadowgraphists ($n=6$) that we employ for the creation of amateur samples comprises a balanced gender representation of 3 males and 3 females, spanning an age range from 9 to 25 years. Among the adults, the hand anthropometric measurements are of $18.75\pm 1.55$ cm in length and $8.66\pm 0.77$ cm in width. For the minors, who obviously have proportionally smaller hand dimensions, these measurements are $14.23\pm 1.16$ cm and $6.73\pm 0.82$ cm respectively. The gender representation among the 14 professional shadow puppeteers is however imbalanced, with 12 male and 2 female shadowgraphists. Figure \ref{fig:fig5} pristinely demonstrates the morphological variations of hand shadow puppets belonging to the `Deer' class due to anatomical and stylistic diversity.
\vspace{-5mm}
\subsection{Comparative Analysis}
\subsubsection{Inter-class Similarity}
Due to the conspicuous resemblance in the anatomical structures of certain animal species, the samples belonging to the classes corresponding to those animals exhibit a notable degree of similarity as well. Figures \ref{fig:fig3e} and \ref{fig:fig3j} are prime examples of such structural similitude that can be observed between the `Deer' and `Moose' classes. These similarities make the image classification task on \haspers quite challenging and culminate to being the reason behind a lot of misclassifications, as discussed in Section \ref{sec:error_analysis}.
\begin{figure*}[t]
    \centering
    \includegraphics[scale=0.84]{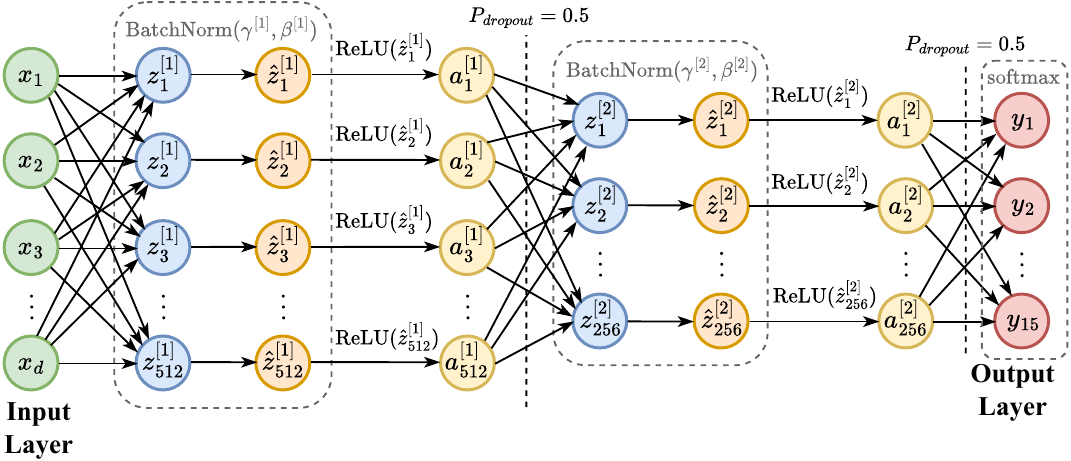}
    \caption{Classifier block attached to the tail-end of the pretrained models. Here, $d$ is the feature dimension of the anterior model. The output features of layer $l$ is $z^{[l]}=W^{[l]}a^{[l-1]}$, where $a^{[l-1]}$ denotes the activation values of the preceding $(l-1)$th layer. The batch normalized value of the $i$th output feature $z^{[l]}_i$ is $\hat{z}^{[l]}_i = \gamma^{[l]}z_{norm}^{[l](i)} + \beta^{[l]}$, where $\gamma$ and $\beta$ are learnable parameters. The activation values of layer $l$ are denoted by $a^{[l]} = g(\hat{z}^{[l]})$ which is computed using the activation function $g = \operatorname{ReLU}$. The predicted probabilities are determined by using the $\operatorname{softmax}$ function on the logits, \textit{i.e.}, $P(\hat{y_i}=1|x) = \operatorname{softmax}(y)_i = \frac{e^{y_i}}{\sum_{j=1}^{15}e^{y_j}}$}
    \vspace{-5mm}
    \label{fig:classifier_block}
\end{figure*}
\subsubsection{Intra-class Dissimilarity}
Some classes include samples of multiple species of the same animal, and these samples are starkly different in appearance from one another. Given the presence of such quasi-disparate samples, along with the individualistic flair that manifests through the puppeteers' stylistic choices, a particular class may show a lot of intra-class dissimilarity. As aforementioned, Figure \ref{fig:fig5} portrays the heterogeneity of this nature among the samples from the `Deer' class.

%% file: sections/4_Statistical_Analysis.tex
\vspace{-2mm}

\subsection{Statistical Analysis}
\label{sec:statistical_analysis}
Table \ref{tab:tab1} presents the statistical properties of the \haspers dataset. It tabulates the proportion of samples belonging to each of the 15 classes and their corresponding training-validation splits.
While searching for hand shadow puppetry performances, we anecdotally observe that certain classes of puppets are more popular than others. Irrespective of this fact, we make sure all 15 classes of puppets are equitably represented in our dataset, with each class having 1,000 image samples ($\approx 6.\dot{6}\%$). As evident in
Table \ref{tab:tab1}, the image samples are evenly distributed across all 15 classes.
The proportions of professionally sourced samples belonging to the `Llama' and `Snake' classes ($14\%$ and $27.3\%$ respectively) are slightly low due to a scarcity of performance clips starring hand shadow puppets of these classes. For cases such as these, we supplement the classes with samples organically created by our novice shadowgraphist cohort. Each class has $\approx 47.827 \pm 1.414\%$ samples from clips of professional performers, and the rest $\approx 52.172 \pm 1.414\%$ samples are sourced from amateur clips. Of the 15,000 samples, approximately $76.53\%$ feature male hands, while $23.64\%$ represent female hands. In tandem, considering the target demographic of ombromanie, around $28.33\%$ of the samples in \haspers consist of children's hands. In totality, we end up with 9,000 samples in the training set and 6,000 samples in the validation set, thereby partitioning \haspers by maintaining a 60:40 ratio.

%% file: sections/5_Developing_Benchmark.tex
\vspace{-2mm}
\section{Methodology for Benchmarking \hasper}
A series of pretrained models are used as feature extractors to develop a benchmark for the dataset. The models are pretrained on the \textsc{ImageNet} \cite{deng2009imagenet} 
dataset and fine-tuned on \hasper. We implement the training pipeline using the Pytorch\footnote{\url{https://pytorch.org/vision/stable/models.html}} framework.
This section presents an overview of the models, evaluation metrics, and experimental results. 
\vspace{-4mm}
\subsection{Experimental Setup}
\subsubsection{Baseline Models}
For this classification task, we use 31 feature extractor models
as baselines, which are listed in Table \ref{tab:tab2}. Some of these models have a track record of good performance across various other image classification tasks \cite{mauricio2023comparing}. We examine both conventional Convolutional Neural Networks (CNNs) and CNNs augmented with attention mechanisms. Some models have multiple variants in terms of size or number of parameters, and we compare the performance among those variants as well. We fuse silhouette-specific features obtained via topological descriptors \cite{blum1967transformation} and polygonization \cite{douglas1973algorithms} with the extracted features from the best-performing vanilla model (see Appendix \ref{appendix:exp_setup} for more information).

\subsubsection{Performance Metrics}
We use top-$k$ validation accuracy values (with $k=1,2,3$), Precision, Recall, and F1-score as evaluation metrics to perform comparative analyses among the aforementioned models. The latter three judgment criteria are used due to the slightly imbalanced nature of \hasper's professional source clips, as evident in Table \ref{tab:tab1}.
\subsubsection{Classifier Network}
We adopt two approaches to arrive at the final 15-dimensional layer since there are a total of 15 classes to predict from. The first approach is to directly append a 15-dimensional fully connected layer at the tail-end of the vanilla models. The second approach incorporates the classifier block portrayed in Figure \ref{fig:classifier_block}.
\subsubsection{Hyperparameters and Optimizer}
We use Stochastic Gradient Descent (SGD) \cite{kiefer1952stochastic}, with a learning rate $\alpha = 0.001$ and momentum $\gamma = 0.9$, as the optimizing method, and Cross Entropy Loss as the loss metric for all the models. To decay the learning rate, we use Step Scheduler, which decays $\alpha$ by $0.1$ every 5 epochs. 
Each model undergoes training for 50 epochs to ensure equitable comparison, and we empirically ascertain that 50 epochs are sufficient for all of the models to achieve convergence.
\begin{table*}[t]
    \centering
    \caption{Performance comparison of the vanilla and modified versions of the image classification models.}
    \footnotesize
    \setlength{\tabcolsep}{0.46em}
    \def\arraystretch{1.2}
    \resizebox{\linewidth}{!}{
\begin{tabular}{|c|c|cccccccccccc|}
\Xhline{2\arrayrulewidth}
\multirow{4}{*}{\textbf{Models}} & \multirow{4}{*}{\textbf{Params.}} &  \multicolumn{12}{c|}{\textbf{Performance Metrics}}                                                                                                                                                                                                                                                                                                                                                                                                                                                                                                                     \\ \cline{3-14} 
                                 & & \multicolumn{6}{c|}{\textbf{Vanilla}}                                                                                                                                                                                                                                                        & \multicolumn{6}{c|}{\textbf{w/ Classifier Block}}                                                                                                                                                                                                   \\ \cline{3-14} 
                                 & & \multicolumn{3}{c|}{\textbf{Top-$k$ Accuracy (\%)}}                                                               & \multicolumn{1}{c|}{\multirow{2}{*}{\textbf{Precision}}} & \multicolumn{1}{c|}{\multirow{2}{*}{\textbf{Recall}}} & \multicolumn{1}{c|}{\multirow{2}{*}{\textbf{F1-score}}} & \multicolumn{3}{c|}{\textbf{Top-$k$ Accuracy (\%)}}                                                               & \multicolumn{1}{c|}{\multirow{2}{*}{\textbf{Precision}}} & \multicolumn{1}{c|}{\multirow{2}{*}{\textbf{Recall}}} & \multirow{2}{*}{\textbf{F1-score}} \\ \cline{3-5} \cline{9-11}
                                 & & \multicolumn{1}{c|}{\textbf{Top-$1$}} & \multicolumn{1}{c|}{\textbf{Top-$2$}} & \multicolumn{1}{c|}{\textbf{Top-$3$}} & \multicolumn{1}{c|}{}                                    & \multicolumn{1}{c|}{}                                 & \multicolumn{1}{c|}{}                                   & \multicolumn{1}{c|}{\textbf{Top-$1$}} & \multicolumn{1}{c|}{\textbf{Top-$2$}} & \multicolumn{1}{c|}{\textbf{Top-$3$}} & \multicolumn{1}{c|}{}                                    & \multicolumn{1}{c|}{}                                 &                                    \\ \Xhline{2\arrayrulewidth}
\multicolumn{1}{|l|}{\textsc{ShuffleNetV2X10} \cite{ma2018shufflenet}}                 & 2.3M &  \multicolumn{1}{c|}{61.73}          & \multicolumn{1}{c|}{78.41}          & \multicolumn{1}{c|}{86.10}          & \multicolumn{1}{c|}{0.6559}                              & \multicolumn{1}{c|}{0.6173}                           & \multicolumn{1}{c|}{0.5970}                             & \multicolumn{1}{c|}{88.73}          & \multicolumn{1}{c|}{93.98}          & \multicolumn{1}{c|}{96.10}          & \multicolumn{1}{c|}{0.8995}                              & \multicolumn{1}{c|}{0.8873}                            & 0.8853                             \\ \hline
\multicolumn{1}{|l|}{\textsc{ViTB16} \cite{dosovitskiy2020image}}                         & 86.6M  & \multicolumn{1}{c|}{69.71}           & \multicolumn{1}{c|}{77.60}          & \multicolumn{1}{c|}{83.28}          & \multicolumn{1}{c|}{0.7276}                              & \multicolumn{1}{c|}{0.6972}                           & \multicolumn{1}{c|}{0.6969}                             & \multicolumn{1}{c|}{68.88}             & \multicolumn{1}{c|}{76.65}          & \multicolumn{1}{c|}{81.36}          & \multicolumn{1}{c|}{0.7192}                              & \multicolumn{1}{c|}{0.6868}                           & 0.6851                             \\ \hline
\multicolumn{1}{|l|}{\textsc{ViTL32} \cite{dosovitskiy2020image}}                         & 306.5M  & \multicolumn{1}{c|}{85.10}          & \multicolumn{1}{c|}{91.56}          & \multicolumn{1}{c|}{94.48}          & \multicolumn{1}{c|}{0.8720}                              & \multicolumn{1}{c|}{0.8510}                           & \multicolumn{1}{c|}{0.8509}                             & \multicolumn{1}{c|}{84.71}          & \multicolumn{1}{c|}{91.80}          & \multicolumn{1}{c|}{94.08}           & \multicolumn{1}{c|}{0.8632}                              & \multicolumn{1}{c|}{0.8472}                           & 0.8465                             \\ \hline
\multicolumn{1}{|l|}{\textsc{AlexNet} \cite{krizhevsky2014one}}                         & 61.1M  & \multicolumn{1}{c|}{87.01}          & \multicolumn{1}{c|}{93.61}          & \multicolumn{1}{c|}{95.46}          & \multicolumn{1}{c|}{0.8840}                               & \multicolumn{1}{c|}{0.8702}                           & \multicolumn{1}{c|}{0.8708}                             & \multicolumn{1}{c|}{88.18}          & \multicolumn{1}{c|}{92.58}          & \multicolumn{1}{c|}{94.80}          & \multicolumn{1}{c|}{0.8887}                              & \multicolumn{1}{c|}{0.8818}                           & 0.8809                             \\ \hline
\multicolumn{1}{|l|}{\textsc{SqueezeNet1\_1} \cite{Iandola2016SqueezeNetAA}}                 & 1.2M  & \multicolumn{1}{c|}{87.56}          & \multicolumn{1}{c|}{92.45}          & \multicolumn{1}{c|}{94.15}          & \multicolumn{1}{c|}{0.8880}                                & \multicolumn{1}{c|}{0.8757}                           & \multicolumn{1}{c|}{0.8744}                             & \multicolumn{1}{c|}{86.21}          & \multicolumn{1}{c|}{92.48}          & \multicolumn{1}{c|}{94.65}          & \multicolumn{1}{c|}{0.8754}                              & \multicolumn{1}{c|}{0.8622}                           & 0.8637                             \\ \hline
\multicolumn{1}{|l|}{\textsc{MobileNetV3Small} \cite{howard2019searching}}                & 2.5M  & \multicolumn{1}{c|}{89.48}          & \multicolumn{1}{c|}{94.31}          & \multicolumn{1}{c|}{95.76}          & \multicolumn{1}{c|}{0.9038}                              & \multicolumn{1}{c|}{0.8948}                           & \multicolumn{1}{c|}{0.8942}                             & \multicolumn{1}{c|}{89.85}          & \multicolumn{1}{c|}{94.35}           & \multicolumn{1}{c|}{96.48}          & \multicolumn{1}{c|}{0.9082}                              & \multicolumn{1}{c|}{0.8985}                           & 0.8976                             \\ \hline
\multicolumn{1}{|l|}{\textsc{SwinB} \cite{liu2021swin}}                           & 87.8M & \multicolumn{1}{c|}{90.50}          & \multicolumn{1}{c|}{95.38}          & \multicolumn{1}{c|}{97.40}          & \multicolumn{1}{c|}{0.9128}                              & \multicolumn{1}{c|}{0.9050}                           & \multicolumn{1}{c|}{0.9042}                             & \multicolumn{1}{c|}{90.20}          & \multicolumn{1}{c|}{95.40}          & \multicolumn{1}{c|}{97.08}          & \multicolumn{1}{c|}{0.9097}                              & \multicolumn{1}{c|}{0.902}                           & 0.9006                              \\ \hline
\multicolumn{1}{|l|}{\textsc{GoogLeNet} \cite{szegedy2015going}}                      & 6.6M  & \multicolumn{1}{c|}{90.73}          & \multicolumn{1}{c|}{94.65}          & \multicolumn{1}{c|}{95.70}          & \multicolumn{1}{c|}{0.9105}                              & \multicolumn{1}{c|}{0.9073}                           & \multicolumn{1}{c|}{0.9059}                             & \multicolumn{1}{c|}{92.18}          & \multicolumn{1}{c|}{95.65}          & \multicolumn{1}{c|}{96.58}           & \multicolumn{1}{c|}{0.9283}                              & \multicolumn{1}{c|}{0.9218}                           & 0.9206                             \\ \hline
\multicolumn{1}{|l|}{\textsc{ResNet18} \cite{he2016deep}}                       & 11.7M  & \multicolumn{1}{c|}{90.91}          & \multicolumn{1}{c|}{95.28}          & \multicolumn{1}{c|}{96.60}           & \multicolumn{1}{c|}{0.9176}                              & \multicolumn{1}{c|}{0.9092}                            & \multicolumn{1}{c|}{0.9069}                             & \multicolumn{1}{c|}{91.25}          & \multicolumn{1}{c|}{95.43}          & \multicolumn{1}{c|}{97.05}          & \multicolumn{1}{c|}{0.9229}                              & \multicolumn{1}{c|}{0.9125}                           & 0.9119                             \\ \hline
\multicolumn{1}{|l|}{\textsc{MobileNetV3Large} \cite{howard2019searching}}               & 5.5M  & \multicolumn{1}{c|}{91.20}          & \multicolumn{1}{c|}{94.48}          & \multicolumn{1}{c|}{95.98}          & \multicolumn{1}{c|}{0.9185}                              & \multicolumn{1}{c|}{0.9120}                           & \multicolumn{1}{c|}{0.9110}                               & \multicolumn{1}{c|}{90.40}          & \multicolumn{1}{c|}{94.53}          & \multicolumn{1}{c|}{95.26}          & \multicolumn{1}{c|}{0.9147}                              & \multicolumn{1}{c|}{0.9040}                            & 0.9024                             \\ \hline
\multicolumn{1}{|l|}{\textsc{ConvNeXt} \cite{liu2022convnet}}                       & 88.6M  & \multicolumn{1}{c|}{91.46}          & \multicolumn{1}{c|}{96.33}           & \multicolumn{1}{c|}{98.05}          & \multicolumn{1}{c|}{0.9220}                              & \multicolumn{1}{c|}{0.9147}                            & \multicolumn{1}{c|}{0.9140}                             & \multicolumn{1}{c|}{92.55}          & \multicolumn{1}{c|}{96.36}          & \multicolumn{1}{c|}{97.96}          & \multicolumn{1}{c|}{0.9306}                              & \multicolumn{1}{c|}{0.9255}                           & 0.9246                             \\ \hline
\multicolumn{1}{|l|}{\textsc{SwinV2B} \cite{liu2022swin}}                         & 87.9M & \multicolumn{1}{c|}{91.58}          & \multicolumn{1}{c|}{96.25}          & \multicolumn{1}{c|}{97.61}          & \multicolumn{1}{c|}{0.9210}                              & \multicolumn{1}{c|}{0.9158}                           & \multicolumn{1}{c|}{0.9151}                             & \multicolumn{1}{c|}{91.48}          & \multicolumn{1}{c|}{96.00}          & \multicolumn{1}{c|}{97.55}          & \multicolumn{1}{c|}{0.9209}                               & \multicolumn{1}{c|}{0.9148}                           & 0.9144                             \\ \hline
\multicolumn{1}{|l|}{\textsc{VGG16} \cite{simonyan2014very}}                          & 138.4M  & \multicolumn{1}{c|}{91.61}          & \multicolumn{1}{c|}{95.08}          & \multicolumn{1}{c|}{96.65}          & \multicolumn{1}{c|}{0.9248}                              & \multicolumn{1}{c|}{0.9162}                           & \multicolumn{1}{c|}{0.9168}                             & \multicolumn{1}{c|}{91.00}          & \multicolumn{1}{c|}{95.21}          & \multicolumn{1}{c|}{96.45}           & \multicolumn{1}{c|}{0.9235}                               & \multicolumn{1}{c|}{0.9100}                           & 0.9119                            \\ \hline
\multicolumn{1}{|l|}{\textsc{MNasNet13} \cite{tan2019mnasnet}}                      & 6.3M  & \multicolumn{1}{c|}{91.66}          & \multicolumn{1}{c|}{95.65}          & \multicolumn{1}{c|}{97.01}          & \multicolumn{1}{c|}{0.9240}                              & \multicolumn{1}{c|}{0.9167}                           & \multicolumn{1}{c|}{0.9149}                             & \multicolumn{1}{c|}{91.45}          & \multicolumn{1}{c|}{95.86}          & \multicolumn{1}{c|}{97.26}          & \multicolumn{1}{c|}{0.9231}                              & \multicolumn{1}{c|}{0.9145}                           & 0.9133                             \\ \hline
\multicolumn{1}{|l|}{\textsc{ConvNextLarge} \cite{liu2022convnet}}                   & 197.8M  & \multicolumn{1}{c|}{91.88}          & \multicolumn{1}{c|}{95.90}          & \multicolumn{1}{c|}{97.70}          & \multicolumn{1}{c|}{0.9254}                              & \multicolumn{1}{c|}{0.9188}                           & \multicolumn{1}{c|}{0.9181}                             & \multicolumn{1}{c|}{88.00}          & \multicolumn{1}{c|}{94.70}          & \multicolumn{1}{c|}{96.56}           & \multicolumn{1}{c|}{0.8942}                              & \multicolumn{1}{c|}{0.8800}                           & 0.8782                            \\ \hline
\multicolumn{1}{|l|}{\textsc{EfficientNetB0} \cite{tan2019efficientnet}}                & 5.3M   & \multicolumn{1}{c|}{91.93}          & \multicolumn{1}{c|}{95.26}          & \multicolumn{1}{c|}{96.71}          & \multicolumn{1}{c|}{0.9257}                              & \multicolumn{1}{c|}{0.9193}                            & \multicolumn{1}{c|}{0.9178}                             & \multicolumn{1}{c|}{90.40}          & \multicolumn{1}{c|}{93.75}          & \multicolumn{1}{c|}{95.10}          & \multicolumn{1}{c|}{0.9131}                              & \multicolumn{1}{c|}{0.9040}                           & 0.9022                             \\ \hline
\multicolumn{1}{|l|}{\textsc{MaxViT} \cite{tu2022maxvit}}                          & 30.9M & \multicolumn{1}{c|}{92.01}          & \multicolumn{1}{c|}{96.50}          & \multicolumn{1}{c|}{97.81}          & \multicolumn{1}{c|}{0.9268}                              & \multicolumn{1}{c|}{0.9202}                           & \multicolumn{1}{c|}{0.9214}                             & \multicolumn{1}{c|}{92.08}           & \multicolumn{1}{c|}{95.98}          & \multicolumn{1}{c|}{97.36}          & \multicolumn{1}{c|}{0.9320}                              & \multicolumn{1}{c|}{0.9208}                            & 0.9237                              \\ \hline
\multicolumn{1}{|l|}{\textsc{EfficientNetV2S} \cite{tan2021efficientnetv2}}               & 21.5M   & \multicolumn{1}{c|}{92.31}          & \multicolumn{1}{c|}{95.75}          & \multicolumn{1}{c|}{96.76}          & \multicolumn{1}{c|}{0.9375}                              & \multicolumn{1}{c|}{0.9232}                           & \multicolumn{1}{c|}{0.9245}                             & \multicolumn{1}{c|}{\textbf{94.45}}          & \multicolumn{1}{c|}{\textbf{97.35}}          & \multicolumn{1}{c|}{\textbf{98.30}}          & \multicolumn{1}{c|}{0.9498}                               & \multicolumn{1}{c|}{\textbf{0.9445}}                           & 0.9438                             \\ \hline
\multicolumn{1}{|l|}{\textsc{VGG19} \cite{simonyan2014very}}                          & 143.7M  & \multicolumn{1}{c|}{92.36}          & \multicolumn{1}{c|}{95.13}          & \multicolumn{1}{c|}{96.10}          & \multicolumn{1}{c|}{0.9354}                              & \multicolumn{1}{c|}{0.9237}                           & \multicolumn{1}{c|}{0.9242}                             & \multicolumn{1}{c|}{91.80}          & \multicolumn{1}{c|}{95.06}          & \multicolumn{1}{c|}{96.15}          & \multicolumn{1}{c|}{0.9296}                              & \multicolumn{1}{c|}{0.9180}                           & 0.9187                            \\ \hline
\multicolumn{1}{|l|}{\textsc{MobileNetV2} \cite{sandler2018mobilenetv2}}                   & 3.5M   & \multicolumn{1}{c|}{92.38}           & \multicolumn{1}{c|}{94.98}          & \multicolumn{1}{c|}{96.05}          & \multicolumn{1}{c|}{0.9303}                              & \multicolumn{1}{c|}{0.9238}                           & \multicolumn{1}{c|}{0.9233}                             & \multicolumn{1}{c|}{92.31}          & \multicolumn{1}{c|}{95.38}          & \multicolumn{1}{c|}{96.91}          & \multicolumn{1}{c|}{0.9311}                              & \multicolumn{1}{c|}{0.9232}                           & 0.9225                             \\ \hline
\multicolumn{1}{|l|}{\textsc{WideResNet50\_2} \cite{zagoruyko2016wide}}                & 68.9M  & \multicolumn{1}{c|}{92.46}          & \multicolumn{1}{c|}{96.28}          & \multicolumn{1}{c|}{97.28}          & \multicolumn{1}{c|}{0.9331}                              & \multicolumn{1}{c|}{0.9247}                           & \multicolumn{1}{c|}{0.9235}                             & \multicolumn{1}{c|}{93.35}          & \multicolumn{1}{c|}{95.73}          & \multicolumn{1}{c|}{97.15}          & \multicolumn{1}{c|}{0.9421}                              & \multicolumn{1}{c|}{0.9335}                            & 0.9330                             \\ \hline
\multicolumn{1}{|l|}{\textsc{ResNet50} \cite{he2016deep}}                       & 25.6M  & \multicolumn{1}{c|}{92.58}          & \multicolumn{1}{c|}{95.56}          & \multicolumn{1}{c|}{96.75}          & \multicolumn{1}{c|}{0.9332}                              & \multicolumn{1}{c|}{0.9258}                           & \multicolumn{1}{c|}{0.9252}                             & \multicolumn{1}{c|}{93.08}          & \multicolumn{1}{c|}{96.48}          & \multicolumn{1}{c|}{97.20}          & \multicolumn{1}{c|}{0.9363}                              & \multicolumn{1}{c|}{0.9308}                           & 0.9299                            \\ \hline
\multicolumn{1}{|l|}{\textsc{RegNetX32GF} \cite{radosavovic2020designing}}                    & 107.8M  & \multicolumn{1}{c|}{92.86}          & \multicolumn{1}{c|}{95.71}          & \multicolumn{1}{c|}{96.93}          & \multicolumn{1}{c|}{0.9348}                              & \multicolumn{1}{c|}{0.9287}                           & \multicolumn{1}{c|}{0.9269}                              & \multicolumn{1}{c|}{92.91}          & \multicolumn{1}{c|}{95.71}          & \multicolumn{1}{c|}{96.95}          & \multicolumn{1}{c|}{0.9366}                              & \multicolumn{1}{c|}{0.9292}                           & 0.9282                             \\ \hline
\multicolumn{1}{|l|}{\textsc{DenseNet121} \cite{huang2017densely}}                    & 8.0M & \multicolumn{1}{c|}{92.93}           & \multicolumn{1}{c|}{95.75}           & \multicolumn{1}{c|}{96.88}          & \multicolumn{1}{c|}{0.9367}                              & \multicolumn{1}{c|}{0.9293}                           & \multicolumn{1}{c|}{0.9282}                             & \multicolumn{1}{c|}{92.95}          & \multicolumn{1}{c|}{95.51}          & \multicolumn{1}{c|}{96.56}          & \multicolumn{1}{c|}{0.9360}                              & \multicolumn{1}{c|}{0.9295}                            & 0.9285                            \\ \hline
\multicolumn{1}{|l|}{\textsc{ResNext101\_32X8D} \cite{xie2017aggregated}}              & 88.8M  & \multicolumn{1}{c|}{93.00}          & \multicolumn{1}{c|}{96.41}          & \multicolumn{1}{c|}{97.23}          & \multicolumn{1}{c|}{0.9364}                              & \multicolumn{1}{c|}{0.9310}                           & \multicolumn{1}{c|}{0.9303}                             & \multicolumn{1}{c|}{94.20}          & \multicolumn{1}{c|}{96.61}          & \multicolumn{1}{c|}{97.58}             & \multicolumn{1}{c|}{\textbf{0.9520}}                              & \multicolumn{1}{c|}{0.9420}                            & 0.9423                             \\ \hline
\multicolumn{1}{|l|}{\textsc{WideResNet101\_2} \cite{zagoruyko2016wide}}               & 126.9M  & \multicolumn{1}{c|}{93.36}          & \multicolumn{1}{c|}{95.81}          & \multicolumn{1}{c|}{96.90}           & \multicolumn{1}{c|}{0.9423}                              & \multicolumn{1}{c|}{0.9337}                           & \multicolumn{1}{c|}{0.9332}                             & \multicolumn{1}{c|}{92.73}          & \multicolumn{1}{c|}{96.35}          & \multicolumn{1}{c|}{97.63}          & \multicolumn{1}{c|}{0.9337}                              & \multicolumn{1}{c|}{0.9273}                            & 0.9267                             \\ \hline
\multicolumn{1}{|l|}{\textsc{InceptionV3} \cite{szegedy2016rethinking}}                   & 27.2M   & \multicolumn{1}{c|}{93.50}          & \multicolumn{1}{c|}{96.48}          & \multicolumn{1}{c|}{97.35}          & \multicolumn{1}{c|}{0.9401}                              & \multicolumn{1}{c|}{0.9350}                           & \multicolumn{1}{c|}{0.9338}                             & \multicolumn{1}{c|}{93.71}           & \multicolumn{1}{c|}{96.36}          & \multicolumn{1}{c|}{97.06}          & \multicolumn{1}{c|}{0.9446}                              & \multicolumn{1}{c|}{0.9372}                           & 0.9371  \\ \hline
\multicolumn{1}{|l|}{\textsc{DenseNet201} \cite{huang2017densely}}                    & 20.0M  & \multicolumn{1}{c|}{93.56}          & \multicolumn{1}{c|}{95.78}          & \multicolumn{1}{c|}{96.73}          & \multicolumn{1}{c|}{0.9450}                              & \multicolumn{1}{c|}{0.9357}                           & \multicolumn{1}{c|}{0.9353}                             & \multicolumn{1}{c|}{94.43}          & \multicolumn{1}{c|}{97.00}          & \multicolumn{1}{c|}{97.61}          & \multicolumn{1}{c|}{0.9492}                              & \multicolumn{1}{c|}{0.9443}                           & \textbf{0.9442}                             \\ \hline
\multicolumn{1}{|l|}{\textsc{ResNet101} \cite{he2016deep}}                     & 44.5M   & \multicolumn{1}{c|}{93.81}          & \multicolumn{1}{c|}{96.23}          & \multicolumn{1}{c|}{97.71}          & \multicolumn{1}{c|}{0.9432}                              & \multicolumn{1}{c|}{0.9382}                           & \multicolumn{1}{c|}{0.9406}                             & \multicolumn{1}{c|}{93.23}          & \multicolumn{1}{c|}{96.93}          & \multicolumn{1}{c|}{98.13}             & \multicolumn{1}{c|}{0.9386}                              & \multicolumn{1}{c|}{0.9323}                           & 0.9321                            \\ \hline
\multicolumn{1}{|l|}{\textsc{ResNet152} \cite{he2016deep}}                      & 60.2M  & \multicolumn{1}{c|}{94.06}          & \multicolumn{1}{c|}{97.06}          & \multicolumn{1}{c|}{98.05}           & \multicolumn{1}{c|}{0.9447}                              & \multicolumn{1}{c|}{0.9407}                            & \multicolumn{1}{c|}{0.9394}                             & \multicolumn{1}{c|}{93.05}          & \multicolumn{1}{c|}{96.73}          & \multicolumn{1}{c|}{97.48}          & \multicolumn{1}{c|}{0.9374}                              & \multicolumn{1}{c|}{0.9305}                           & 0.9297                             \\ \hline
\multicolumn{1}{|l|}{\textsc{ResNet34} \cite{he2016deep}}                       & 21.8M  & \multicolumn{1}{c|}{\textbf{94.97}}          & \multicolumn{1}{c|}{\textbf{97.23}}          & \multicolumn{1}{c|}{\textbf{98.23}}          & \multicolumn{1}{c|}{\textbf{0.9516}}                              & \multicolumn{1}{c|}{\textbf{0.9497}}                           & \multicolumn{1}{c|}{\textbf{0.9491}}                             & \multicolumn{1}{c|}{91.98}          & \multicolumn{1}{c|}{95.95}          & \multicolumn{1}{c|}{97.20}          & \multicolumn{1}{c|}{0.9266}                              & \multicolumn{1}{c|}{0.9198}                           & 0.9189                                                        \\ \Xhline{3\arrayrulewidth}
\multicolumn{1}{|l|}{\textsc{ResNet34} w/ Silhouette Polygonization} & 21.8M 
        & \multicolumn{1}{c|}{92.72} & \multicolumn{1}{c|}{96.41} & \multicolumn{1}{c|}{97.51} 
        & \multicolumn{1}{c|}{0.9328} & \multicolumn{1}{c|}{0.9272} & \multicolumn{1}{c|}{0.9257} 
        & \multicolumn{1}{c|}{92.95 {\scriptsize\textcolor{ForestGreen}{$(+1.05\%)$}}} 
        & \multicolumn{1}{c|}{95.75} & \multicolumn{1}{c|}{96.61} 
        & \multicolumn{1}{c|}{0.9352 {\scriptsize\textcolor{ForestGreen}{$(+0.93\%)$}}} 
        & \multicolumn{1}{c|}{0.9295 {\scriptsize\textcolor{ForestGreen}{$(+1.05\%)$}}} 
        & \multicolumn{1}{c|}{0.9283 {\scriptsize\textcolor{ForestGreen}{$(+1.02\%)$}}} 
        \\ \hline
\multicolumn{1}{|l|}{\textsc{ResNet34} w/ Topological Features} & 21.8M 
        & \multicolumn{1}{c|}{93.72} & \multicolumn{1}{c|}{96.43} & \multicolumn{1}{c|}{97.78} 
        & \multicolumn{1}{c|}{0.9432} & \multicolumn{1}{c|}{0.9372} & \multicolumn{1}{c|}{0.9359} 
        & \multicolumn{1}{c|}{94.05 {\scriptsize\textcolor{ForestGreen}{$(+2.25\%)$}}} 
        & \multicolumn{1}{c|}{96.45 {\scriptsize\textcolor{ForestGreen}{$(+0.52\%)$}}} 
        & \multicolumn{1}{c|}{97.53 {\scriptsize\textcolor{ForestGreen}{$(+0.34\%)$}}} 
        & \multicolumn{1}{c|}{0.9476 {\scriptsize\textcolor{ForestGreen}{$(+2.27\%)$}}} 
        & \multicolumn{1}{c|}{0.9405 {\scriptsize\textcolor{ForestGreen}{$(+2.25\%)$}}} 
        & \multicolumn{1}{c|}{0.9401 {\scriptsize\textcolor{ForestGreen}{$(+2.31\%)$}}}
        \\ \Xhline{2\arrayrulewidth}
\end{tabular}
}
\label{tab:tab2}
\end{table*}
\subsubsection{Data Augmentation and Preprocessing}
In order to generate a more diverse pool of training samples, we also incorporate data transformation techniques\footnote{\url{https://pytorch.org/vision/stable/transforms.html}}---Random Resize, Random Perspective, Color Jitter, Random Invert, Random Horizontal Flip, Random Crop, Random Rotation, Gaussian Blur, and Random Affine with translation and shearing---while training the models. We choose these data augmentation techniques since the classes in \haspers are mostly rotationally asymmetric and incongruent. Consequently, the augmented samples aid in eliciting better generalization abilities and robustness for all the models. The input images that are fed to the models are appropriately resized
\textit{a priori} using Bicubic Interpolation.

\vspace{-3mm}
\subsection{Results and Findings}

\subsubsection{Performance Analysis}
The \textsc{ResNet34} model yields the best performance with a top-$1$ accuracy of $94.97\%$. The vanilla version of the model also yields the highest top-$2$ accuracy, top-$3$ accuracy, Precision, Recall, and F1-scores of $97.23\%$, $98.23\%$, $0.9516$, $0.9497$, and $0.9491$ respectively. 
Upon being equipped with the classifier block shown in Figure \ref{fig:classifier_block}, the \textsc{EfficientNetV2S} model yields the highest top-$k$ accuracies and Recall. In contrast, the \textsc{ResNext101\_32X8D} and \textsc{DenseNet201} models demonstrate the best performance across Precision and F1-score metrics respectively. At this recess of the performance analysis, we consider the top-$1$ accuracy metric to be the most statistically significant metric. As evident in Table \ref{tab:tab2}, the vanilla models listed in the upper part's penultimate row and above lag behind the \textsc{ResNet34} model when it comes to the top-$1$ accuracy value (as well as the other metrics), which is why we adjudicate that \textsc{ResNet34} is the best-performing model. We hypothesize that residual connections in ResNets help preserve low-level edge and contour information through identity mappings, ensuring that crucial silhouette boundaries aren't lost as the network deepens, making them better at capturing subtle variations.

\begin{figure*}[t]
    \centering
    \subfloat[Vanilla \textsc{ResNet34}]{\includegraphics[width=0.5\linewidth]{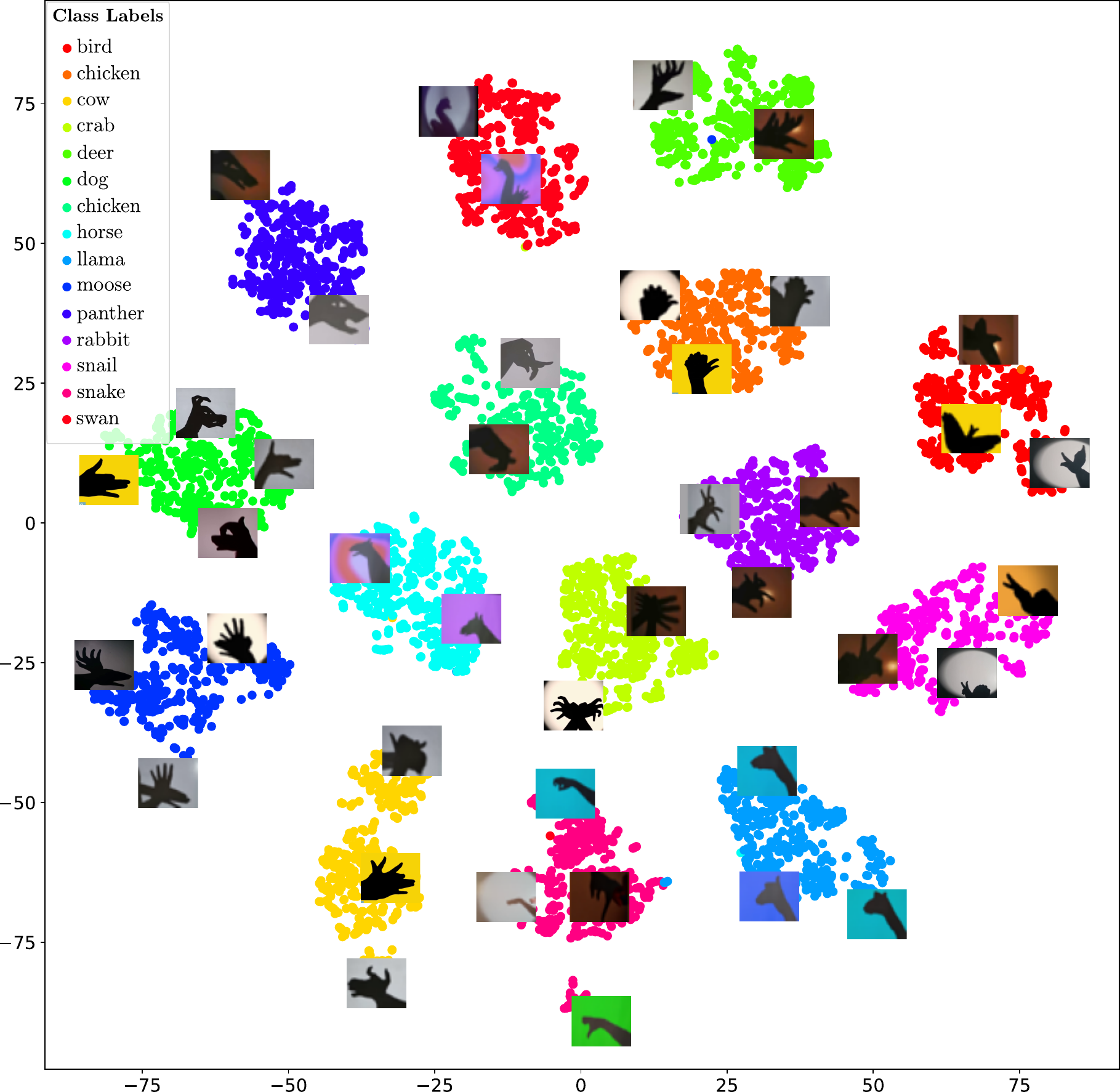}\label{fig:fig11a}}
    \subfloat[\centering\textsc{ResNet34} w/ Classifier Block]{\includegraphics[width=0.5\linewidth]{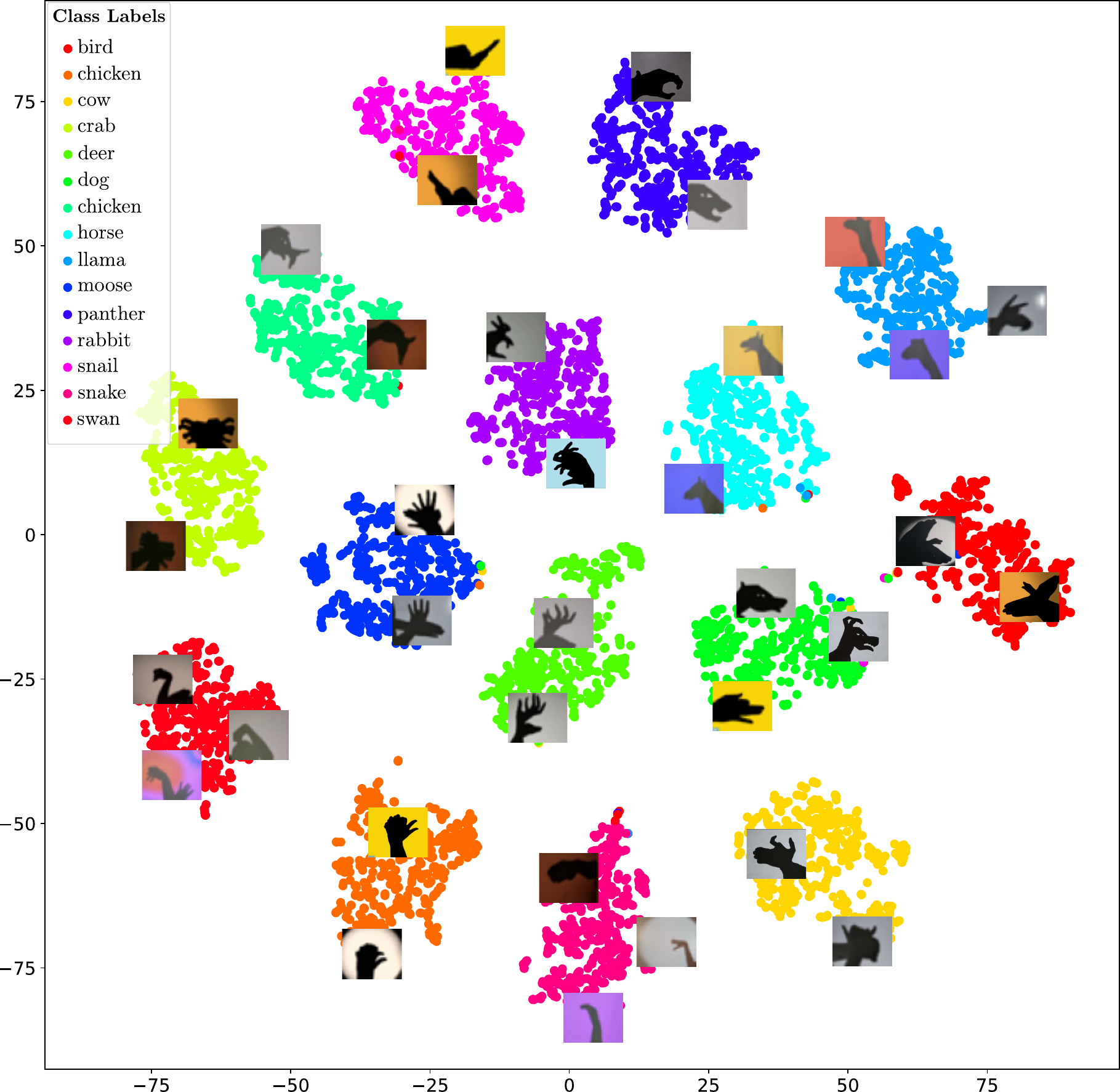}\label{fig:fig11b}}
    \caption{$t$-Distributed Stochastic Neighbor Embedding ($t$-SNE) feature representations of \textsc{ResNet34}.}
    \label{fig:fig11}
\end{figure*}
\subsubsection{Feature Space Visualization and Analysis}
In order to visualize the learned feature space of \textsc{ResNet34}, we resort to the dimensionality reduction technique called $t$-Distributed Stochastic Neighbor Embedding \cite{van2008visualizing} since it can preserve the proximity of high-dimensional data points. For high-dimensional data residing on or proximate to a low-dimensional, non-linear manifold, it becomes imperative to preserve this proximity of the collapsed low-dimensional representations for closely resembling data points. Achieving such proximity preservation is often unattainable through linear mappings such as Principal Component Analysis (PCA), which is why we opt for the $t$-SNE dimensionality reduction approach. We can pragmatically infer from the 2D-collapsed visualizations of the high-dimensional feature representations in Figure \ref{fig:fig11}, that the classes are nicely clustered and congealed with minimal overlaps and outliers. This enables the model to easily determine the decision surface in the high-dimensional feature space and perform very well on the classification task.

\subsubsection{Qualitative Analysis and Explainability}
As depicted in Figure \ref{fig:fig10}, we adopt a plethora of explainable AI \textsc{(xAI)} techniques for the best-performing \textsc{ResNet34} model to understand its decision-making.
While viewing the GradCAM (Gradient-weighted Class Activation Mapping) \cite{selvaraju2017grad} attention heatmaps, it becomes apparent that the model puts more gravitas on the common-sense distinguishing traits. For example, in Figure \ref{fig:fig10b}, we observe the regions of the image samples predominantly influencing their respective classification scores---the wingspan and beak of a bird, the gallinaceous comb of a chicken, the horns and concave head of a cow, the appendages of a crab, the horns of a deer, the long-slanted head of a dog, the tusks of an elephant, the long maxilla-mandibular jaw of a horse, the long-eared and tapered head of a llama, the upright horns of a moose, the big eyes and small ears of a panther, the petite hands and head of a rabbit, the shell and antennae of a snail, the lateral hood expansion of a snake, as well as the slender neck and wing feathers of a swan. As human beings, we evoke these same distinguishing characteristics while classifying the images using our own visual reasoning faculties. As exemplified in Figure \ref{fig:fig10c}, for local interpretation, we use the model-agnostic technique called LIME (Local Interpretable Model-agnostic Explanations) \cite{ribeiro2016should}. The green highlights indicate regions of the image that contribute positively to the probability of the assigned label, while the red highlights signify areas that reduce this probability. We also demonstrate the spatial support of the top-$1$ predicted classes by generating the saliency maps \cite{simonyan2014deep} in Figure \ref{fig:fig10d}. These maps are rendered using a solitary back-propagation pass through the \textsc{ResNet34} model, and they accentuate the salient areas of the given image, characterized by their discriminative attributes with respect to the given class.
\begin{figure}[t]
    \centering
    \subfloat[Camera feed]{\includegraphics[scale=0.16]{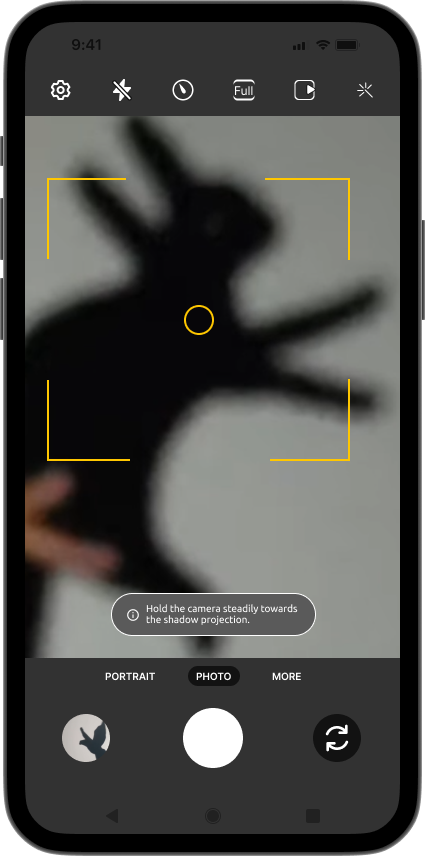}}\hspace{1cm}
    \subfloat[\centering Prediction]{\includegraphics[scale=0.16]{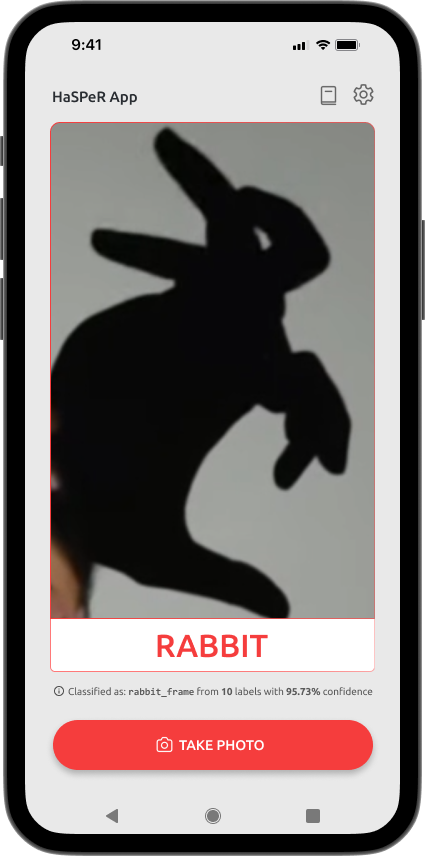}}
    \setlength{\belowcaptionskip}{-4mm}
    \caption{Android application for shadow puppet recognition.}
    \label{fig:app_screenshots}
\end{figure}
\begin{figure}[!tb]
    \centering
    \subfloat{\includegraphics[width=0.84\linewidth]{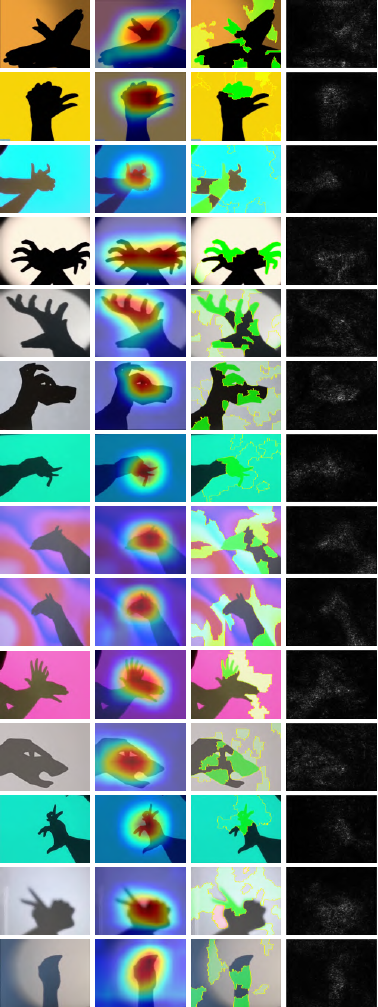}}\\\vspace{1mm}
    \setcounter{subfigure}{0}
    \subfloat[\centering{Raw Image}]{\includegraphics[width=0.19892495054949927\linewidth]{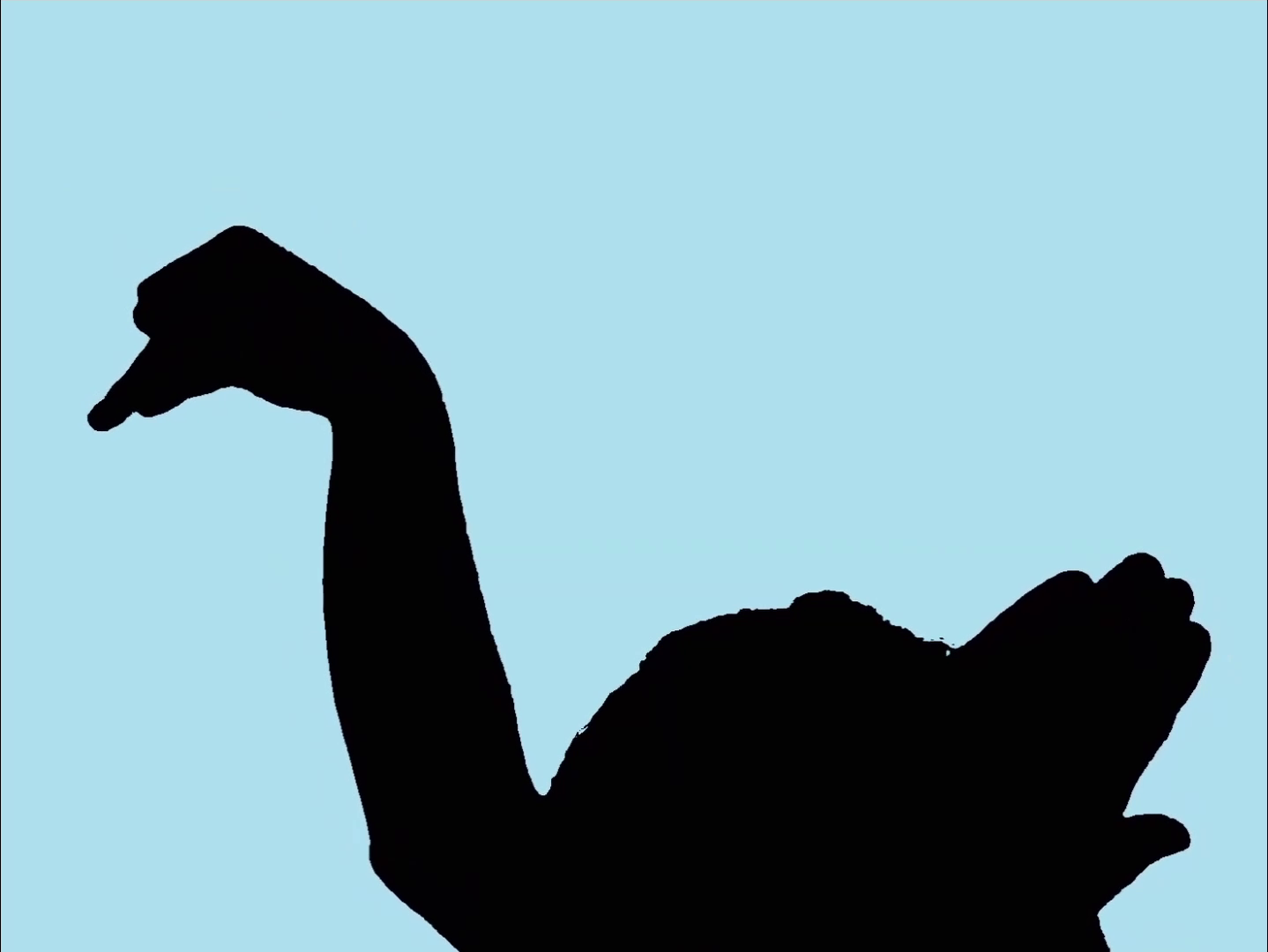}\label{fig:fig10a}}\hspace{0.12145272709532635cm}\subfloat[\centering GradCAM Heatmap]{\includegraphics[width=0.19892495054949927\linewidth]{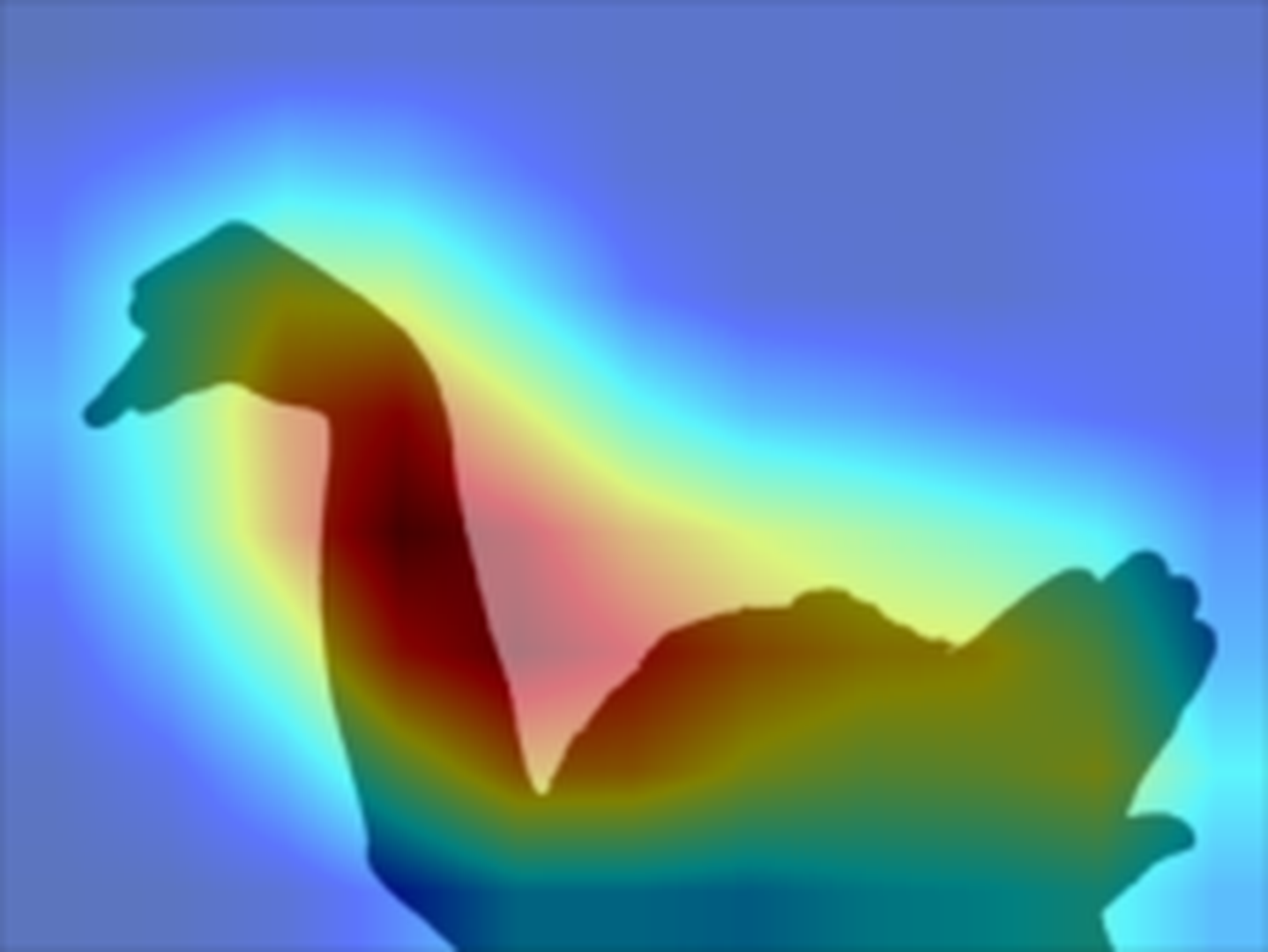}\label{fig:fig10b}}
    \hspace{0.12145272709532635cm}\subfloat[\centering LIME]{\includegraphics[width=0.19892495054949927\linewidth]{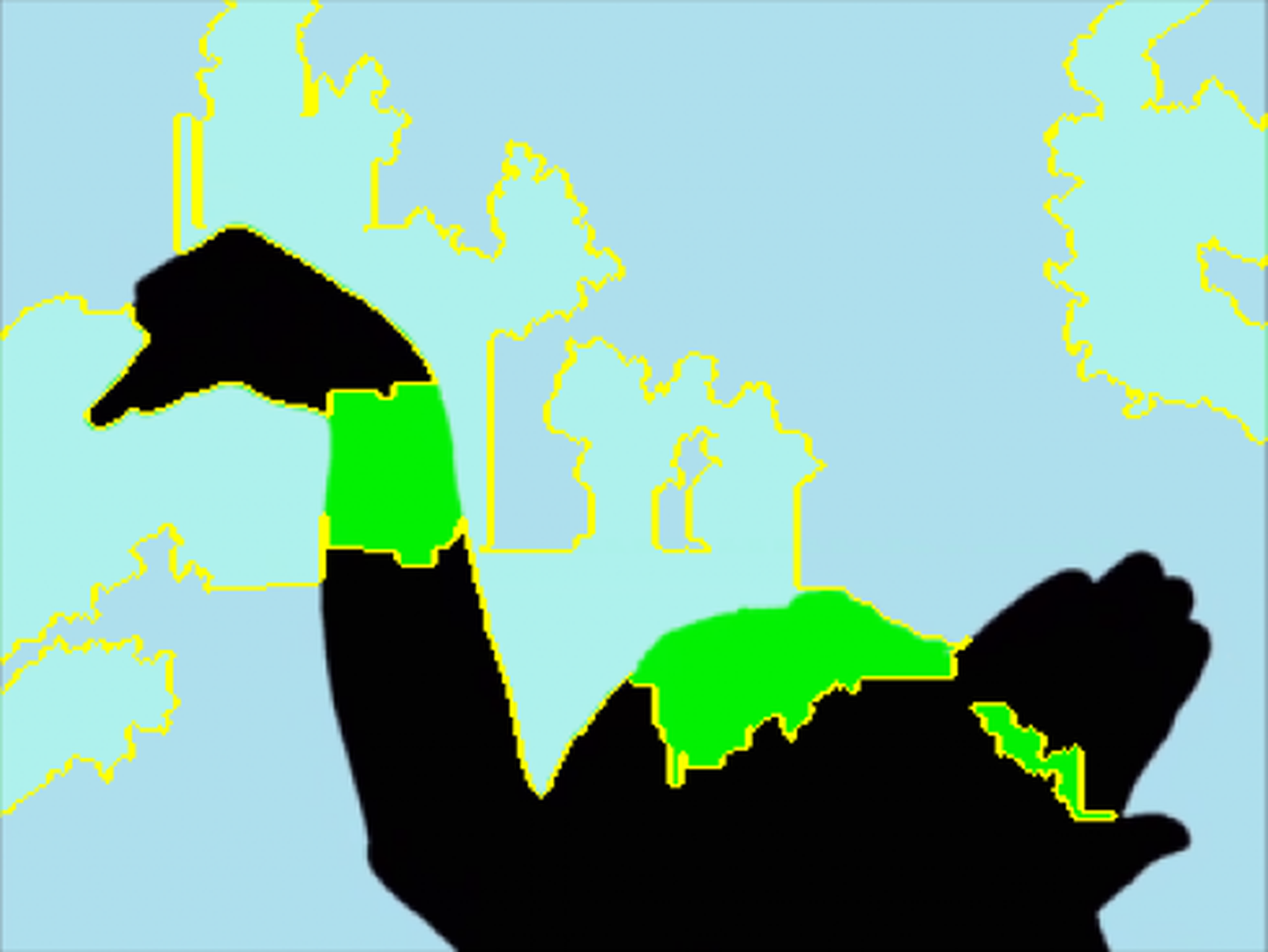}\label{fig:fig10c}}\hspace{0.12145272709532635cm}\subfloat[\centering Saliency Map]{\includegraphics[width=0.19892495054949927\linewidth]{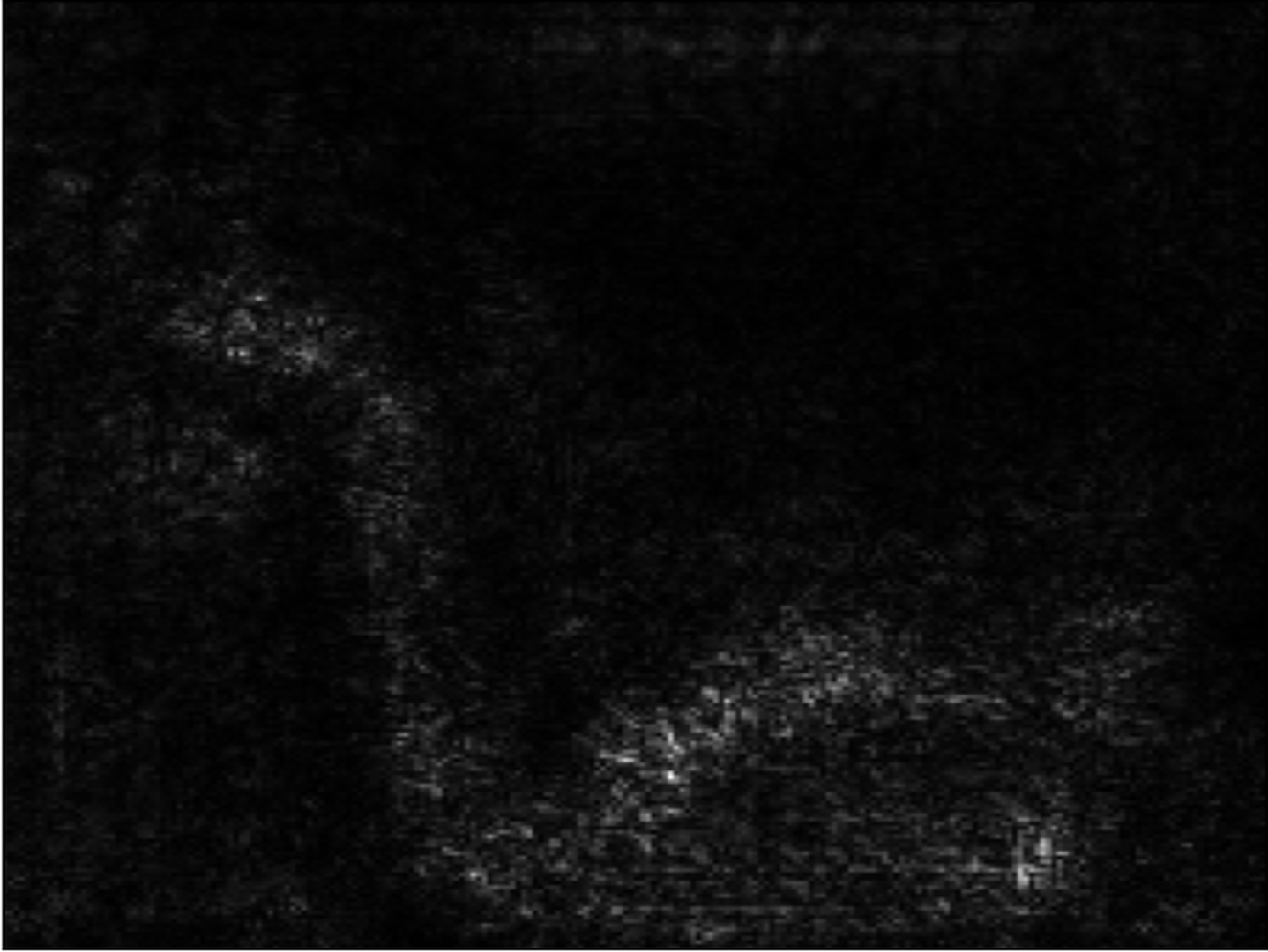}\label{fig:fig10d}}
    \setlength{\belowcaptionskip}{-7mm}
    \caption{The juxtaposition of original image samples from the {\hasper} dataset with their corresponding GradCAM Heatmaps, LIME Visualizations, and Saliency Maps (for the best-performing vanilla \textsc{ResNet34} model).}
    \label{fig:fig10}
\end{figure}
\subsubsection{Practicality Analysis as a Teaching Tool}
It is noteworthy to point out that \textsc{MobileNetV2}, with only 3.5 million parameters, managed to surpass many of the other models in terms of performance. This indicates the suitability of this image classification task for lighter, low-latency models that can be used in mobile applications and embedded devices. We create a simple prototype Android application using Flutter to test the efficacy of \textsc{MobileNetV2} in classifying hand shadow puppet images from the phone's camera feed. In order to make the prototype work as seamlessly as intended, we make sure the vicinity is well-lit, and the camera accurately captures a sharply focused silhouette. We find that the model has a memory footprint of $29$ MB and achieves an average inference time of $880$ $\mu$s on the Snapdragon 8 Gen 2 mobile chipset featured in the Samsung Galaxy S23 smartphone. Figure \ref{fig:app_screenshots} portrays the snapshots of the prototype application. There are several other practical implementation challenges involved in this endeavor that we can identify for an educational mobile application to comport well with the target demographics \cite{altwaijry2021arabic} and real-world settings. 
    The system must operate with minimal computational overhead, ensure real-time responsiveness, and maintain low latency. This necessitates a pragmatic tradeoff between the FLOP count and classification accuracy, with the requisite model compression and optimization techniques.
    Given the potential variability in device camera capabilities, the application must have preprocessing steps including, but not limited to, denoising, adaptive contrast enhancement, and sharpening the input feed to mitigate artifacts. In recognition of the diverse motor capabilities of users, particularly younger learners and users with dexterity impairments, the application must have intelligent motion compensation to stabilize the shaky camera inputs.
    Adding to the desiderata is the consideration of a suitable UI/UX that is tailored for the pediatric user base, as is done in handwriting teaching apps for children \cite{altwaijry2021arabic}, because an intuitive and enjoyable learning experience is of paramount importance for educational apps. This includes providing step-by-step tutorials, interactive guides with progressive difficulty, and illustrative diagrams to demonstrate the creation of hand shadow puppets.
  A holistically sound approach to this application development endeavor will surely potentiate the pedagogical feasibility of ombromanie.

  \begin{figure}[t]
    \centering
    \includegraphics[width=1\linewidth]{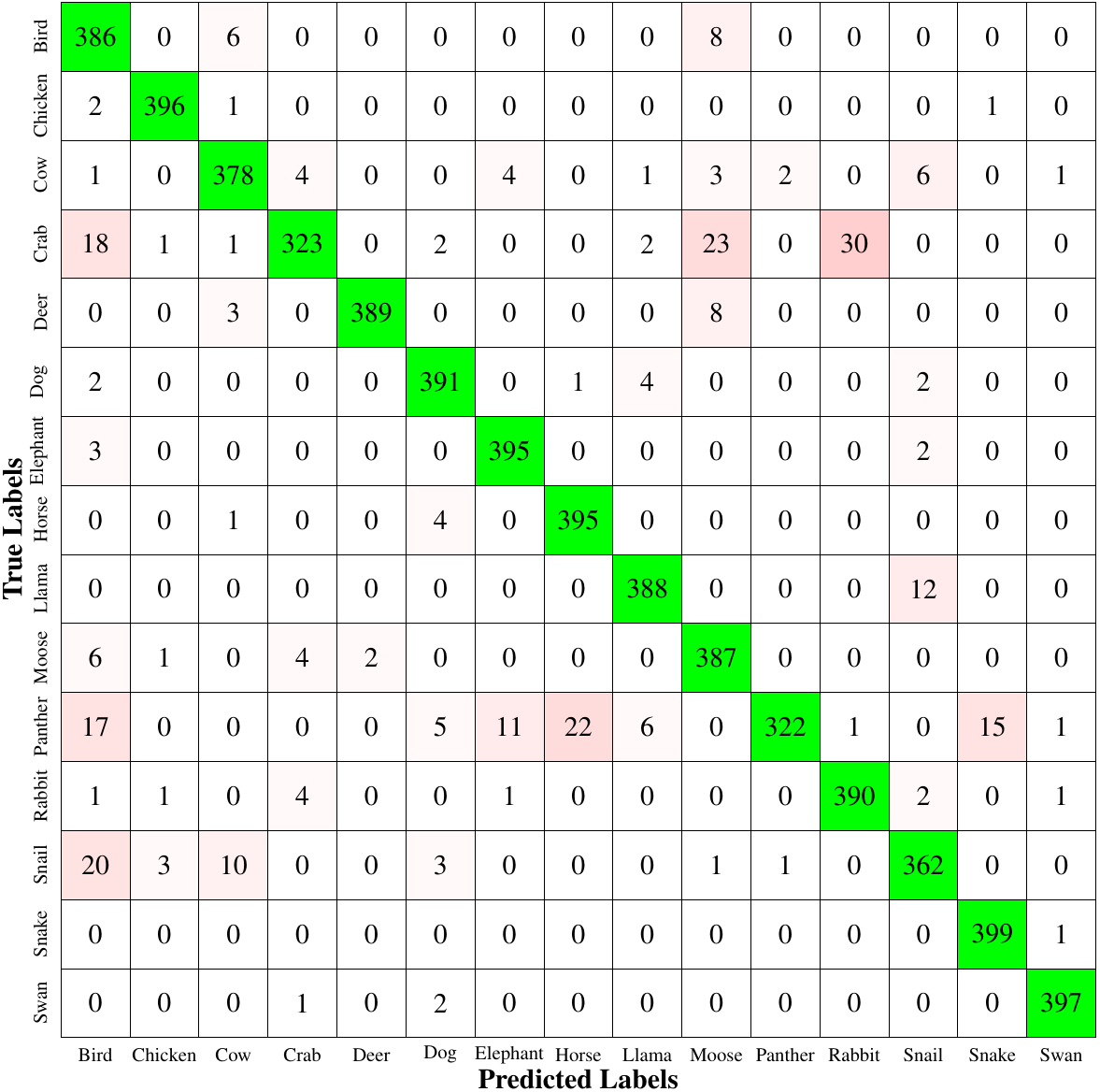}
    \setlength{\belowcaptionskip}{-4mm}
    \caption{Confusion Matrix of vanilla \textsc{ResNet34}.}
    \label{fig:confusion_matrix}
\end{figure}
\begin{figure*}[t]
    \centering
    \subfloat[\centering Actual Label: Crab, Predicted Class: Rabbit]{\includegraphics[width=0.19\linewidth]{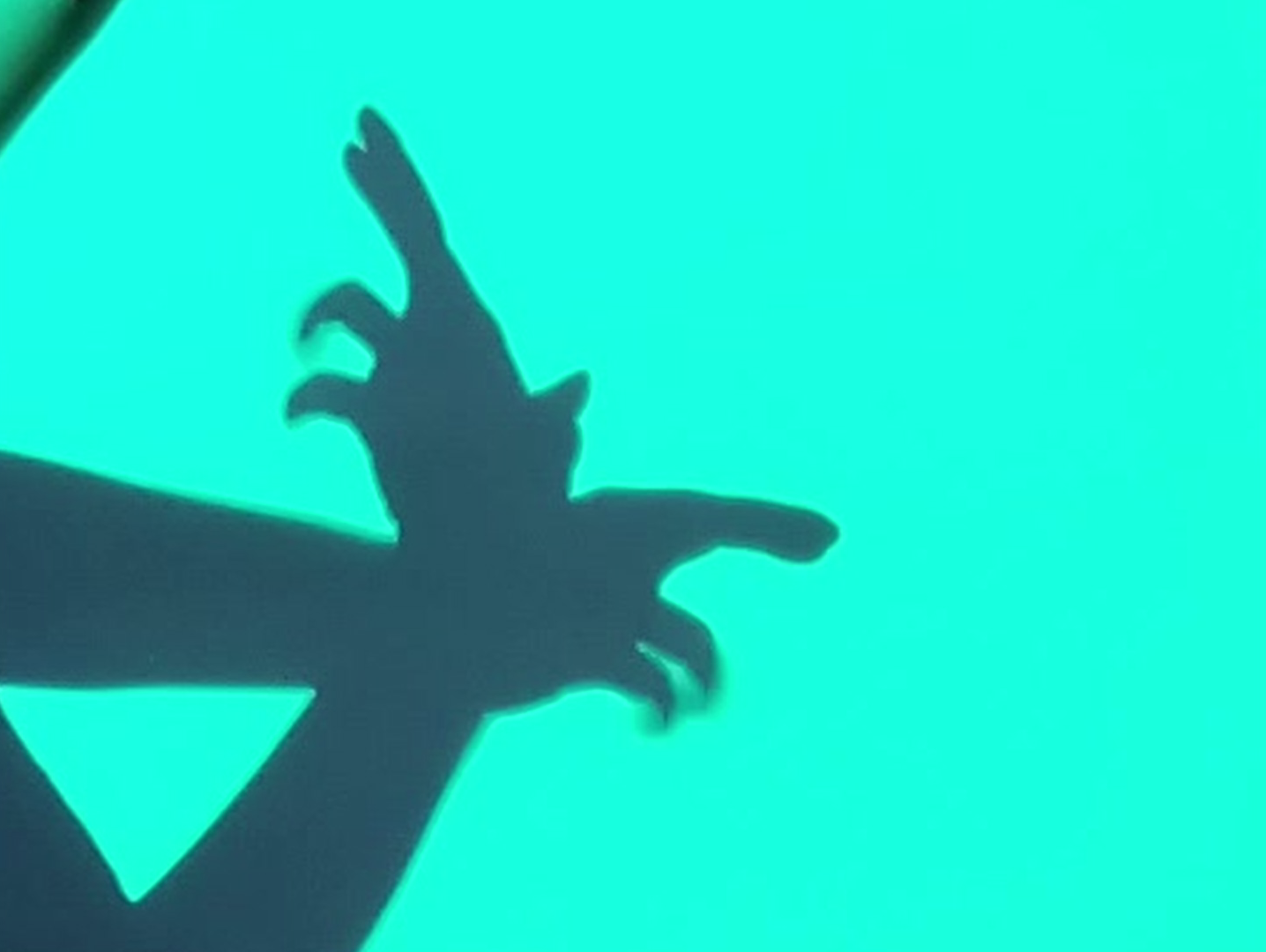}\label{fig:fig9a}}
    \hfil
    \subfloat[\centering Actual Label: Moose, Predicted Class: Bird]{\includegraphics[width=0.19\linewidth]{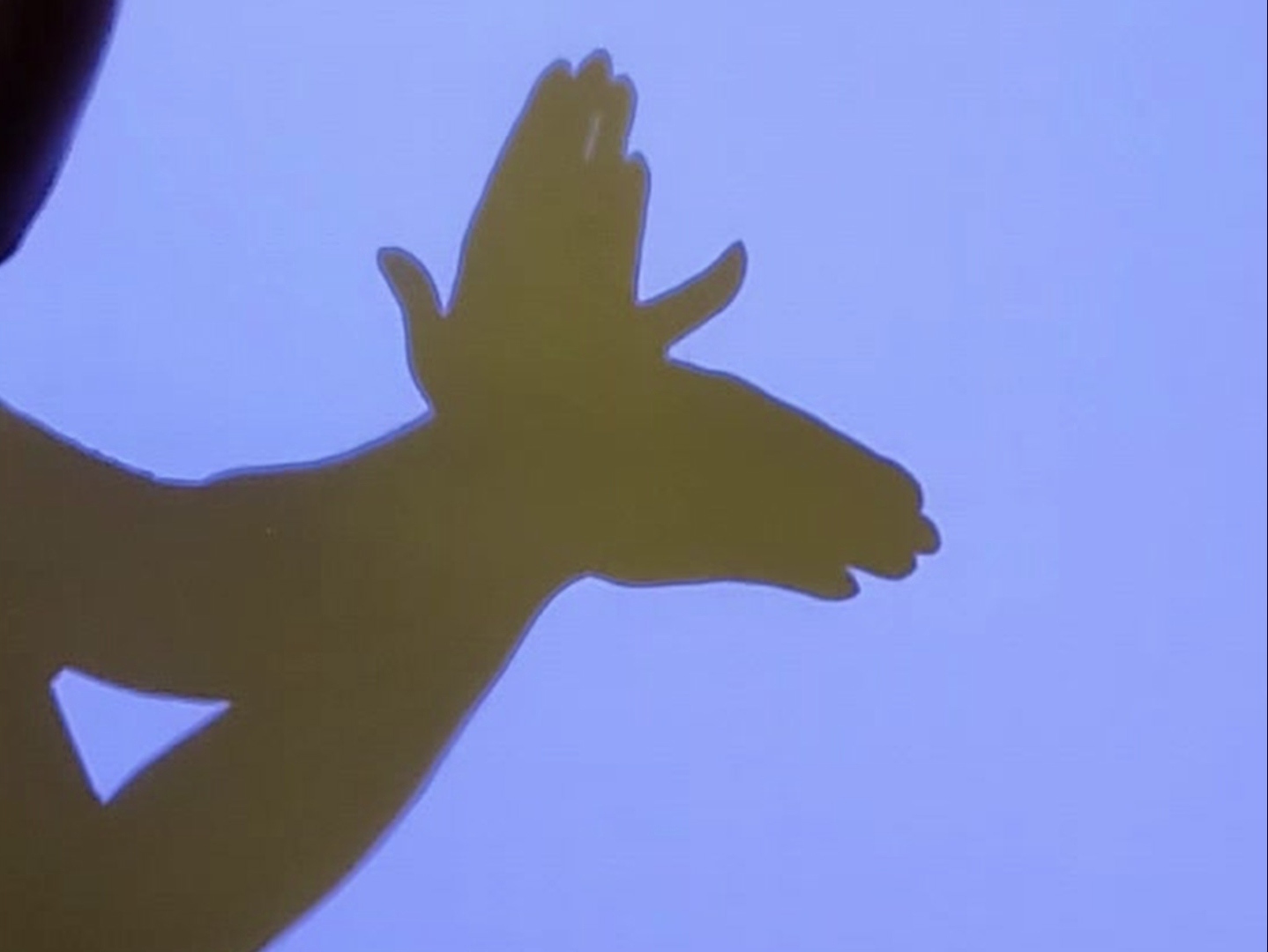}\label{fig:fig9b}}
    \hfil
    \subfloat[\centering Actual Label: Panther, Predicted Class: Bird]{\includegraphics[width=0.19\linewidth]{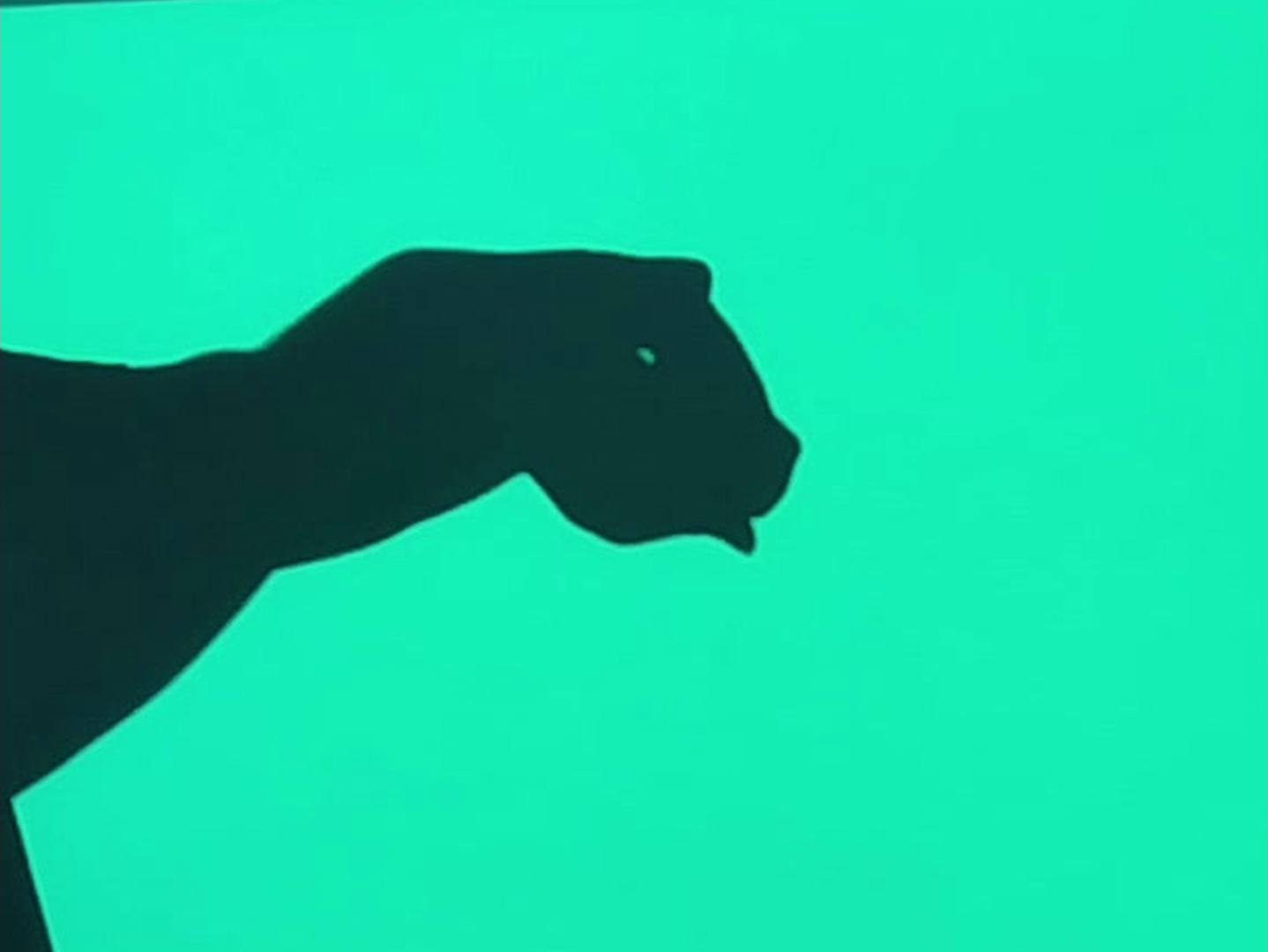}\label{fig:fig9c}}
    \hfil
    \subfloat[\centering Actual Label: Llama, Predicted Class: Snail]{\includegraphics[width=0.19\linewidth]{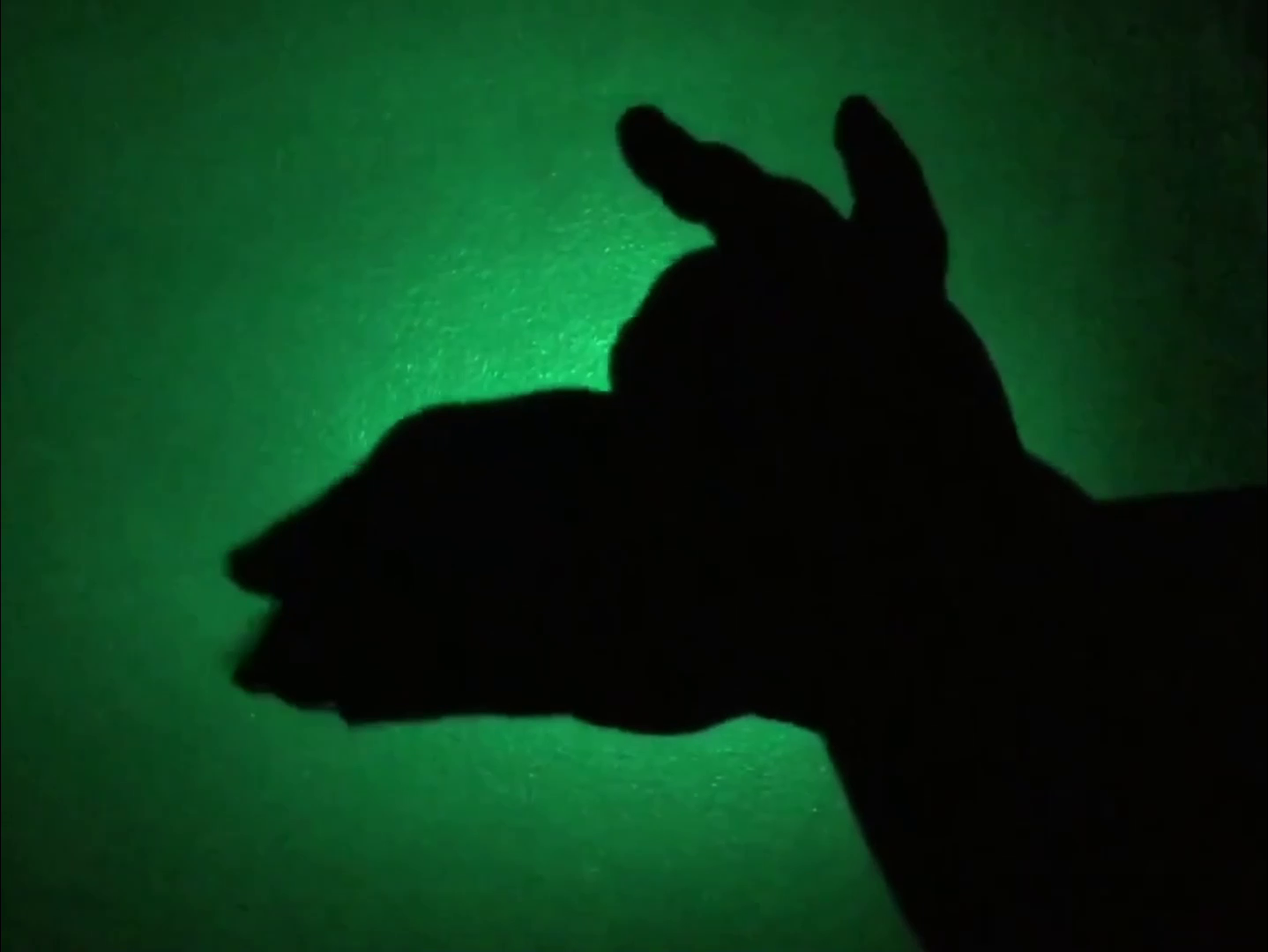}\label{fig:fig9d}}
    \hfil
    \subfloat[\centering Actual Label: Snail, Predicted Class: Bird]{\includegraphics[width=0.19\linewidth]{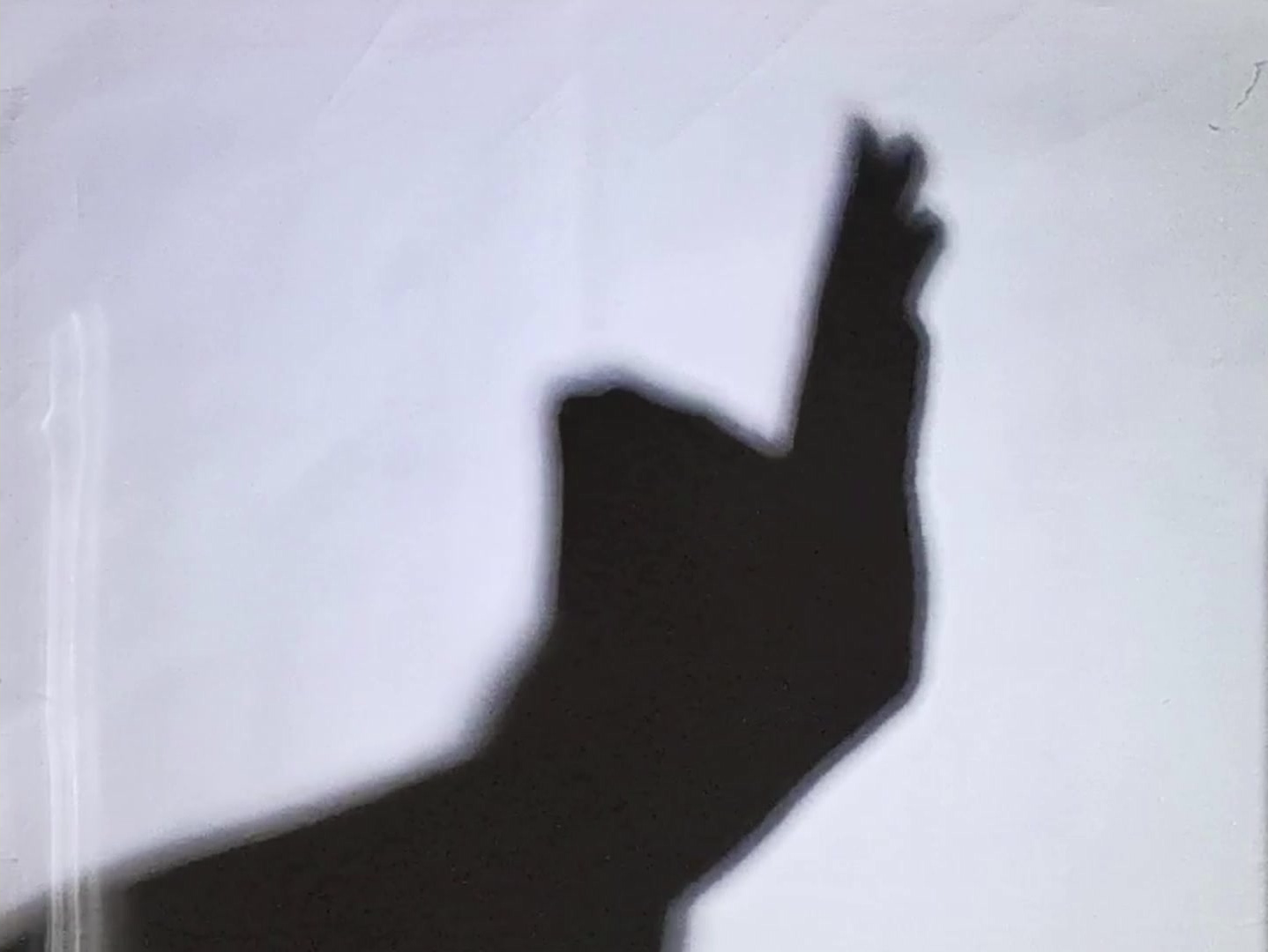}\label{fig:fig9e}}\\
    \vspace{1mm}
    \subfloat[\centering Similar sample from the `Rabbit' class]{\includegraphics[width=0.19\linewidth]{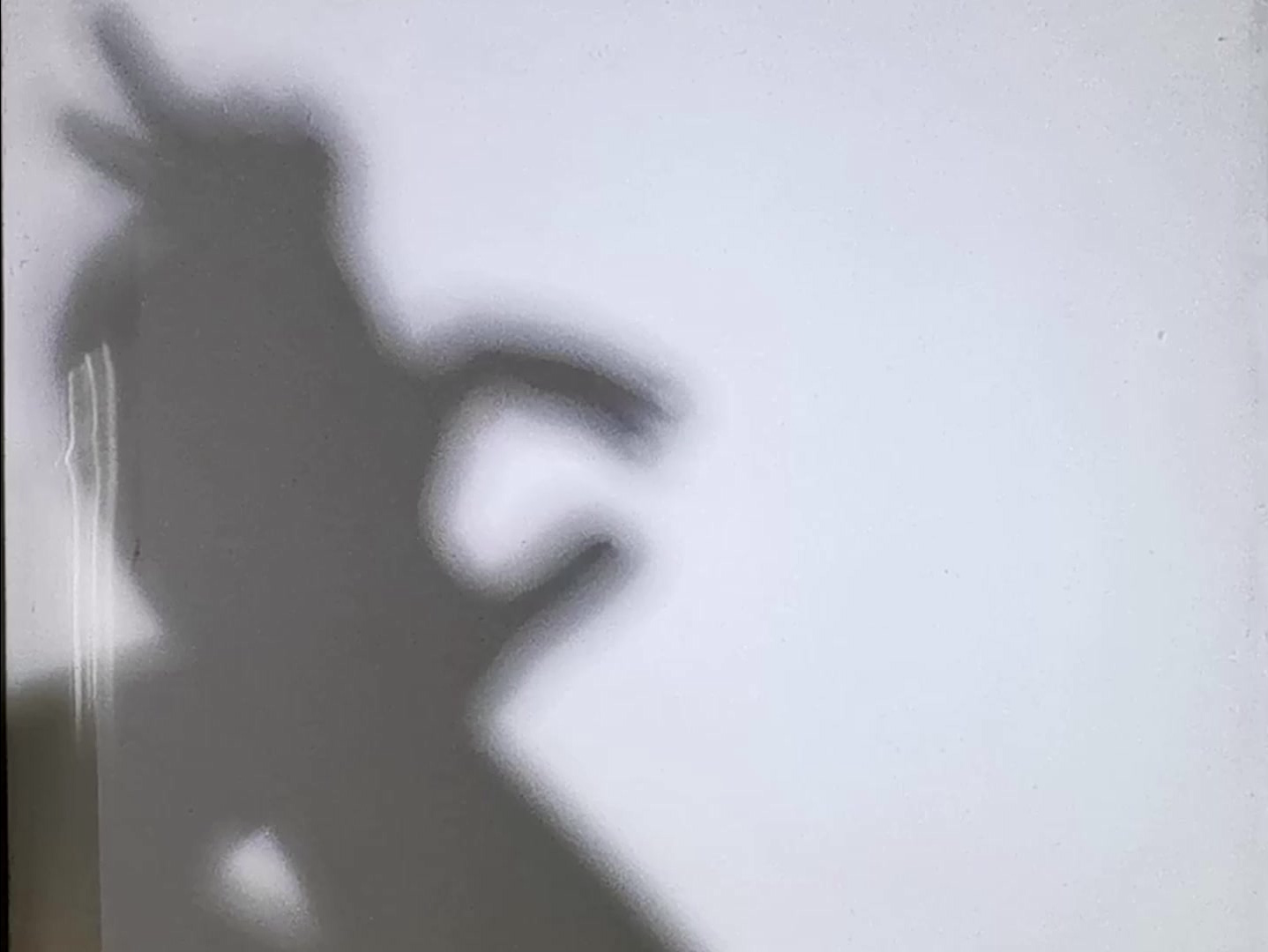}\label{fig:fig9f}}
    \hfil
    \subfloat[\centering Similar sample from the `Bird' class]{\includegraphics[width=0.19\linewidth]{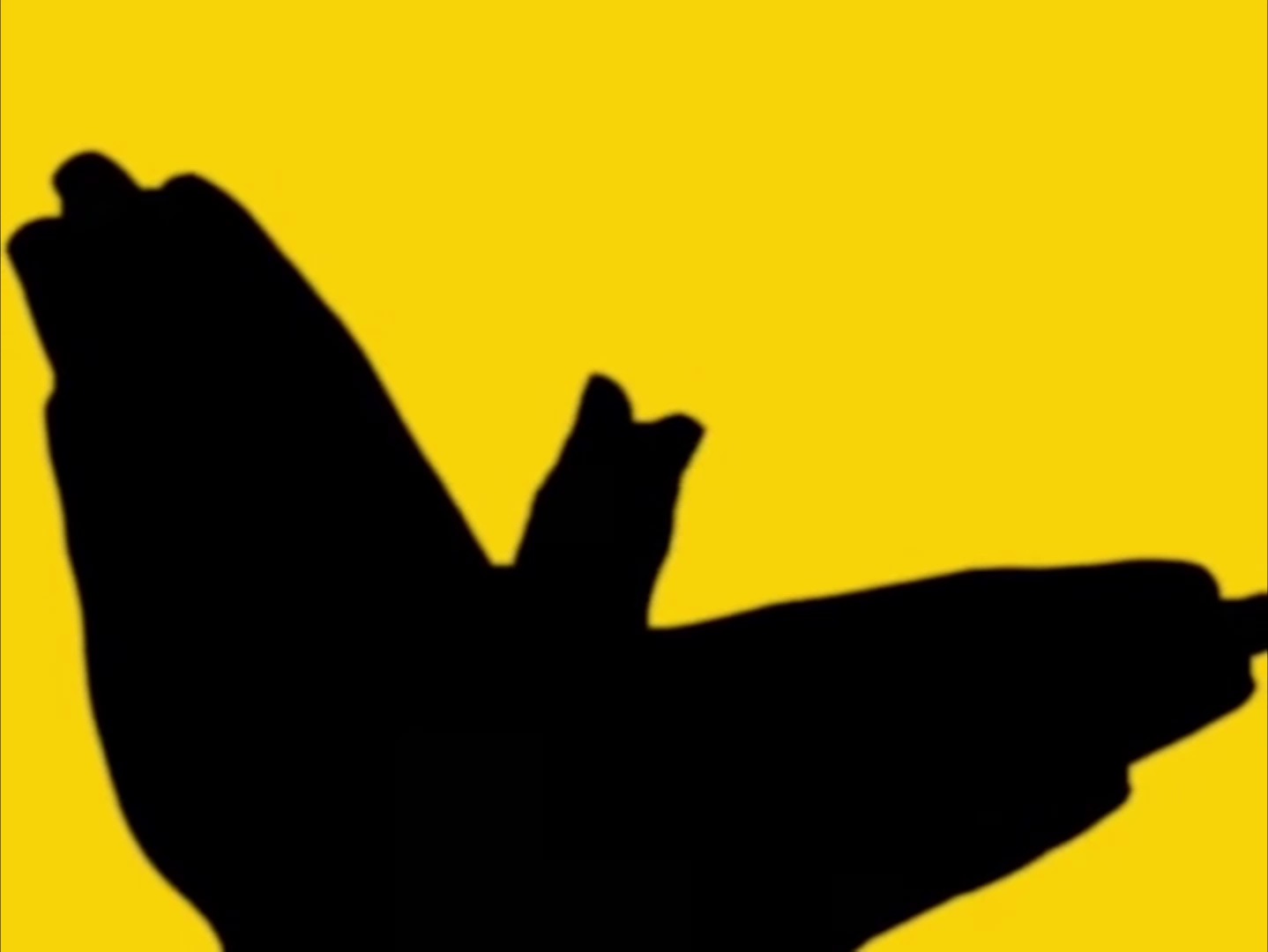}\label{fig:fig9g}}
    \hfil
    \subfloat[\centering Similar sample from the `Bird' class]{\includegraphics[width=0.19\linewidth]{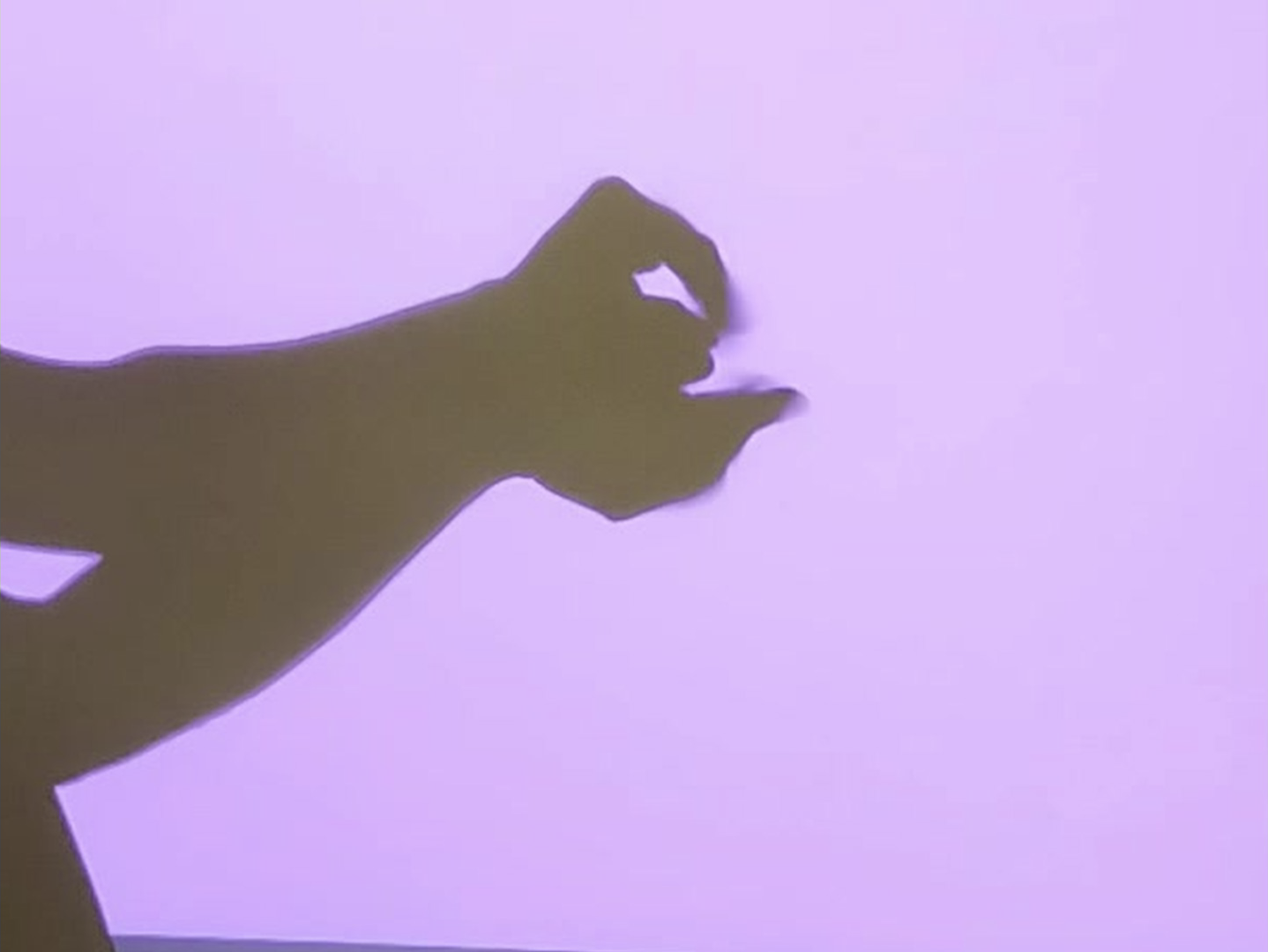}\label{fig:fig9h}}
    \hfil
    \subfloat[\centering Similar sample from the `Snail' class]{\includegraphics[width=0.19\linewidth]{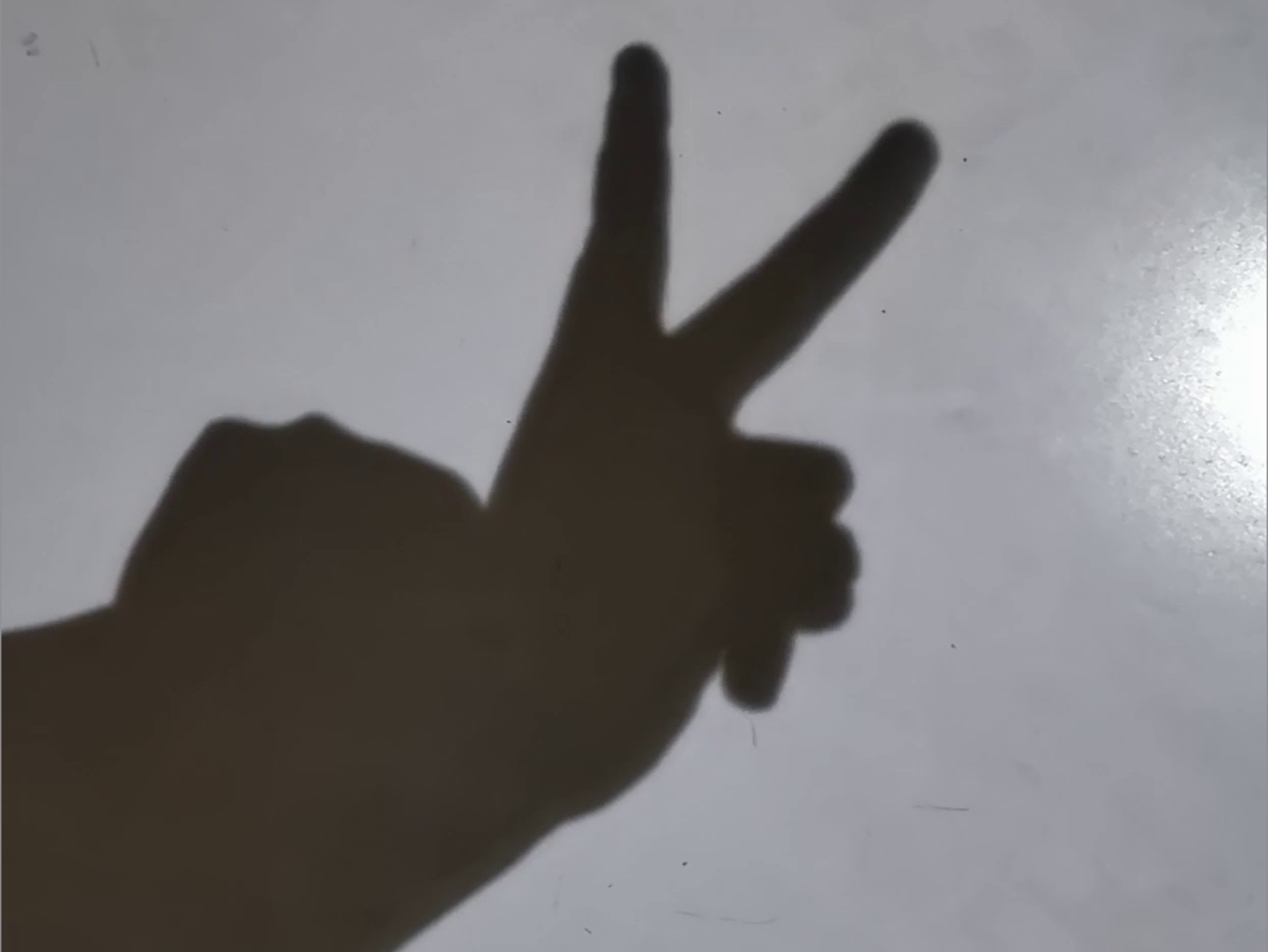}\label{fig:fig9i}}
    \hfil
    \subfloat[\centering Similar sample from the `Bird' class]{\includegraphics[width=0.19\linewidth]{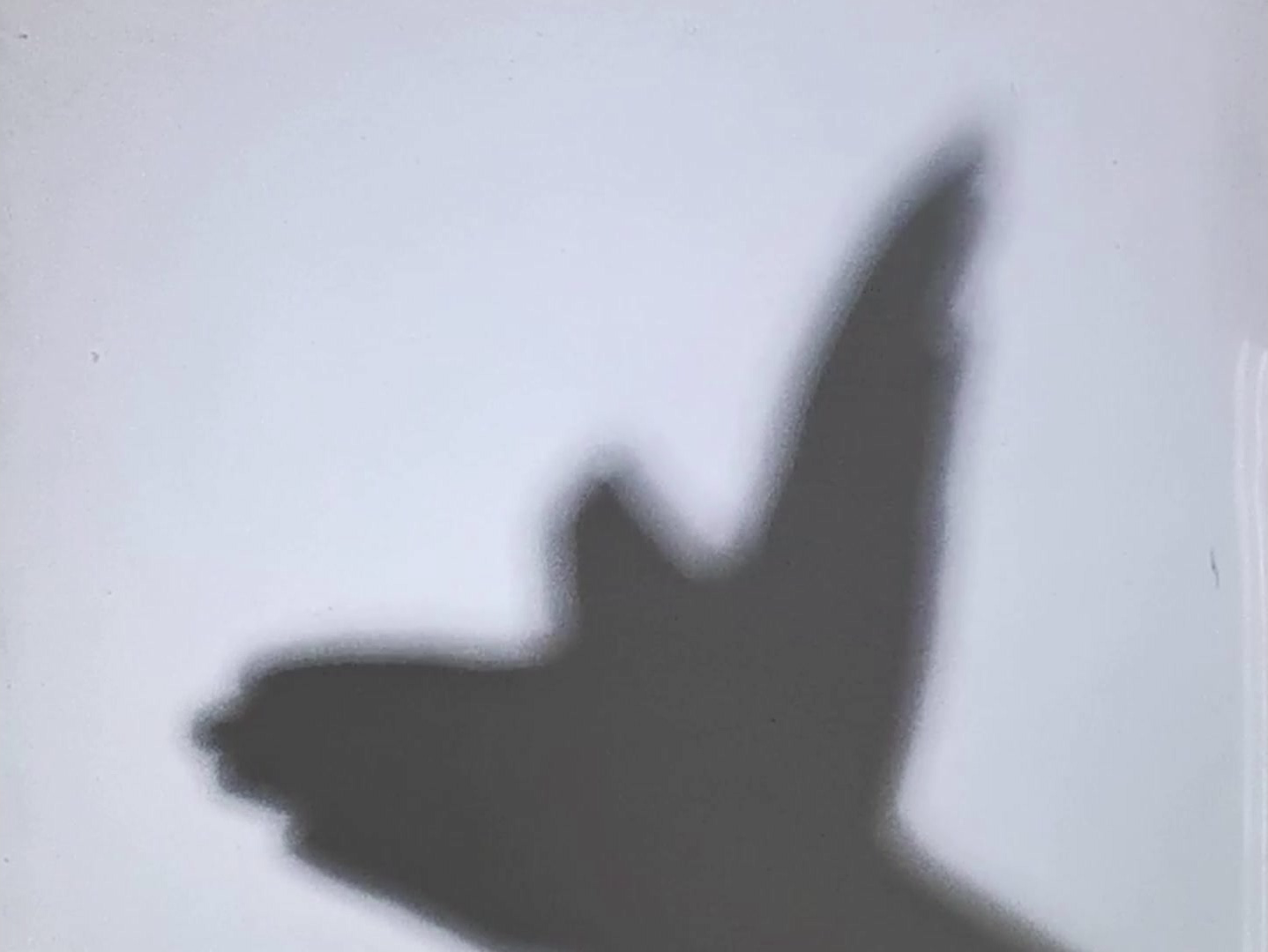}\label{fig:fig9j}}
    \setlength{\belowcaptionskip}{-4mm}
    \caption{Misclassified samples with visually similar samples of the predicted class.}
    \label{fig:fig9}
\end{figure*}
\vspace{-2mm}
\subsection{Error Analysis}
\label{sec:error_analysis}
The confusion matrix for the \textsc{ResNet34} model on our dataset, presented in Figure \ref{fig:confusion_matrix}, reveals that the `Crab' class exhibits the highest count of misclassifications. One obvious reason for this is the somewhat significant inter-class similarity among the `Bird', `Moose', `Rabbit', and `Crab' classes. 
Most of the misclassified samples are from visually similar classes. We can posit that navigating the intricacies of visually similar classes poses a significant challenge in this image classification task, as evident from the other pale-red entries of the confusion matrix in Figure \ref{fig:confusion_matrix}. Even to the keen human eye, distinguishing between these classes may be perplexing, as they share common visual features, shapes, or color patterns that result in a high degree of resemblance. We examine various aspects, such as the distinctive features or characteristics that might have led to confusion and the degree of similarity between the misclassified classes. 
Figure \ref{fig:fig9a} and \ref{fig:fig9f} show the confusion between a `Crab' sample and a `Rabbit' sample which look visually quite similar. The same holds for one of the six `Moose' samples that are misclassified as `Bird' samples by the \textsc{ResNet34} model, as depicted in Figure \ref{fig:fig9b} and \ref{fig:fig9g}. We observe that misclassifications of this type occur when images belonging to different, but visually akin categories, are erroneously assigned to the wrong class. Another reason for misclassifications is the ambiguity of shape present in mid-action frames. For instance, the `Bird' sample in Figure \ref{fig:fig9h} is a transition frame between two successive wing flaps. However, due to the presence of this ambiguous sample in the training set of the `Bird' class, the \textsc{ResNet34} ends up misclassifying the `Panther' sample in Figure \ref{fig:fig9c} as a `Bird' sample. The misclassification portrayed by the pair of Figure \ref{fig:fig9d} and \ref{fig:fig9i} is due to the combination of poor lighting and ineptitude of the amateur child puppeteer in creating `Llama' shadows. The model confuses the ear protrusions of the `Llama' sample to be the tentacular eyes of a `Snail' sample. For Figure \ref{fig:fig9e}, we can pontificate the misclassification reasons to be the overlapping of the fingers and the distorted angle at which the sample was captured. The `Snail' sample thereby gets wrongly classified as a `Bird' sample due to the existence of the analogous sample portrayed in Figure \ref{fig:fig9j}. 
The green entries along the diagonal of the confusion matrix in Figure \ref{fig:confusion_matrix} indicate the reasonably good classwise prediction performance of the \textsc{ResNet34} model, which is, to some extent, due to the perfectly balanced sample distribution in \hasper. 
\vspace{-4mm}
\subsection{Avenues of Improvement}
To reduce the number of misclassifications, models need to be imbued with the ability to learn certain nuanced features. In light of the contemporary image classification literature, we can opt to use Kolmogorov--Arnold Networks (KAN) \cite{liu2024kan} instead of a simple Multilayer Perceptron (MLP) as the classifier block. The use of Convolutional KANs \cite{bodner2024convolutional} has yielded good results in many image classification benchmarks. The task of image classification on \haspers can be dubbed as a silhouette classification task, which is why we can leverage topological features of the shadow contours to achieve better results \cite{lima2023image}. The silhouette polygonization algorithm (PoG), as proposed by \citet{goccmen2023polygonized}, may aid in achieving better classification accuracy. Other possible avenues may involve the use of ensemble methods coupled with voting schemes, or resorting to data augmentation with synthetically generated samples, but we defer the exploration of these hypotheses for future research endeavors.

%% file: sections/7_Conclusion.tex
\vspace{-2mm}
\section{Conclusion and Future Work}
In this paper, we explore an intriguing subject matter, hand shadow puppetry, in the realm of computer vision. We introduce \hasper, a novel dataset with a sizable collection of 15,000 hand shadow puppet images distributed across 15 classes, taken from both expert and amateur puppetry performance clips, by dint of a sophisticated frame extraction process involving optical flow estimation. We fine-tune 31 pretrained image classification models on \haspers to establish a benchmark for the dataset.
To explain and analyze the performance of the most erudite model, \textsc{ResNet34}, in comparison with other baseline models, we visually manifest its feature space using the $t$-SNE dimensionality reduction technique. We also perform thorough qualitative and error analyses for the \textsc{ResNet34} model. We envisage the possibility of developing applications for imparting the art of shadowgraphy, via mobile and embedded devices. We claim that this work is novel and significant since it is the first publicly available dataset and study on image classification benchmarking that focuses only on ombromanie.
There are many avenues in our work that warrant further investigation.
We hope to reconcile those desiderata by enriching our dataset with numerous permutations of arm positions and finger movements, preferably by employing more skilled individuals with varying palm and wrist structures, thereby creating more diverse silhouettes.
We also plan to experiment with a gesture detection technology such as MediaPipe\footnote{MediaPipe --- \url{https://developers.google.com/mediapipe}} or Microsoft Kinect\footnote{Kinect for Windows --- \url{https://learn.microsoft.com/en-us/windows/apps/design/devices/kinect-for-windows}} for leveraging depth coordinates of hand landmarks \cite{Mahmud2023}, and assess their efficacy in classifying hand shadow puppets.
\vspace{-2mm}

%% file: sections/8_Acknowledgments.tex
\section{Acknowledgments}
We convey our heartfelt gratitude, in advance, to the anonymous reviewers for their constructive criticisms and insightful feedback which will surely be conducive to the improvement of the research work outlined in this paper. We also appreciate the Systems and Software Lab (SSL) of the Islamic University of Technology (IUT) for the generous provision of computing resources during the course of this project. Additionally, we wish to acknowledge Shahriar Ivan, Department of Computer Science and Engineering, Islamic University of Technology, for his invaluable assistance in proofreading and offering a preliminary review of this manuscript. We further thank Mohammad Ishrak Abedin and Reaz Hassan Joader, from the same department, for lending their assistance, which was instrumental in refining this paper's illustrations and diagrams. Syed Rifat Raiyan, in particular, wants to thank his parents, Syed Sirajul Islam and Kazi Shahana Begum, for everything.
\section*{Declaration of Competing Interest}
The authors declare that they have no known competing financial interests or personal relationships that could have appeared to influence the work reported in this paper.

\section*{Data Availability}
The codes and datasets generated, written, and/or analyzed during the research project are available in the \haspers GitHub repository (\url{https://github.com/Starscream-11813/HaSPeR}) and in the \haspers Hugging Face repository (\url{https://huggingface.co/datasets/Starscream-11813/HaSPeR}).

\section*{Funding Sources}
This research did not receive any specific grant from funding agencies in the public, commercial, or not-for-profit sectors. All associated costs were self-funded by the authors. Access to computing resources was provided by the Systems and Software Lab (SSL) of the Islamic University of Technology (IUT).